\documentclass[jair,twoside,11pt,theapa]{article}

\newcommand{\papertitle}{Human Activity Recognition in an Open World}

\usepackage[implicit=true,unicode=true,
 bookmarks=true,bookmarksnumbered=true,bookmarksopen=true]
 {hyperref}
\hypersetup{
  pdftitle={\papertitle},
  pdfauthor={Derek S. Prijatelj},
  pdfsubject={\papertitle},
  pdfkeywords={Human Activity Recognition, Open World Recognition, Novelty Detection, Incremental Learning},
  colorlinks=true,
  allcolors=black,
  pdfpagelayout=OneColumn,
  pdfnewwindow=true,
}

\usepackage[pscoord]{eso-pic}

\newcommand{\placetextbox}[4]{
  \setbox0=\hbox{#4}
  \AddToShipoutPictureFG*{
    \if#3r
    \put(\LenToUnit{\paperwidth-#1},\LenToUnit{\paperheight-#2}){\vtop{{\null}\makebox[0pt][r]{\begin{tabular}{r}#4\end{tabular}}}}%
    \else
    \put(\LenToUnit{#1},\LenToUnit{\paperheight-#2}){\vtop{{\null}\makebox[0pt][l]{\begin{tabular}{l}#4\end{tabular}}}}%
    \fi
  }%
}%

\usepackage{jair, theapa, rawfonts}
\jairheading{81}{2024}{935-971}{12/2022}{12/2024}
\ShortHeadings{\papertitle}
{Prijatelj, Grieggs, Huang, Du, Shringi, Funk, Kaufman, Robertson \& Scheirer}
\firstpageno{935}

\usepackage{refstyle}

\usepackage{times}
\usepackage{epsfig}
\usepackage{graphicx}
\usepackage{amsmath}
\usepackage{amssymb}
\usepackage{booktabs}

\newcommand{\HARCodeURL}{\url{https://github.com/prijatelj/human-activity-recognition-in-an-open-world}}

\let\svthefootnote\thefootnote
\newcommand\blankfootnote[1]{%
  \let\thefootnote\relax\footnotetext{#1}%
  \let\thefootnote\svthefootnote%
}
\let\svfootnote\footnote
\renewcommand\footnote[2][?]{%
  \if\relax#1\relax%
    \blankfootnote{#2}%
  \else%
    \if?#1\svfootnote{#2}\else\svfootnote[#1]{#2}\fi%
  \fi
}

\newcommand\blfootnote[1]{%
  \begingroup
  \renewcommand\thefootnote{}\footnote{#1}%
  \addtocounter{footnote}{-1}%
  \endgroup
}

\usepackage{enumitem}
\usepackage{subcaption}

\DeclareCaptionSubType*{figure}

\usepackage[mathscr]{eucal}
\DeclareMathAlphabet\mathrsfso      {U}{rsfso}{m}{n}

\usepackage{algorithm}
\usepackage{algorithmic}

\usepackage{multicol}

\usepackage{xr}

\makeatletter
\newcommand*{\addFileDependency}[1]{%
  \typeout{(#1)}
  \@addtofilelist{#1}
  \IfFileExists{#1}{}{\typeout{No file #1.}}
}
\makeatother

\newcommand{\PredSpace}{\mathcal{\Pred}}

\newcommand{\Input}{X}

\newcommand{\MeasurableInput}{\mathrsfso{\Input}}

\newcommand{\sampleInput}{x}

\newcommand{\Output}{Y}

\newcommand{\MeasurableOutput}{\mathrsfso{\Output}}

\newcommand{\sampleOutput}{y}

\newcommand{\Pred}{{\hat{Y}}}

\newcommand{\PredictorFunc}{F}

\newcommand{\samplePredictorFunc}{f}

\newcommand{\eg}{\textit{e.g.}}

\begin{document}
\title{\papertitle}

\author{\name Derek S. Prijatelj \href{https://orcid.org/0000-0002-0529-9190}{\includegraphics[height=.8em]{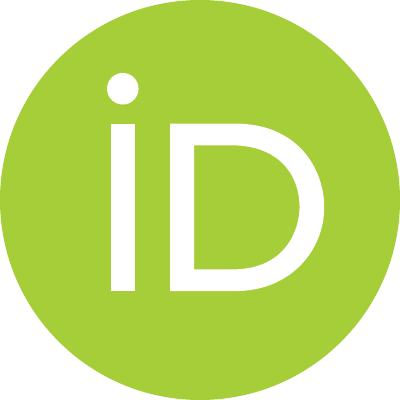}} \email dprijate@nd.edu\\
\addr Department of Computer Science and Engineering\\
University of Notre Dame\\
Notre Dame, IN 46556, USA\\
\vspace{-1.1em}
\AND
\name Samuel Grieggs \email sgrieggs@iup.edu\\
\addr Department of Mathematical and Computer Sciences \\
Indiana University of Pennsylvania\\
Indiana, PA 15705, USA\\
\vspace{-1.1em}
\AND
\name Jin Huang \email jhuang24@nd.edu\\
\addr Department of Computer Science and Engineering\\
University of Notre Dame\\
Notre Dame, IN 46556, USA\\
\vspace{-1.1em}
\AND
\name Dawei Du \email cvdaviddo@gmail.com\\
\name Ameya Shringi \email ashringi@google.com\\
\name Christopher Funk \email christopher.funk@kitware.com\\
\addr Kitware, Inc.\\
1712 Route 9, Suite 300, Clifton Park, NY 12065, USA\\
\vspace{-1.1em}
\AND
\name Adam Kaufman \email awk11@outlook.com\\
\name Eric Robertson \email  robertsone361@gmail.com\\
\addr PAR Government\\
421 Ridge St, Rome, NY 13440, USA\\
\vspace{-1.1em}
\AND
\name Walter J. Scheirer \email walter.scheirer@nd.edu\\
\addr Department of Computer Science and Engineering\\
University of Notre Dame\\
Notre Dame, IN 46556, USA\\
}

\maketitle
\thispagestyle{empty}

\placetextbox{3cm}{1.2cm}{l}{\footnotesize Journal of Artificial Intelligence Research 81 (2024) 935-971}
\placetextbox{3cm}{1.2cm}{r}{\footnotesize  Submitted 12/2022; published 12/2024}
\placetextbox{3cm}{25.4cm}{l}{\footnotesize  
    \copyright 2024 The Authors. 
    Published by AI Access Foundation under Creative Commons Attribution License CC BY 4.0.
}

\vspace{-2em}
\begin{abstract}
    Managing novelty in perception-based human activity recognition (HAR) is critical in realistic settings to improve task performance over time and ensure solution generalization outside of prior seen samples. 
Novelty manifests in HAR as unseen samples, activities, objects, environments, and sensor changes, among other ways.
Novelty may be task-relevant, such as a new class or new features, or task-irrelevant resulting in nuisance novelty, such as never before seen noise, blur, or distorted video recordings.
To perform HAR optimally, algorithmic solutions must
    be tolerant to nuisance novelty,
    and learn over time in the face of novelty.
This paper
    1) formalizes the definition of novelty in HAR building upon the prior definition of novelty in classification tasks,
    2) proposes an incremental open world learning (OWL) protocol and applies it to the Kinetics datasets to generate a new benchmark KOWL-718,
    3) analyzes the performance of current state-of-the-art HAR models when novelty is introduced over time,
    4) provides a containerized and packaged pipeline for reproducing the OWL protocol and for modifying for any future updates to Kinetics.
The experimental analysis includes an ablation study
of how the different models perform 
under various conditions as annotated by Kinetics-AVA.
The code may be used to analyze different annotations and subsets of the Kinetics datasets in an incremental open world fashion, as well as be extended as further updates to Kinetics are released.
\end{abstract}

%
\begin{figure}[t]
    \centering
    \includegraphics[width=.7\linewidth]{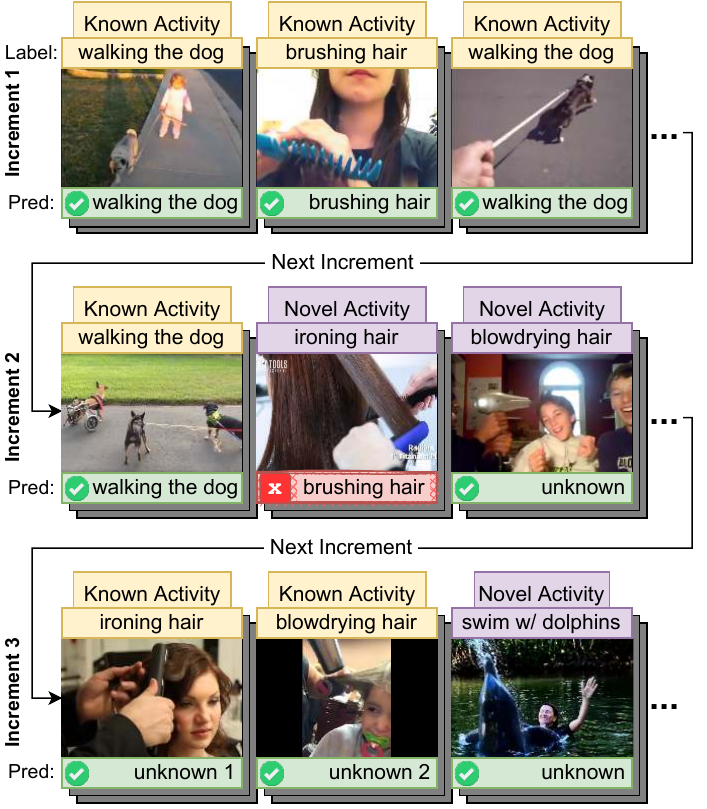}
    \vspace{-0.5em}
    \caption{
        \small
        An example of open world visual human activity recognition starting from  Kinetics-400~\cite{kay_kinetics_2017} known activities and \textbf{incrementally learning novel activities} from Kinetics-700-2020~\cite{carreira_short_2019,smaira_short_2020}.
        In Increment 2, without handling novelty, all unknown activities would be misclassified, when it is more correct to label them as ``unknown.'' 
        In an open world, learning unknown activities is desired, as in Increment 3 where ``ironing hair'' and ``blowdrying hair'' have been learned by the predictor and represented with stand-in unknown classes 1 and 2 to differentiate from the general unknown class.
        These stand-in labels remain until the predictor is given (or determines) a human readable label.
        \label{fig:teaser}
        \vspace{-1.2em}
    }
\end{figure}

\section{Introduction}
Human activities consist of various visual phenomena whose mappings to semantic classes are often noisy and uncertain.
These activities may range from swimming alone in the ocean, to playing in team sports such as football, or to reading an academic article.
Along with the activities, the environments where they take place may also be very different~\shortcite{minh_dang_sensor-based_2020}.
The task of Human Activity Recognition (HAR) is to recognize and classify these activities in visual recordings~\shortcite{beddiar_vision-based_2020}.
Given the extent of possible human activities and the variation within a single activity, datasets are often under-sampled representations of the task.
That incomplete representation results in a generalization problem for the HAR predictors when they encounter novel human activities, including novel instances of known activities.
This necessitates learning over time as more data is released and serves as a practical example of incremental open world learning.

The Open World Learning (OWL) setting~\shortcite{bendale_towards_2015,boult_learning_2019,langley_open-world_2020} is designed to expect novelty, including the lack of exhaustive samples for known classes and the lack of any samples for unseen, novel classes.
As such, OWL is a more practical experimental setting for HAR that has yet to be introduced in the HAR literature.
OWL may take on the form of incremental learning~\cite{belouadah_comprehensive_2021} where new batches of samples are made available over the time-steps.
The task is to perform HAR and learn the activity classes over time as new samples are introduced.
An instantiation of HAR can be found in the Kinetics datasets, which have been released since 2017, introducing new classes and samples, including
    Kinetics-400~\cite{kay_kinetics_2017},
    Kinetics-600~\shortcite{carreira_short_2018},
    Kinetics-700-2020~\cite{carreira_short_2019,smaira_short_2020},
    and Kinetics-AVA~\shortcite{li_ava-kinetics_2020}.
If we refer to Kinetics-700 we mean the 2020 updated version.
Ideally,
HAR predictors should be comparable and generalize across the Kinetics datasets.
Given the inherent lack of samples to characterize all human activities, HAR predictors must learn over time as data is made available, as in the case of the Kinetics datasets, and they should adapt to novelties that affect task performance~\cite{bendale_towards_2015}.
To evaluate such predictors, we define an OWL protocol with accompanying code to configure and run a Kinetics Open World Learning (KOWL) experiment.

\paragraph{Contributions}
To better evaluate predictors' generalization in Open World HAR,
we introduce:
\begin{enumerate}[noitemsep,nolistsep]
\item
A definition of Open World HAR in Section~\ref{sec:formal} as an instantiation of the newly introduced framework by \shortciteA{boult_towards_2021}, which establishes a common language for novelty across AI domains;
\item
An OWL protocol in Section~\ref{sec:owl_protocol} to create experiments to analyze novelty in HAR datasets,
     applied to
        Kinetics 400~\cite{kay_kinetics_2017},
        600~\cite{carreira_short_2018},
        and 700-2020~\cite{smaira_short_2020} to create KOWL-718, accompanied by baseline predictors for benchmarking;
\item
A detailed analysis in Section~\ref{sec:exps} of the performance of pre-existing state-of-the-art visual HAR models in the presence of novelty, including X3D~\cite{feichtenhofer_x3d_2020} and TimeSformer~\cite{bertasius_is_2021}.

\item
Pipeline code for configuring and running the open world HAR experiments\footnote{
    Code repository with containers for reproducing and extending: 
    \HARCodeURL
    }
    with a Docker image to both reproduce the Kinetics experiments below, as well as enable any configuration for open world Kinetics experiments in the future. 
\end{enumerate}
\section{Background and Related Work}
\label{sec:bg}
Fundamentally, novelty is something that is new to something else.
This could be related to the task being learned, such as a sample of a new class in the evaluation set, or new features that correlate strongly with a known class~\cite{chalmers_strong_2006}.
Such things are deemed ``task-relevant'' if they contain information to accomplish the task~\shortcite{kullbackInformationSufficiency1951,tishby_information_1999,gershman_novelty_2015}.
Novelty can also be task-irrelevant, for example the color of one's new clothes is a feature of a video that should be mostly uncorrelated with the activity of standing in place. 
Such task-irrelevant novelty is termed  ``nuisance novelty'' and its affect on task performance is to be minimized.
Such concepts of novelty have been defined in a formal abstraction with a relation to agents performing their tasks in general AI settings by \citeA{boult_towards_2021}.
Note that \citeA{boult_towards_2021} define their abstraction referring to the task's algorithmic solution as an ``agent,'' where in this work we refer to them as ``predictors.''
The primary actions of the predictors observed in this work are their predicted classification probability vectors and an optional feedback request through the ordering of the sample identifiers based on priority of interest to the predictor.

Open Set Recognition~\shortcite{scheirer_toward_2013,geng_recent_2021}, Anomaly Detection~\shortcite{pang_deep_2021,ruff_unifying_2021}, Novelty Detection~\shortcite{roitberg_informed_2018}, and Out-of-Distribution Detection~\shortcite{zaeemzadeh_out--distribution_2021} are all intended to detect when a sample occurs that contains novel information that was not represented by the prior data.
For simplicity, this paper refers to these similar tasks solely as ``novelty detection.''
In terms of classification tasks, as in Open Set Recognition, a sample with a large amount of novel information cannot be reliably classified as a known class, and instead should be labeled as an unknown class.
Novelty detection does not involve incremental learning to adapt to novelty; it only involves the detection of novelty.
This may be seen as detecting out-of-distribution samples based on a single time-step.

OWL~\shortcite{boult_learning_2019,langley_open-world_2020,geng_recent_2021} is an example of learning-based classification or recognition with novelty in mind.
OWL is the natural extension of novelty detection where new classes are learned alongside the known classes.
Recent work has proposed a practical OWL protocol for image classification~\shortcite{dhamija_self-supervised_2021} as an extension of Incremental Classifier and Representation Learning~\shortcite{rebuffi_icarl_2017}, but no OWL protocol, benchmark dataset, or model exist for HAR at the time of writing.
In this paper we apply OWL to HAR, a more visually complicated problem than static image classification.
Our work has multiple key differences to this prior work, including a substantially more complete performance analysis of the different classification tasks over time, including the predictor's novelty reaction time.
Our protocol records each confusion matrix and its derived measures twice per increment, while theirs only records the cumulative classification accuracy of the known classes alone, known classes with a catch-all unknown class, and binary classification of novelty detection.
Section~\ref{sec:eval} addresses the nuance in the confusion matrix's derivative measures and what information is lost when collapsing or removing subsets of labels.
The confusion matrix captures all classification information over a period of time, ignoring the order of samples.
We measure the predictor's novelty reaction time to capture this time-series performance.
We also incorporated the budgeted $b\%$ feedback into the protocol and assessed performance given varying feedback budgets.
Partial feedback $0\% < b < 100\%$ results in another subtask for the predictor to prioritize for which samples it requests feedback.

Another difference to \citeA{dhamija_self-supervised_2021} is that in Section~\ref{sec:prior_know_novelty} we emphasize that not knowing how the predictor's prior knowledge relates to the current experiment makes knowing what is novel impossible, as in pretrained models, which invalidates the assessment of predictor performance when encountering novelty.
This distinction is important, especially when considering the current dominant trend in AI of using large pretrained models for transfer learning to down-stream tasks~\shortcite{qiu_pre-trained_2020} where what is known prior is often left implicit, resulting in uncertainty of what is novel to the predictor.

There are some HAR works related to Open Set Recognition~\shortcite{roitberg_open_2020,gutoski_deep_2021,brighi_activityexplorer_2021,inacio_osvidcap_2021}, but none that attempt to provide a benchmark or solution for OWL in HAR, which involves incremental learning in a world with unknown human activity classes and their associated videos.
Three methods approach this goal.
Deep Metric Learning for Open Set HAR~\cite{gutoski_deep_2021} uses metric learning to improve feature representation models for open set recognition, but the associated experiments do not involve incremental learning.
Activity Explorer~\cite{brighi_activityexplorer_2021} is described as having an ability to recognize potentially new classes, but is only assessed within an OSR context, rather than in an OWL context. It does not include logic on how to update classes over time, leaving it to the human user.
OSVidCap~\cite{inacio_osvidcap_2021} is for the task of Video Captioning, which is related to HAR, but is not HAR specifically. Like the other two approaches, it is evaluated in an Open Set Recognition mode, not OWL.

A feature representation space that generalizes to the task and captures the world's task-relevant information is an important part in solving HAR with novelty.
The state of representation learning consists of large semi-supervised deep neural networks with copious amounts of data~\shortcite{radford_learning_2021,schiappa_self-supervised_2022}, or generative models that leverage adversarial examples or augmentations for contrastive learning~\cite{ho_contrastive_2020}.
Feature representation is important for OWL because it determines what information is available to learn the task for the down-stream components, such as the classifier (see Figure~\ref{fig:owhar}).
Not only do OWL classifiers need to be able to learn new classes over time, but ideally their representation learners also are able to adapt to include missing task-relevant information.
This is similar to the popular pretrained models in natural language processing where a semi-supervised model, such as a transformer, is pretrained on an upstream task with a large dataset and in doing so learns information relevant to many downstream tasks~\cite{qiu_pre-trained_2020}.
With a feature representation that contains information not only relevant to the current downstream task but also contains possibly future relevant task information, the classifiers are able to distinguish between known and unknown, learning novel classes over time.
Incremental learning model types that perform fine-tuning are one way to adapt the feature representation space over the increments~\cite{belouadah_comprehensive_2021}.
Frozen or fixed representations rely on a feature representation that already contains the information relevant to learning novel classes in the future, where any downstream fine-tuning or classifier changes will change their learned function to use task-relevant information in the feature representation ignored prior to the novel class being introduced, limited by the data processing inequality~\cite[Sec.~2.8]{cover_elements_1991}.
\section{Formalization of Open World HAR}
\label{sec:formal}
\cite{boult_towards_2021} provided a convenient framework for formalizing novelty problems in AI.
A formal definition of novelty for HAR provides a standardized way to communicate and understand the impact of novelty on the performance of a learning-based system meant to recognize activities.
This enables informed discussion about the nuances of OWL, including new classes and new types of task-relevant phenomena such as new visual features, \eg, associating the visual depiction of a football with the sport. 

The formalization starts with defining a task. 
HAR is a classification task where the predictor learns a deterministic function
$\samplePredictorFunc \in \{\PredictorFunc: \MeasurableInput \rightarrow \MeasurableOutput\}$ that maps from the visual space $\MeasurableInput$ to the space of human activity classes $\MeasurableOutput$.
A predictor estimates this function
from paired data samples $(x, y)$: a video sample $x \in \MeasurableInput$ and the most likely human activity class label $y \in \MeasurableOutput$.
The paired spaces $(\MeasurableInput, \MeasurableOutput)$ form the system state space in which the task is performed and novelty may occur.
Given the space of class labels $y \in \MeasurableOutput$ is never complete in OWL, novel activity classes may be encountered in future HAR video samples.
As such, the problem of learning an optimal mapping $\samplePredictorFunc$ from data persists over time in OWL where one of the class labels stands for the unknown classes.
This unknown class is used to represent the detection of novel classes.
Novelty recognition can then identify distinct unknown classes from the detected novel classes through a semi-supervised learning approach.

A performance measure is necessary to define the task, such as regret, which is
the quantification of the undesirability a predictor's $\Pred$ decision or the resulting state of that decision.
which may differ from the dissimilarity measures over $\MeasurableInput$ and $\MeasurableOutput$.
A dissimilarity measure is a quantification of how different states are to one another, possibly with a threshold for binary dissimilarity decisions and possibly task-specific to ignore some task-irrelevant information.
The experience determines what is known and thus what is novel given the dissimilarity measures over the spaces.
It is a time-indexed history of the states that have occurred over time.
    The final piece is the State Transition Function, which is
    a function that moves from the current state to the next state given a change in time or an action taken by the predictor.
    For HAR, we consider the smallest time-step as the index of the current sample, and mostly consider the index of the OWL increments.
    In the following experiments, we refer to time-steps at the level of half an increment where the whole step is receiving the increment's videos and the half step is receiving any feedback labels for those videos.
    When given an increment's videos, the predictor decides the class, detects novelty, possibly updates its state, and, if possible, may request feedback.

Our paper primarily focuses on the spaces in which novelty may occur related to classifying human activities: the input space $\MeasurableInput$, the predictor output space $\PredSpace$, and the ground truth target label space $\MeasurableOutput$ obtained from a dataset.
The relation of these to the primary three is as follows:
\begin{itemize}
    \item 
    The world defines that which is considered the truth to be learned, as defined by the dataset, meaning the paired mapping of video $\sampleInput$ to ground truth label $\sampleOutput$.
    The dataset as in order of presentation to the predictor forms the world experience.
    The order of presentation is based on the chosen experimental design and serves as the state transition function of the observed world.
    The regret measures are based on the relative comparison of predictions to the ground truth labels.
    Section~\ref{sec:eval} goes into explicit detail for any KOWL experiment.
    In this paper, all of these parts are controlled and executed by the ``evaluator,'' as seen in Figure~\ref{fig:single_step}, which is the algorithmic execution of the experiment.
    \item
    The observation space relates to how these human activities were captured visually, such as camera sensors.
    The original dataset determines the source of the visual recordings and may include any metadata defining those recordings.
    Nuisance novelty can occur here, including never before experienced perspective rotations, hue shifts, blurs and noise.
    \item
    The predictor space consists of the predictor's output space, experience, and internal state.
    Internal regret measures are those such as the optimization function in which the predictor optimizes as it learns, such as losses for neural networks.
    State transition functions for the predictor are its update functions as it receives videos and any feedback from the evaluator.
    See Section~\ref{sec:abstract_predictor} for the complete abstraction and Section~\ref{sec:models} for examples.
\end{itemize}
More detailed explanations of the world, observation, and predictor spaces as structured in~\cite{boult_towards_2021}  for defining novelty in HAR are available in Appendices~\ref*{sec:spaces_novelty} and \ref*{sec:har_novelty_sets}.

\subsection{Nuisance Novelty}
\label{sec:nuisance}
To learn a task, the predictor
maximizes the task-relevant information in the predictions and filters out the task-irrelevant information~\cite{kullbackInformationSufficiency1951,tishby_information_1999}, as measured by the correlation between the predictions and the ground truth labels.
Task-irrelevant information is noise or other irrelevant phenomena that occurs in the world and is possibly captured by the sensor.
Nuisance novelties are never before encountered task-irrelevant phenomena that the predictor is to be tolerant to in order to maintain task performance.
An example in HAR is that if someone is dancing, then in general their clothes should not be a strongly correlated feature to the recognized dancing activity for the predictor to generalize across dancing videos.
Nuisance novelty is defined in \cite{boult_towards_2021} as a novelty for which a pair of regret measures significantly disagree. In this example, the predictor may falsely detect novelty with respect to a new activity class when in reality it is only someone dancing in a never-before-seen colored shirt and the activity class label is still dancing.

Often is the case that some novelty may occur in the world, such as a sensor being  knocked over and the visual input being rotated.
The input being rotated is novel if rotations were never seen in training, but if all of the actions being recorded are known, then the predictor is desired to be tolerant to rotations of its field of view.
However, if this rotation causes the predictor to repeatedly detect novel classes, despite them being known, then this is a nuisance novelty. 
Predictors need to maintain task performance in the face of such nuisance novelties.
Further explanations of the different types of novelty with human activity are provided in Appendix~\ref{sec:har_novelty_sets}.

\begin{figure*}[t]
    \centering
    \includegraphics[width=1.05\linewidth]{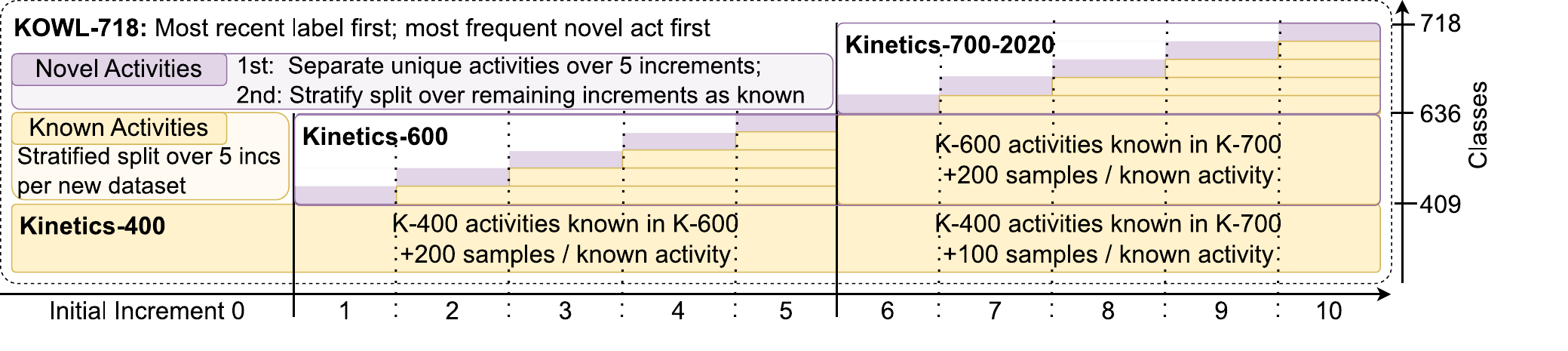}
    \caption{
        \small
        A visual representation of KOWL-718's class density per increment as used in the experiments
        when the OWL protocol is applied to Kinetics-400 to Kinetics-700.
        The incremental learning uses a unified mapping of the Kinetics activity classes based on the most recent label first.
        Each increment contains a new set of unknown activities and the prior known activities persist as long as the original Kinetics datasets contained samples for them.
        The most frequent novel classes were introduced first within their respective dataset to help balance the samples over increments.
        See Section~\ref{sec:protocol}.
    }
    \label{fig:protocol}
\end{figure*}

\section{Open World Learning Protocol for HAR}
\label{sec:owl_protocol}

\label{sec:protocol}
Any labeled HAR dataset may be partitioned in such a way that yields an OWL HAR dataset whose increments consist of videos depicting known and novel activities.
The following focuses on forming increments on the training data split, as depicted in Figure~\ref{fig:protocol}.
The validation and testing data splits are similarly be split using the same class sets per increment such that they are aligned to the training data.
Given the number of desired increments $N$, a labeled dataset that is an unordered sequence of pairs $(x, y)$, and the starting known label set $K \subset \MeasurableOutput$, then
    the starting unknown label set $U$ consists of the unique labels not in $K$, where $U = \{y : y \in \MeasurableOutput, y \notin K\}$.
The pairs with known labels may be split into the $N$ increments in any way desired, such as a stratified split that maintains the balance of the known labels across the increments as much as possible. 
With the known labels partitioned, the increments must also introduce novel classes.
To do so, the unknown labels are separated uniformly across the $N$ increments such that there are $\lfloor\frac{|U|}{N}\rfloor$ unknown classes within in the first $N-1$ increments and the remainder in the final increment.
At this point, every increment shares the known classes and has a subset of unknown classes disjoint to all other increments.

Given that we want the novel classes introduced at each increment to remain present in future increments,
each increment's unknown class samples need to be spread across the remaining increments that follow in time.
To counteract class imbalance, we order the unknown classes by descending order of sample frequency.
Then, insert them by the most frequent unknown class first over the time ordered increments.
The increment at time-step $t$ has its unknown label set $U_t$ that is stratified split into $n - t$ splits, where $n \in [0, N-1]$ using zero-indexing.
One of those splits remains in the current $t$ increment.
The other $n - t - 1$ splits are added to their respective future increment, where $t \in \{0, 1, ..., n-2, n-1\}$.
This results in an OWL HAR dataset that can be used to train and evaluate a predictor incrementally where the sample index corresponds to time and there are $N$ class distributional changes where novel classes are introduced.
The distribution here is the ground truth labels as a random variable whose set of possible labels at each increment grows.
If training, validation, and testing splits are not predefined, each increment from above may be split using the typical methodologies, such as 8:2 train-to-test ratios, or even multi-fold cross validation.
If they are pre-defined, then the validation and testing splits are separated into increments based on their classes corresponding to the classes in the training split, ensuring the validation and testing splits contain the training classes at each increment.

\begin{table}[t]
    \scalebox{.9}{{\small
\begin{tabular}{ll|lllll|lllll}
\toprule
Increment & 0          &      1 &      2 &      3 &      4 &      5 &      6 &      7 &      8 &      9 &     10 \\
\midrule
Classes  &&&&&&&&&&& \\
\multicolumn{1}{r}{Known} & 409 &    409 &   455 &   501 &   547 &   593 &   636 &   653 &   670 &   687 &   704 \\
\multicolumn{1}{r}{Novel} & 0 &     46 &     46 &     46 &     46 &     43 &     17 &     17 &     17 &     17 &     14 \\

\midrule
Train Samples  &&&&&&&&&&& \\
\multicolumn{1}{r}{Known} & 218371 &  16082 &  19148 &  24661 &  32391 &  44687 &  34287 &  36112 &  38527 &  41855 &  47128 \\
\multicolumn{1}{r}{Novel} & 0 &   3163 &   5513 &   7730 &  12298 &  25538 &   1870 &   2415 &   3331 &   5273 &  10144 \\

\midrule

Test* Samples    &&&&&&&&&&& \\
\multicolumn{1}{r}{Known} & 35256 &   1128 &  307 &  491 &   732 &   1096 &  4892 &  4990 &  5197 &  5473 &  5883 \\
\multicolumn{1}{r}{Novel} & 0 &   179 &   232 &   266 &   372 &   612 &   163 &   208 &   277 &   411 &   684 \\
\bottomrule
\end{tabular}
}
}
    \vspace{-0.5em}
    \caption{
        \small
        \textbf{KOWL-718}'s total known and novel classes with their samples \textbf{per increment} using the configuration in Section~\ref{sec:exps}:
        Most recent label first, first come first assigned data split, and most frequent novel class first per future dataset.
        Increment 0 uses the Kinetics-400 data splits without the 18,016 validation samples listed above.
        The test split consists of Kinetics-600's validation and test splits for the increments in range [1, 5] and the Kinetics-700 validation split for increments in range [6, 10] given the Kinetics-700-2020 test set labels are not yet publicly released.
        Totaling 630,524 training samples and 68,849 test samples overall.
        *See Appendix~\ref{sm:sec:data} for a breakdown and a note on link rot where videos are lost.
        \vspace{-0.5em}
    }
    \label{tab:kowl718}
\end{table}

This protocol was used to construct a Kinetics Open World Learning experiment with 718 classes (KOWL-718) that followed the chronological release of the original datasets.
KOWL-718 starts with Kinetics-400~\cite{kay_kinetics_2017} as the initial increment, followed in order by Kinetics-600~\cite{carreira_short_2018} split into 5 increments, and then Kinetics-700-2020~\cite{carreira_short_2019,smaira_short_2020} split into 5 increments, as seen in Figure~\ref{fig:protocol} and Table~\ref{tab:kowl718}.
The human activity labels between the original Kinetics datasets differ as more were added and existing classes were either removed, renamed, or reorganized.
Some classes were separated into two or more classes, or samples that were in one class were moved to another.
Given this inconsistency between labels, we constructed a Kinetics unified label set that prioritizes the most recent changes made to the datasets first: following Kinetics-700, then Kinetics-600, then Kinetics-400.
This means that the samples that were in prior versions and are still in Kinetics-700 have a Kinetics-700 class, while the samples removed from future versions keep their most recent class.
The resulting most-recent-first label set totals 718 activities.
Following this label order enables the use of existing pretrained predictors at the checkpoints which correspond to the end of the Kinetics datasets, if so desired.
However, see Section~\ref{sec:prior_know_novelty} on the nuance of this with respect to novelty.
The unified Kinetics criteria are as follows:
\begin{itemize}[noitemsep,nolistsep]
    \item \textbf{Label assignment}:
        \textit{Most recent label first}.
        Thus, Kinetics-700-2020 labels are used for all samples when available.
        If the samples in Kinetics-600 are not used in Kinetics-700, 
        then they are assigned activities from Kinetics-600.
        Same applies for samples in Kinetics-400 that are unused in either Kinetics-600 or -700. 
        Under this configuration, some activities are not represented in later datasets, which matches the actual release of the data.
    \item \textbf{Data split assignment}:
        \textit{First come, first assigned}.
        This follows how the data were released throughout the years, keeping persistent samples within their first assigned data split, and avoiding the issue of training samples from earlier Kinetics datasets being used in validation or test sets of later Kinetics datasets.
    \item \textbf{Introduction of novel activities}:
        \textit{Most frequent first}. 
        When adding a dataset as released in time, this means novel activities within Kinetics-600 and Kinetics-700-2020, the most frequent novel activity is introduced first such that there are more samples of that class across the $N=5$ increments.
        This aids in balancing the number of samples across the dataset's remaining increments.
\end{itemize}

While using the most recent labels means those re-labeled samples from older datasets are more accurate, it does result in class imbalance in the prior Kinetics datasets.
As such, when using the OWL protocol to partition Kinetics-600 and -700 samples whose labels were fewer than the remaining increments, they were simply included only within the increment they were introduced in.
This occurs within the increments' validation splits due to their smaller size relative to the training and testing splits.
Given this we combined the validation and test splits of Kinetics-600 into our test set at those increments as seen in Figure~\ref{fig:protocol} and Table~\ref{tab:kowl718}.
Due to using the most recent label first scheme, the Kinetics-400 and -600 training splits have classes that are not contained within their validation or testing sets throughout the incremental learning until the final increment.
The Kinetics-700 validation data serves as the test data in KOWL-718 because the original test labels are not publicly released.

\begin{figure}[t]
    \centering
    \includegraphics[width=0.75\textwidth]{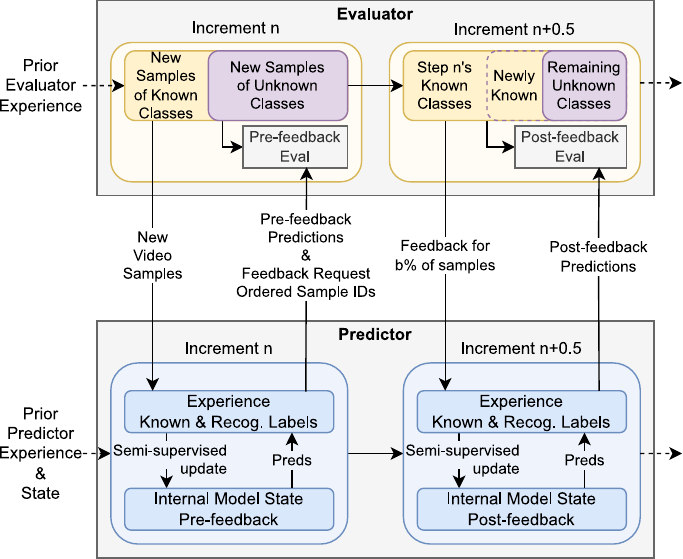}
    \caption{
        \small
        The $n$-th increment of the OWL protocol depicting the exchange of data between the evaluator and predictor as they update their internal state.
        An increment has a pre-feedback phase (whole step $n$) and a post-feedback phase (half step $n+0.5$), if any feedback is given.
        During pre-feedback, the evaluator gives new samples to the predictor without any labels.
        The predictor updates given this new unlabeled data, such as semi-supervised learning, and returns its predictions
        along with the samples ordered by priority for feedback.
        The evaluator assesses the pre-feedback predictions and saves the samples for which feedback was provided, if any.
        The ground truth class labels given as feedback are then known to the predictor.
        The predictor may update its state given any feedback it receives and perform a post-feedback prediction, which is then re-evaluated.
        This process repeats until the experiment's increments are exhausted.
        At the initial increment, prior knowledge obtained from external data, such as feature representations from other models, needs to be explicitly stated to be evaluated appropriately.
        See Section~\ref{sec:abstract_predictor}.
        \vspace{-1em}
    }
    \label{fig:single_step}
\end{figure}

\subsection{Evaluation}
\label{sec:eval}
Using an incremental OWL HAR dataset,  incremental learning starts with fitting on the training data in the initial increment that includes the known classes in $K$, which matches typical supervised learning classification.
As seen in Figure~\ref{fig:single_step}, the next increment's data is introduced and the predictor is evaluated on all of that data prior to any provided feedback, such as the ground truth labels.
Novelty detection and recognition of human activities is evaluated at this pre-feedback increment.
Each prediction may be a single nominal class or an estimated probability vector of the mutually exclusive classes.

With the pre-feedback class predictions, the evaluator measures the predictor's performance on HAR classification, novelty detection, and novelty recognition.
The confusion matrix captures all of the information from the samples of the marginal and joint distributions
    between the predicted most likely classes and the actual labels, which enables deriving measures often useful in evaluating classification and clustering.
The evaluation measures' types (bold) necessary to assess the predictors are as follows along with our chosen measure (italics):
\begin{enumerate}[noitemsep,nolistsep]
    \item \textbf{Exact Symbol Matching}: \textit{Accuracy}:
        The primary classification task is designed such that the predictor's output matches understandable class labels.
        For example, we want unobfuscated activity labels, such as ``walking the dog'' instead of ``class\_379''.
        Accuracy is a common measure that evaluates how many samples have the correct label, albeit susceptible to class imbalance.
    \item \textbf{Linear Correlation}:
        \textit{Matthews Correlation Coefficient (MCC)}:
        While the accuracy is easily human interpretable, at times it may be greater for ill-performing predictors, such as when a predictor outputs the most frequent label in an imbalanced label set.
        Linear correlation measures, such as Pearson's r coefficient or the equivalent Matthews Correlation Coefficient (MCC), have low values for such undesired behavior.
        These values range from $[-1, 1]$, where -1 is inversely or negatively correlated, 0 is no correlation, and 1 is positively correlated.
    \item \textbf{Task-Relevant Information}:
        \textit{Arithmetic Mean of Entropies Normalized Mutual Information}:
        Non-linear correlations ignore the exact symbol and focus on the information learned, which includes both linear and non-linear relationships.
        Mutual information measures the amount of task-relevant information in the predictions~\cite{tishby_information_1999,kinney_equitability_2014}.
        It is an unbounded measure and requires enough samples to properly estimate the actual information of the task and predictions.
        With that said, normalized variants ranging $[0,1]$ are available, such as the information quality ratio or the normalization by the arithmetic mean of the entropies, where the former is more exact when enough samples are available, and the latter is a better estimate when not enough samples are available, albeit not as interpretable. 
        When normalized, a mutual information
            of $1$ indicates the predictions contain all of the task-relevant information.
\end{enumerate}
To observe only one of these measures ignores important performance information in the others.
A well-performing predictor will maximize all of these point-estimate measures in- and out-of-sample.

\begin{figure*}[t]
    \centering
    \includegraphics[width=.9\linewidth]{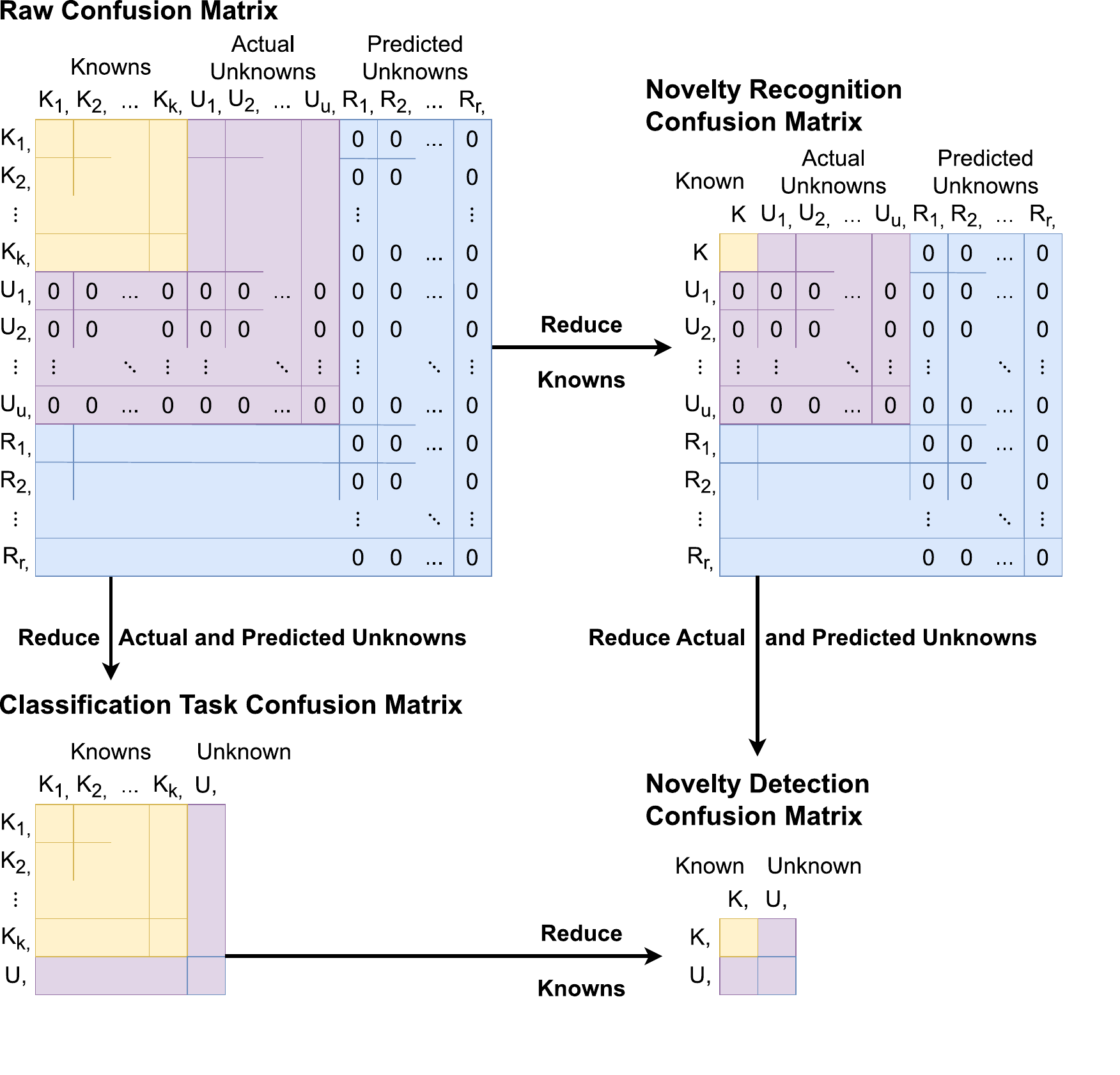}
    \vspace{-2.2em}
    \caption{
        \small
        The raw confusion matrix and its reductions are used to assess the activity prediction performance
        over a window of time, ignoring sample order. 
        The raw version compares the true known and unknown classes (columns) to the predictions (rows), including any recognized unknown classes.
        A predictor can never predict the true unknown labels, otherwise they'd be known.
        A predictor's recognized unknowns are their recognized class-clusters, which never match the true labels. 
        The zeros denote where no pair of predicted and actual labels can occur.
        A class reduction is an information lossy transform that focuses more on the performance on the unreduced classes by ignoring any confusion between the reduced classes.
        When the unknowns are reduced, novelty detection is assessed, ignoring recognition.
        See Section~\ref{sec:eval}.
        \vspace{-1em}
    }
    \label{fig:confusion}
\end{figure*}

Open world recognition, which is inherently a classification task, may be evaluated at each time step using a confusion matrix and any of its derived measures, such as those mentioned above.
Depending on how classes are treated, one may observe the ``raw'' confusion matrix where the paired predicted labels and unknown ground truth labels are read by the evaluator as written by the predictor and ground truth data. 
In this work, we use three different reductions of the confusion matrix's labels to assess three specific subtasks listed below for open world recognition.
These reductions are depicted in Figure~\ref{fig:confusion}.
In addition to evaluating these three subtasks, a fourth subtask is necessary to assess
the time-series information of reacting to novelty as it occurs through the incremental learning. In detail, the four subtasks are: 
\begin{enumerate}[noitemsep,nolistsep]
    \item \textbf{Classification Task}:
    The primary activity classes of interest along with a single unknown class to account for any novel or unknown classes.
    To evaluate this, reduce the predictor's recognized unknown classes into the general ``unknown'' class, indicating when the sample is deemed something other than that which is known. 
    This is the typical assessment used in the original Kinetics protocol, minus the catch-all ``unknown'' class.
    \item \textbf{Novelty Detection}:
    the binary classification of known vs. unknown.
    All known classes are reduced into a single ``known'' class, and all unknown classes are reduced into a single ``unknown'' class.
    \item \textbf{Novelty Recognition}: the unsupervised class-clustering separate from the classification of the known classes. 
    To evaluate this, reduce all known classes into a single ``known'' class.
    The measures derived from the resulting confusion matrix must account for the class labels not using the same symbols, even though they are the same cluster.
    This includes measures such as mutual information, as used in this work.
    \item \textbf{Novelty Reaction Time}: The time it takes to react to novelty, which in the case of HAR is measured in the number of samples till detection of novel activities after the first novel sample has occurred.
    Here, as in novelty detection, the known classes are reduced and the unknown classes are reduced, resulting in a binary classification assessment.
    The difference here is that the novelty reaction time measures how many samples pass after the first occurrence of a novel activity's sample.
\end{enumerate}

\begin{figure}
    \centering
    \includegraphics{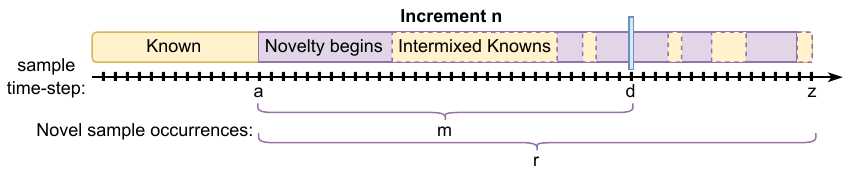}
    \vspace{-1.2em}
    \caption{
        \small
        Confusion matrices and their derivative measures ignore information across time, such as sample order, making it important to separately measure the novelty reaction time.
        This figure depicts the variables involved in the novelty reaction time measured in this work in Equation~\ref{eq:react}.
    }
    \label{fig:visual_react}
\end{figure}

    We use the following procedure to measure novelty reaction time within a single increment whose variables are visualized in Figure~\ref{fig:visual_react}:
    When the first novel sample occurs at the sample index $a$, record the number of novel samples $m$ observed before the first sample index of novelty $d$ is detected by the predictor at or after $a$.
    Let $z$ be the final sample index of the increment and $r$ be the total count of novel samples within the increment.
    The following harmonic mean is used to measure the novelty reaction time:
    \begin{equation}
    \label{eq:react}
        (\frac{
            (\frac{d-a}{z+1-a})^{-1} + (\frac{m}{r})^{-1}
        }{2}
        )^{-1}
        =
        \frac{2}{
            \frac{z+1-a}{d-a} + \frac{r}{m}
        }
    \end{equation}
    where $\frac{d-a}{z+1-a}$ is the fraction of samples that pass before novelty is first detected and $\frac{m}{r}$ is the fraction of total novel samples that occur before detection occurs.
    $z+1$ is used because it reflects the total number of samples with a zero index. If novelty is never detected, then that corresponding fraction would be 1, otherwise it approaches 1 in the limit of $z$ to infinity.
    Note that in the demonstration of KOWL-718 below, all samples in the increment are given to the predictor at once by the evaluator, rather than in ``sub-increments'' or batches, which is different from detecting novelty as the samples are given as all the information in the increment can be used to determine where the novelty occurs.
    If desired, one may assess novelty detection and reaction time at finer granularity, such as per-sample or per-batch.

All of these measurements require a clear distinction between what is known and what is novel.
In the majority of this paper's experimental demonstration in Section~\ref{sec:exps}, the known classes are based on what activity class labels the predictor knows within its experience at that moment in time, \textit{i.e.}, at the current half of the increment.
All of their samples are thus deemed known or novel to the predictor based on this alone.
This may be referred to as \textit{novel-to-predictor}. 
This is different from being \textit{novel-to-evaluator} where novelty is determined by whether the experiment has yet provided a class sample, referred to as ``actual novelty'' by \citeA{boult_towards_2021}.
The evaluator may record the measurements from the above four subtasks above from either perspective.

\subsection{Requesting Feedback}
\label{sec:feedback_request}
After evaluating the predictor on the new increment's data, feedback may be offered to it.
The
supervised
incremental learning paradigm~\cite{masana_class-incremental_2021,losing_incremental_2018} releases all of the labels ($b = 100\%$) to the predictor.
The experiment may instead only release a subset of the labels, such as none ($b=0\%$), or partial feedback, such as $b=50\%$ of the samples for this new increment.
Our OWL protocol gives the predictor another subtask of requesting feedback for itself from the evaluator after every pre-feedback evaluation.
This feedback request takes the form of an ordered list of the samples' unique identifiers.
The evaluator then determines how much feedback will be returned and returns it starting with the prioritized ordering requested by the predictor.
This experimental design matches human-in-the-loop feedback in practice where human labeling is costly and figuring out a sample subset of interest to be annotated is desirable.
Another option for feedback types is returning the performance measurements on the increment, which is simply a controlled validation dataset where the actual labels are not released, although we assess being given the ground truth labels directly.

\subsection{Evaluating Tolerance to Nuisance Novelty}
\label{sec:eval_nuisance}
The dataset and above measures assess the learning of task-relevant information, and how novel activities are handled over time.
It may be of use, as it is in practice, to assess the tolerance of the predictors to certain variables they are desired to be invariant against, such as task-irrelevant information or nuisance novelties.
To do this in HAR, the videos may be visually transformed in ways unseen in training and any corresponding change in the above performance measures may be recorded.
If they do not encounter a significant change in performance, then they are deemed tolerant to such visual transformations, such as perspective rotation, hue shifts, visual blur, or noise.
Otherwise, such transformations that dramatically affect performance can indicate predictor model weaknesses and areas of improvement.
This may be assessed such that different novel visual transforms occur over time and thus are part of learning over time.
Otherwise, the predictor's state may be saved and then reloaded in isolation to evaluate how the predictor at that moment would handle the encountered nuisance novelty.
The experiment in Section~\ref{sec:tolerance} takes the latter approach, loading the saved checkpoints of post-feedback internal model state for that increment and then continuing the incremental learning until the next feedback is given using only the visually transformed videos of the original increments' samples.

\subsection{A Template for OWL HAR Predictors}
\label{sec:abstract_predictor}
\begin{figure*}[t]
    \centering
    \includegraphics[width=\textwidth]{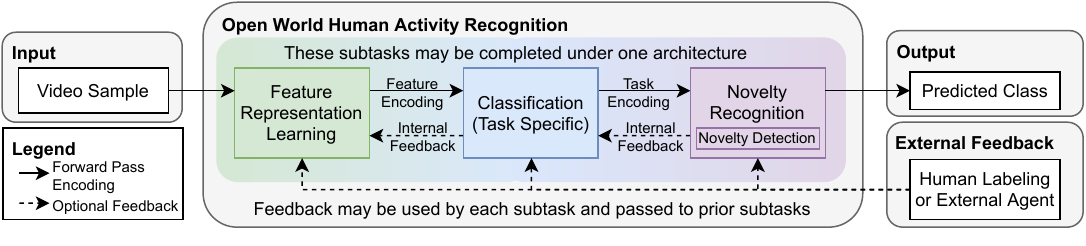}
    \caption{
        \small
        An OWL \textbf{predictor template} for HAR with novelty.
        Learning tasks in an open world requires the subtasks of feature representation, task-specific classification, and novelty recognition to not only detect, but also learn novel classes over time.
        Feedback is essential to inform the predictor of when to improve at subtasks.
        Upon request from the predictor, external feedback may be provided from humans as in active learning or from other predictors.
        0\%, 50\%, and 100\% human feedback is examined.
        More predictor details can be found in Section~\ref{sec:models}.
    }
    \label{fig:owhar}
\end{figure*}

Given the above formalization, a template for OWL predictors that manage novelty in a HAR domain may be defined to provide an outline of what such algorithms must include to perform well in practice.
Figure~\ref{fig:owhar} shows a template that includes the subtasks of feature representation learning, task-specific classification, and novelty recognition.
The predictors must account for performing the tasks over time and should adapt to novelties that affect task performance~\cite{bendale_towards_2015}.
These predictors may be able to request external feedback either from human experts as in active learning or from other predictors that are accessible to them.

The first subtask
of all OWL predictors for HAR is feature representation learning, 
which is the learning of the mapping of the input video sample to an informative encoding.
As mentioned in Section~\ref{sec:bg}, feature representation plays a key role in not only learning HAR but also detecting and recognizing novel activities as it serves as the filter of what information remains as potentially relevant features for the classification and recognition tasks.

The second subtask is the classification of known human activities as defined by the task given to this predictor, which means there is a subset of human activities that are task-relevant and use a specific semantic class labeling system.
Class labels may differ between datasets and the specific HAR task.
The original Kinetics dataset's evolution from 400 to 700 classes is an example of not only more classes being available over time that are deemed relevant, but also a change in the actual class structure.
For example, some samples are labeled as ``brushing hair'' in Kinetics-400 were given the new label ``combing hair'' in Kinetics-700 to differentiate between using a brush versus a comb.

The third substask is handling novelty, which is referred to in Figure~\ref{fig:owhar} as novelty recognition.
Novelty recognition includes novelty detection, where being able to detect novelty facilitates the incremental learning of new information, as it is collected over time.
In Figure~\ref{fig:owhar}, these subtasks are ordered by sequence of typical performance, especially when focused on novel activities.
Performing novelty detection and OWL in the feature representation space less specific to a task is beneficial because it provides more information about the input, however this includes task-irrelevant information and as such nuisance novelties may trigger irrelevant novelty detections and possibly affect novelty recognition performance.
When the feature representation learning begins to approach closer to the classification task, current task-irrelevant information is naturally lost where task-relevancy is determined by the $K$ known classes.
Performing novelty detection or recognition within a feature representation space that is closer to the task may result in ignoring novel information, including future task-relevant information.

As noted in Figure~\ref{fig:owhar}, these three subtasks may be completed separately by different processing modules or together by one or two modules.
The important underlying design is that these subtasks share information with one another and tend to be performed sequentially as ordered here, especially when the novelty of interest is in the class activities.

As part of updating the predictor's internal state, external feedback may be leveraged,
 when possible, that is either provided from human annotators as in active learning, or from other predictors.
Feedback from human annotators could be labels of the detected unknown classes, labels of notably incorrect samples in the known classes, or some other signal of performance informed by human annotators.
Feedback from predictors can be similar or even more nondescript signals of feedback.
To properly use this external feedback is a subtask in itself as sometimes the feedback may be in a form that does not directly relate to internal mappings of classes or may even contain noise.
The partial feedback of input samples results in semi-supervised updating of the predictor's internal state.

\subsubsection{Incremental Learning with Prior Knowledge and Novelty}
To enable informed comparison of predictors on a KOWL experiment, we relate the six properties of incremental learning algorithms from \cite{belouadah_comprehensive_2021} to the terminology we've used throughout this paper. 
If we have a prior introduced term for it, then their term is in parentheses.
Otherwise, we use the original names and quote their definitions.
We also add a seventh property, ``prior knowledge,'' as it relates to feature representation learning and is fundamentally an important part to differentiate types of predictors on this benchmark and to determine what is known or novel.

\begin{enumerate}[noitemsep,nolistsep]
    \item \textbf{Complexity}:
        ``[The] capacity to integrate new information
with a minimal change in terms of the model structure.''
        This refers to the predictor's internal state, specifically what that state is and how it changes, \textit{e.g.}, the predictor's number of parameters and changes in their values.
    \item \textbf{Experience (Memory)}:
        The prior mentioned predictor experience may be limited in memory space.
        When this is the case, performing OWL for HAR may become more difficult and nuanced as the predictor needs to handle maintaining the key information of the HAR classes and any information of use for differentiating novel from known activities classes.
    \item \textbf{Regret (Accuracy)}:
        The predictor's internal assessment of its own performance.
        This is a separate and possibly similar or entirely different calculation than what the evaluator uses to assess the predictor's performance.
    \item \textbf{Timeliness}:
        ``[The] delay needed between the occurrence of new data and their integration in the incremental models.''
        In the case of the KOWL protocol, the time for the predictor to perform its operations, such as updating its state and performing inference.
    \item \textbf{Plasticity}:
        ``[The] capacity to deal with new classes that are significantly different from the ones that were learned in the past.''
        In the case of KOWL experiments, the predictor may require feedback or may be able to learn in a semi-supervised manner.
    \item \textbf{Scalability}:
        ``[The] aptitude to learn a large number of classes, typically up to tens of thousands, and ensure usability in complex real-world applications.''
        KOWL-718 has 718 activity classes and the rate of novel classes per increment is rather consistent as seen in Table~\ref{tab:kowl718}.
    \item \textbf{Prior Knowledge}:
        A predictor's prior knowledge must be \textit{explicitly stated} to know how to properly evaluate it and compare it to other predictors.
        Prior knowledge determines what may be considered novel to the predictor.
        If any of the new activities introduced in Kinetics 600 or 700 are known prior, then they are \textit{not} novel to the predictor.
        \textbf{Prior knowledge that shares any information with the dataset \textit{invalidates} the experiment for assessing how a predictor performs and learns the task when that information is supposed to be \textit{novel} to the predictor once observed in the experiment}, which thus invalidates the original purpose of this OWL protocol. 
\end{enumerate}

\label{sec:prior_know_novelty}
        There are predictors \textit{with no prior knowledge}, \textit{i.e.}, those without information shared with the experiment data.
        For example, predictors without any prior training on data external to this benchmark or without any information about the classes considered novel to the evaluator, such as how the predictor performed on prior runs of KOWL experiments with the same data or similarly informed Bayesian priors.
        Before the predictor has seen the first sample in the initial step's training split, the predictor should essentially know nothing.
        What prior knowledge is irrelevant to the task is difficult to know, even more so to know what degree of relevance, which is why none is preferred.
        No prior knowledge 
        is favorable to the scientific method here as it preserves the dependency on order in which the videos are encountered and preserves the closed system for controlled evaluation of a predictor's learning on this benchmark data under this protocol.
        A controlled experiment with appropriate control of when different information is introduced ensures we can more appropriately assess causality, and thus assess if the task-relevant information was learned from within the experiment, rather than if it were known prior to experimentation. 

        A predictor \textit{with prior knowledge} may still be evaluated on a KOWL experiment, but how such an informed predictor behaves towards novelty \textit{cannot} be evaluated unless exactly what is prior known is \textit{not} the novel information being added incrementally throughout the experiment.
        This is difficult to do properly, and so prior knowledge is discouraged if you desire to assess predictor performance in the presence of novelty. 
        We \textbf{strongly recommend} comparing predictors with prior knowledge against those with similar, ideally the same, prior knowledge.
        For example, if there are pretrainings being used, either use the same pretraining feature representations or be trained on the same external data before starting a KOWL experiment.
        Otherwise, if the point is still to assess the predictor's learning of the HAR task within the KOWL experiment, special care needs taken to demonstrate that the prior knowledge is irrelevant to learning the HAR task within the KOWL experiment.
        If one only cares if their predictor can perform well on the benchmark regardless of  informative priors, for example to answer the question of does the prior knowledge share information with the benchmark's task, then of course they are free to do so, but it must be acknowledged that this is not assessing the predictor's learning of the task as represented by the KOWL experiment's data alone.

\section{Experimental Demonstration}
\label{sec:exps}
The following experiments demonstrate the OWL protocol as applied to the Kinetics dataset series to form the KOWL-718 benchmark.
Multiple feedback budgets were examined, where 0\%, 50\%, and 100\% feedback as ground truth labels of the incremental training split were provided to the predictor.
Alongside the protocol evaluation,  
    we conducted an ablation study of the predictors' performances given Kinetics-AVA annotations.
We also examined the state-of-the-art feature representation tolerance to visual transforms to the Kinetics-400 videos.

\subsection{HAR Predictors for Evaluation}
\label{sec:models}

There were two feature representation models explored in this work: X3D~\cite{feichtenhofer_x3d_2020} and TimeSformer~\cite{bertasius_is_2021}.
X3D is a 3D ResNet~\shortcite{he_deep_2016} and TimeSformer is a video transformer~\shortcite{dosovitskiy_image_2022}.
Specific implementation details  for these models can be found in Appendix~\ref{sm:sec:frepr}.
Both models are initialized at their Kinetics-400 pretraining state and the layer just before the softmax readout is used as the feature representation encoding.
Starting with Kinetics-400 pretrainings ensures no information about the future increments is known early, preserving the controlled incremental learning experiment for novelty.
We note that X3D and TimeSformer pretrained on Kinetics-400 were trained in a supervised fashion meaning that their feature representation more closely approaches the classification task, containing more task-relevant information as defined by the known classes, but may lack future task-relevant information in the videos yet to be deemed task-relevant at the initial increment.

The fine-tuned classifier that serves as our demonstrative baseline for this OWL HAR benchmark is a single fully connected layer to the feature representation that then outputs to the final softmax classification layer, denoted as ``ANN''.
The fine-tuned classifier is the predictor's only state updated during the incremental learning.
The ANN classifier always has an unknown class in its softmax classifier and as novel classes become known, the classifier's output increases in size.
Novelty detection is handled using a threshold over the classifier's softmax output where the initial increment's validation partition is used to set that threshold with an accepted level of error, which was 10\% in these experiments.
The accepted level of error represents a prior expectation of how well the validation data is believed to represent the known classes, and was used only to ensure some novelty detection occurred for demonstration purposes.
The ANNs have no method of handling novelty recognition, only detection.

Given the ANNs cannot handle novelty recognition, a semi-supervised classifier was used to depict novelty recognition by learning the potential unknown classes.
This classifier, denoted ``GMM FINCH'', consisted of a Gaussian Mixture Model (GMM) on the feature representation space where the Gaussian components were determined by the FINCH~\shortcite{sarfraz_efficient_2019} clustering algorithm performed on each class's sample sets.
If the known labels are available, then those samples form different groups of points by class in which FINCH calculated the clusters within each separately.
In the hope of better learning the feature representation space's density for finding unknown classes, a Gaussian distribution was fit over a class's clusters to form the class's GMM.
Each Gaussian distribution was fit using the 
the sample mean and covariance in feature representation space.
The unknown samples 
served as their own unknown class, which was fit with a GMM in a similar fashion.
The unknown samples were based on the classifier's predictions after being fit after thresholding on the logarithmic probabilities.
The threshold was found in a similar fashion to the ANNs using the validation data as all knowns and a hyperparameter to establish an accepted amount of error on knowns in hope of enabling better novelty detection in the future.
A sample is deemed novel to the known classes by this threshold if the sample has a logarithmic probability less than the minimum maximum-likely class logarithmic probability minus the likelihood adjustment found from the validation data given the accepted error.

The Extreme Value Machine (EVM)~\cite{rudd_extreme_2018}, which is an open world-specific classifier, was also explored.
However, its performance was found to be worse than the ANN on the initial increment and in addition to its memory constraints and runtime issues, it was left out of further incremental learning analysis.
The EVM in itself is only a novelty detector and cannot perform novelty recognition on its own.
See Appendix~\ref{sm:sec:finetunes} for details.

The demonstrative predictor baselines' properties are: 
\begin{enumerate}[noitemsep] 
    \item \textbf{Complexity}: 
        All baseline predictors used fixed feature representations of either X3D or TimeSformer on Kinetics-400. 
        Finetuned ANNs only adjusted their weights when given feedback and they involve only one hidden layer.
        GMM FINCH was updated in a semi-supervised fashion when given unseen data samples, with or without evaluator feedback.
    \item \textbf{Experience}:
        All predictors examined have an unlimited memory buffer in the demonstration examples.
        However, this could be an experimental parameter changed when analyzing different predictors on KOWL-718.
    \item \textbf{Regret}:
        The Kinetics-400 pretrained feature representations used the losses respective to their model, X3D or TimeSformer.
        Cross-entropy was used to train the finetuned ANNs.
        GMM FINCH used the logarithmic probability density function for each class's GMM.
    \item \textbf{Timeliness}:
        No time-limits were placed on a predictor's inference or fitting operations.
    \item \textbf{Plasticity}:
        The baseline finetuned ANNs can only detect unknowns.
        They do not support recognizing unknown classes and must be given feedback labels to account for new activity classes over time.
        The GMM FINCH predictors learn a GMM in the fixed feature space per activity class and as such may identify unknown classes without feedback as Gaussian distributions within the feature space.
        The GMM FINCH predictors thus update whenever any new data is presented to it, even without feedback.
    \item \textbf{Scalability}:
        All predictors were fit on 407 activity classes in the initial increment and then incrementally encounter the remaining novel 311 classes, totaling 718 classes.
        The new classes per increment is seen in Table~\ref{tab:kowl718}.
        No further assessment of scalability is demonstrated.
        See Appendix~\ref{sm:sec:agents} for implementation details.
    \item \textbf{Prior Knowledge}:
        No external prior knowledge was used.
        Pretrainings of X3D and TimeSformer on Kinetics-400 were used only to expedite the experimental process by skipping their fitting on the initial increment in KOWL-718, which is the same input data for each model.
\end{enumerate}

Task performance for Kinetics is often measured using classification accuracy of top-1, top-5, or $1 - \frac{\text{top-1} + \text{top-5}}{2}$.
Given that our unified labeling is inherently imbalanced, we examine not only top-1 accuracy, but also MCC and the NMI as they are measures that better capture performance.

\begin{figure*}
    \centering
    \small
    \textbf{Raw HAR Non-Cumulative Performance on KOWL-718}
    \includegraphics[width=\linewidth]{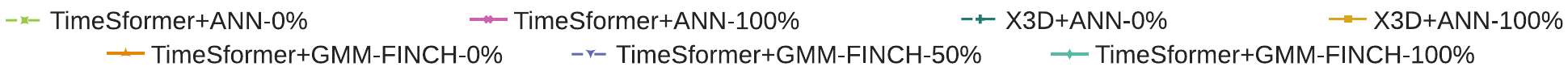}
    \subcaptionbox{
        Training Set
        \label{fig:raw:train}
    }[.495\linewidth]{
        \includegraphics[width=\linewidth]{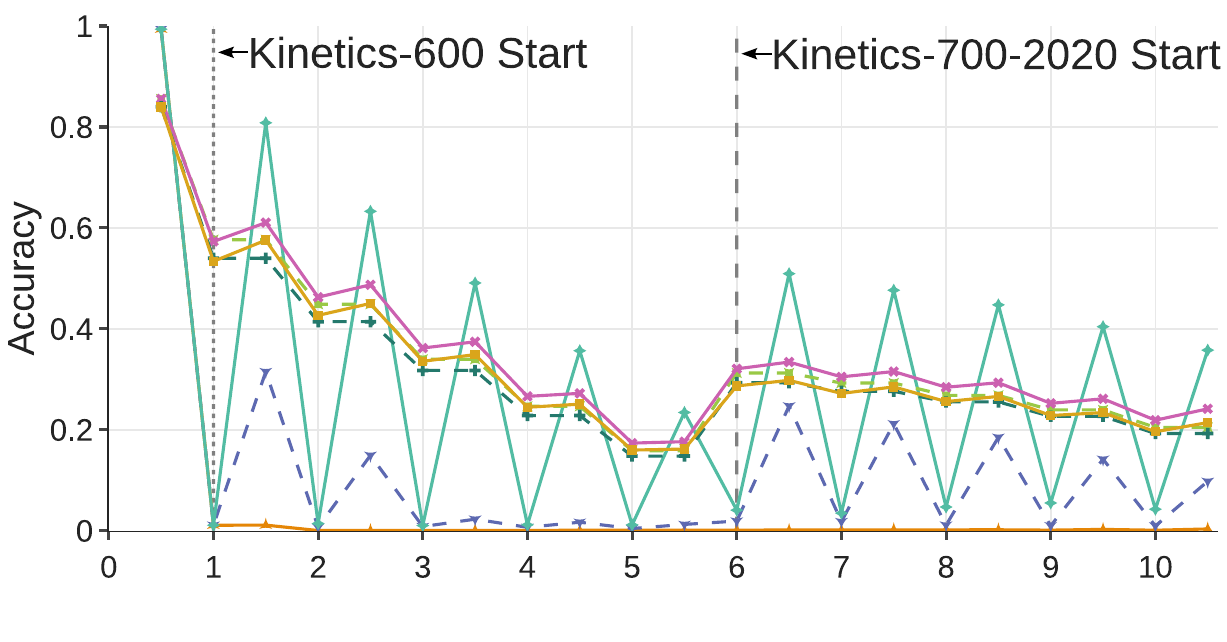}
        \includegraphics[width=\linewidth]{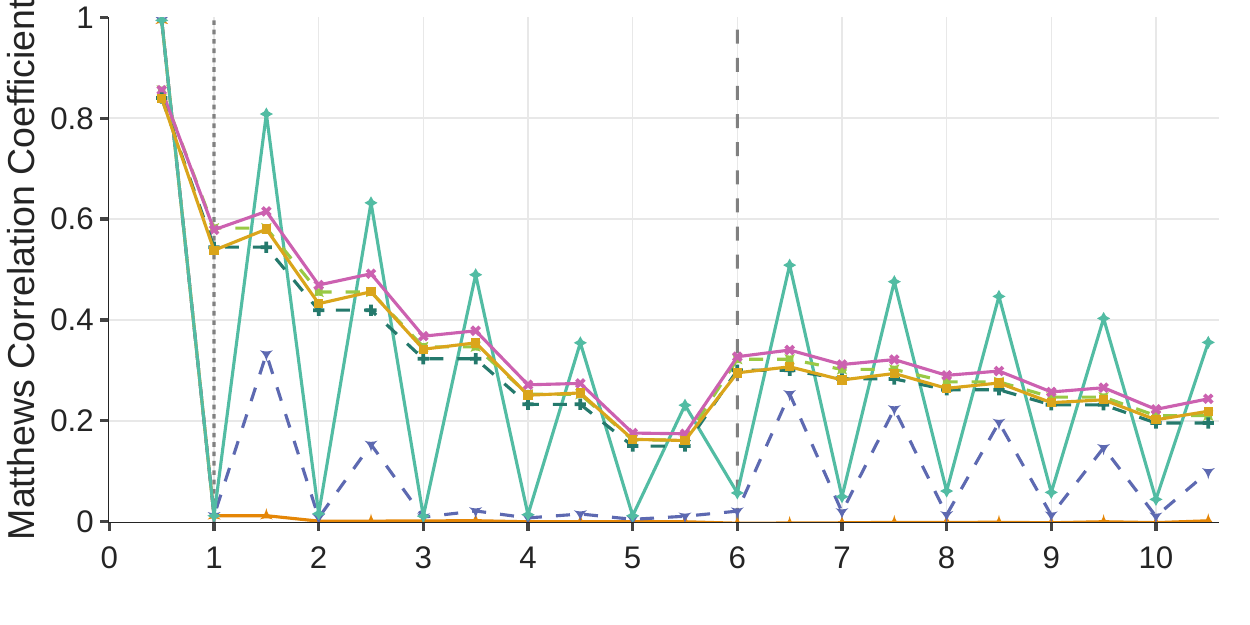}
        \includegraphics[width=\linewidth]{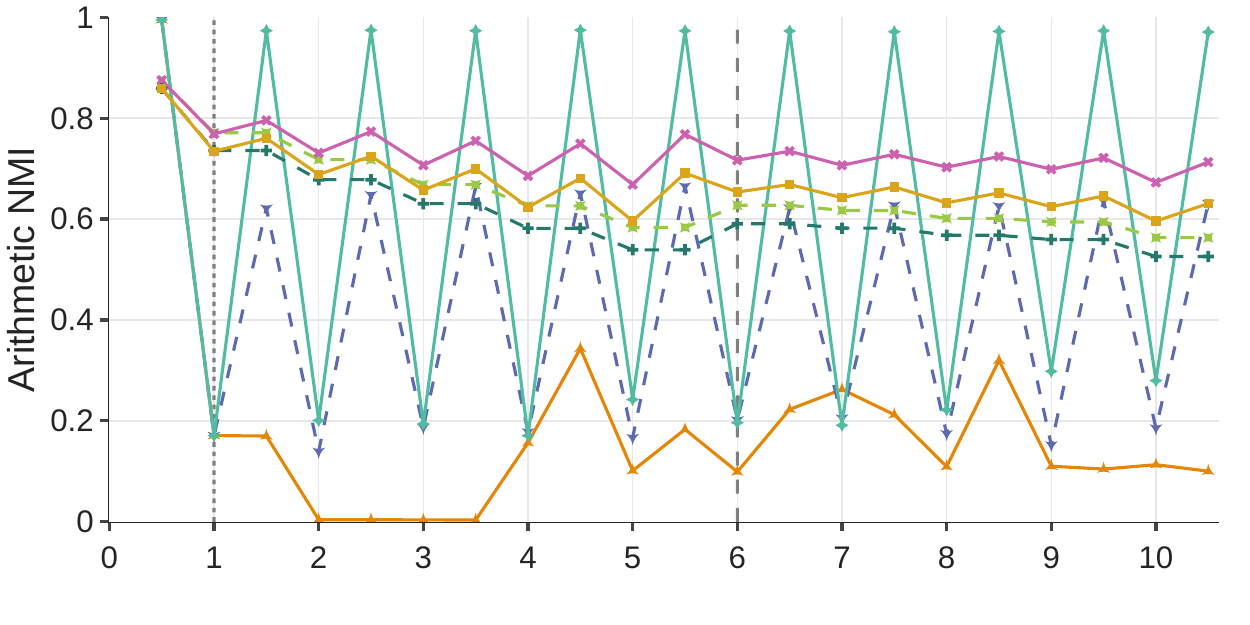}
    }
    \subcaptionbox{
        Testing Set
        \label{fig:raw:test}
    }[.495\linewidth]{
        \includegraphics[width=\linewidth]{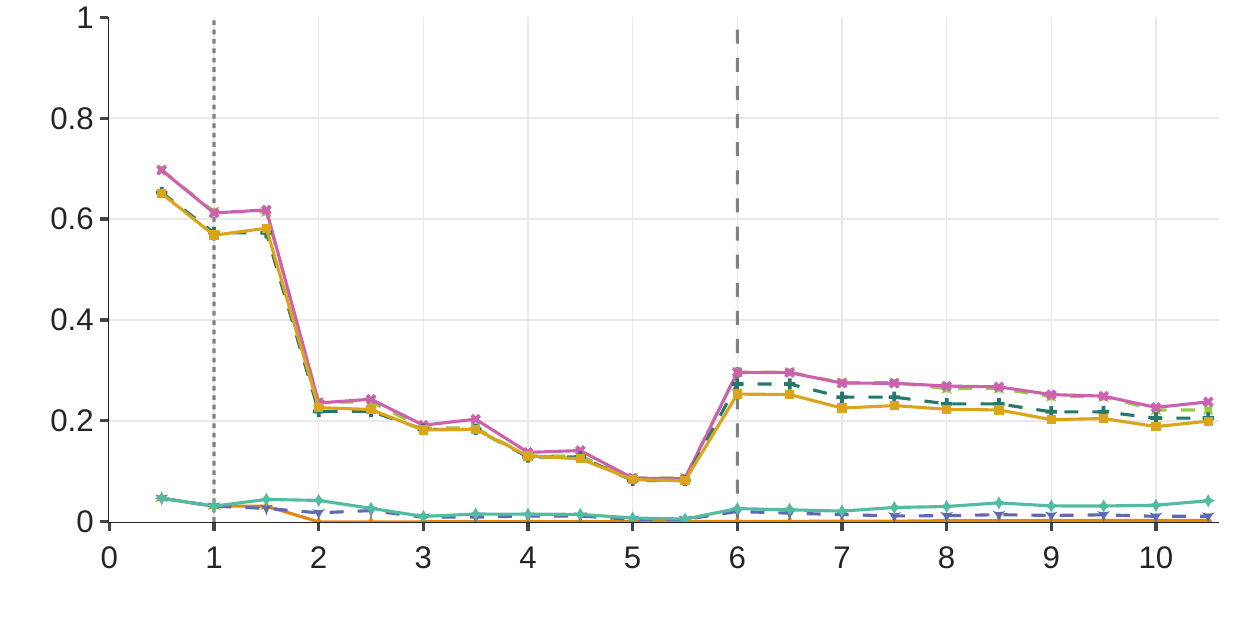}
        \includegraphics[width=\linewidth]{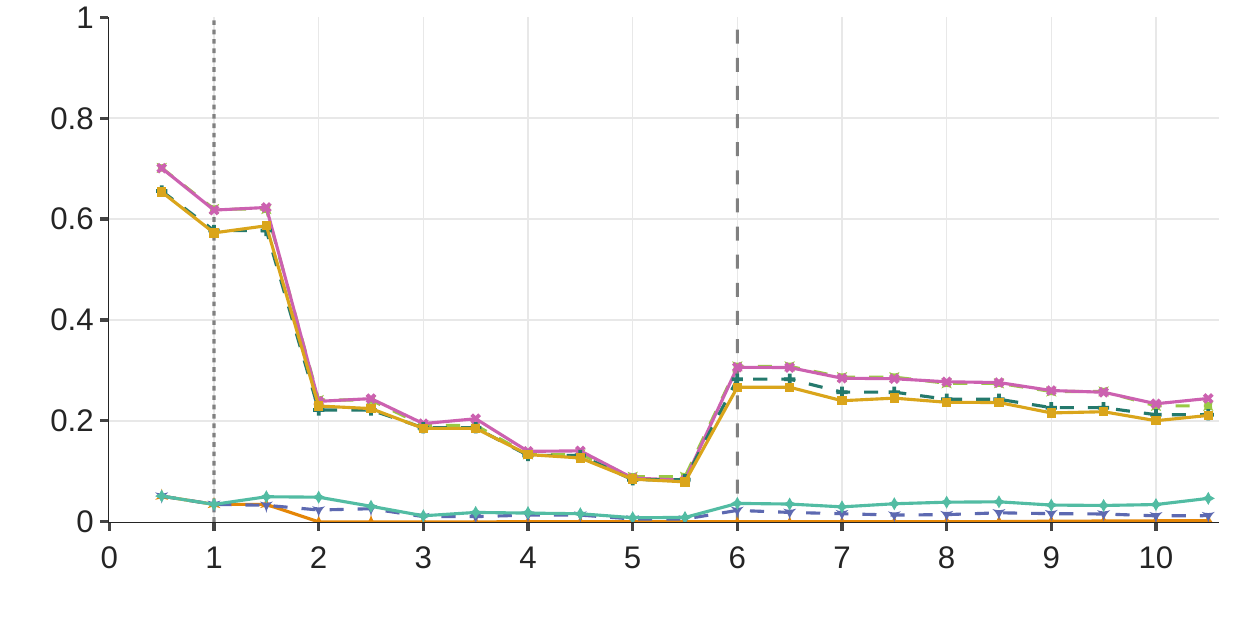}
        \includegraphics[width=\linewidth]{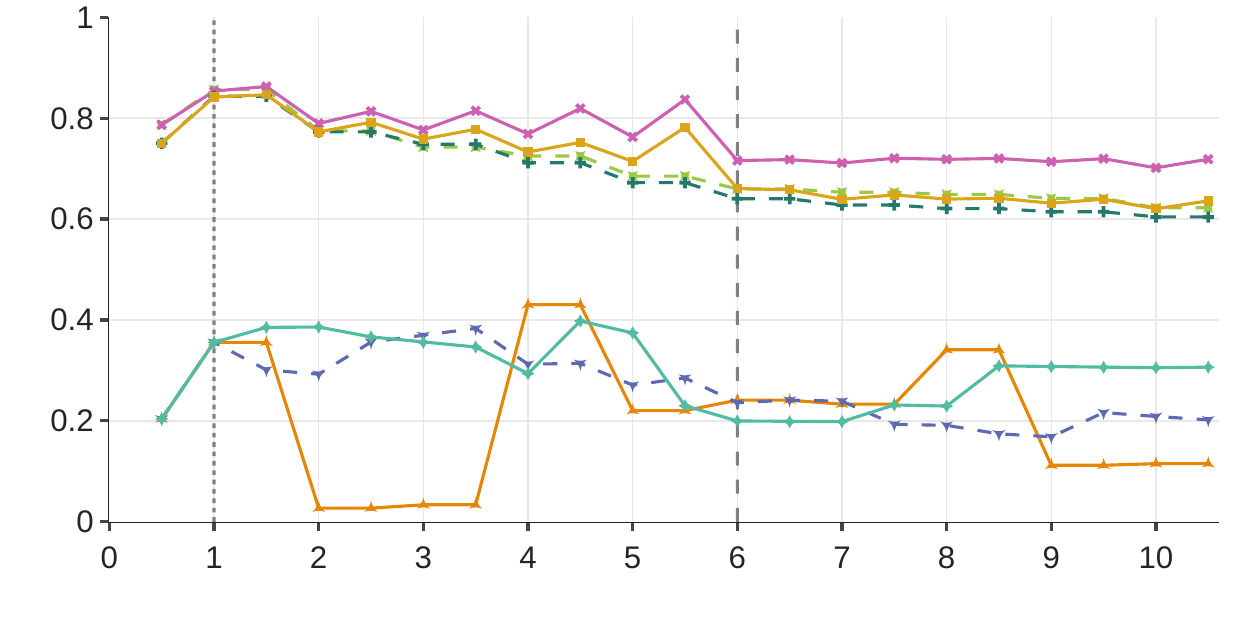}
    }
    \caption{
        \small
        The 
        \textbf{non-cumulative} incremental HAR \textbf{raw confusion matrix} performance of the baseline predictors
            where the incremental evaluation for
            \textbf{pre-feedback is on whole numbers} 
            and 
            \textbf{post-feedback is on half steps}.
        All pre-feedback samples are out-of-sample, regardless of whether they are in the future training data. 
        The raw confusion matrix compares the top-1 predictions to the target label without any reductions of knowns or unknowns.
        The \textbf{predictors were fit on all prior and current experienced data} at each increment,
        \textit{except} that the ANNs did not use prior seen validation / test sets as unlabeled data as the GMM FINCH recognizer did.
        \vspace{-1em}
    }
    \label{fig:raw}
\end{figure*}

The predictors require different learning paradigms depending on their feedback budget.
100\% human label feedback is the typical supervised incremental learning case where all ground truth labels are given to the predictor after an initial performance assessment at each increment.
0\% feedback is when no feedback of any form is given to the predictor and its performance is examined based how well it detects novel classes as unknown, and, if the predictor learns to recognize unknown classes over time, how its class-cluster mappings correlate to the unknown ground truth.
The ANN predictor baselines with 0\% feedback do not update under the complete supervised learning paradigm.
However the GMM FINCH predictors continue to update in a semi-supervised manner without any further feedback labels.
50\% feedback is the most nuanced as the predictor must prioritize the increment's samples by how informative the predictor believes feedback to be for those samples.
Given the small spread of raw performance difference between the 100\% and 0\% feedback ANNs  (see Figure~\ref{fig:raw}), the performance figures depict only GMM FINCH performing 50\% feedback for the simplicity of demonstration and for the figures' readability.

\subsection{Open World HAR: Novel Activities Over Time}
\label{sec:kowl_exp}
\begin{figure*}[t]
    \centering
    \small
    \textbf{HAR Novel-to-Predictor Classification Task Non-Cumulative Performance on KOWL-718}
    \includegraphics[width=\linewidth]{graphics/exp2/legend_usage.pdf}
    \subcaptionbox{
        Training Set
        \label{fig:clsf:train}
    }[.495\linewidth]{
        \includegraphics[width=\linewidth]{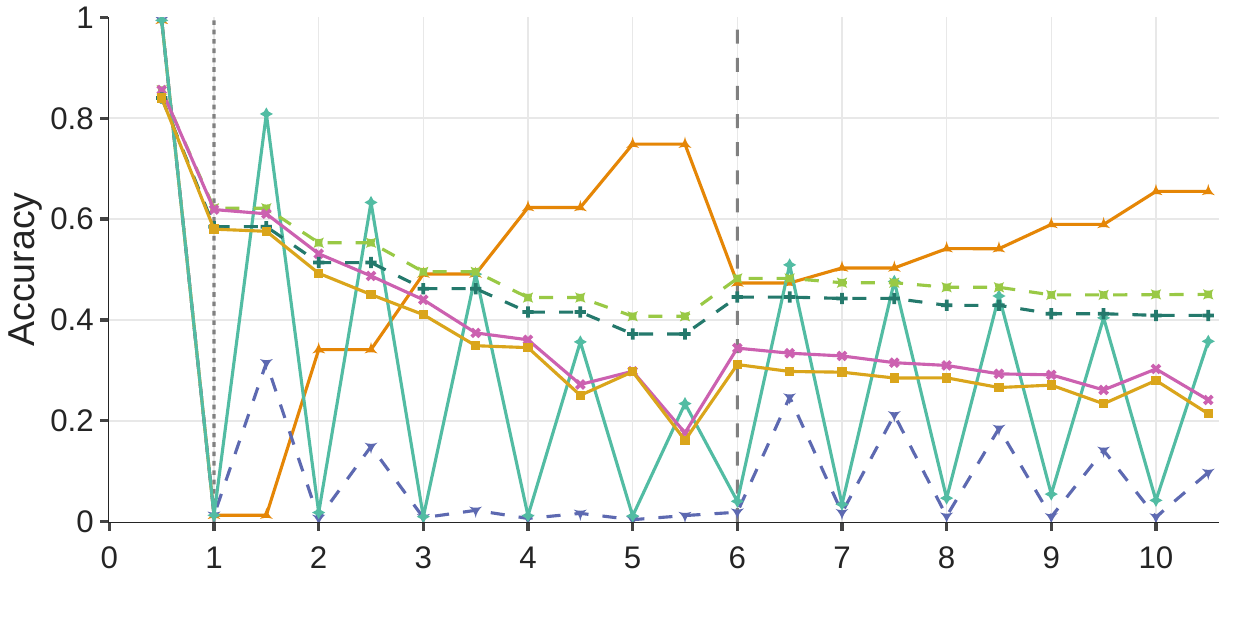}
        \includegraphics[width=\linewidth]{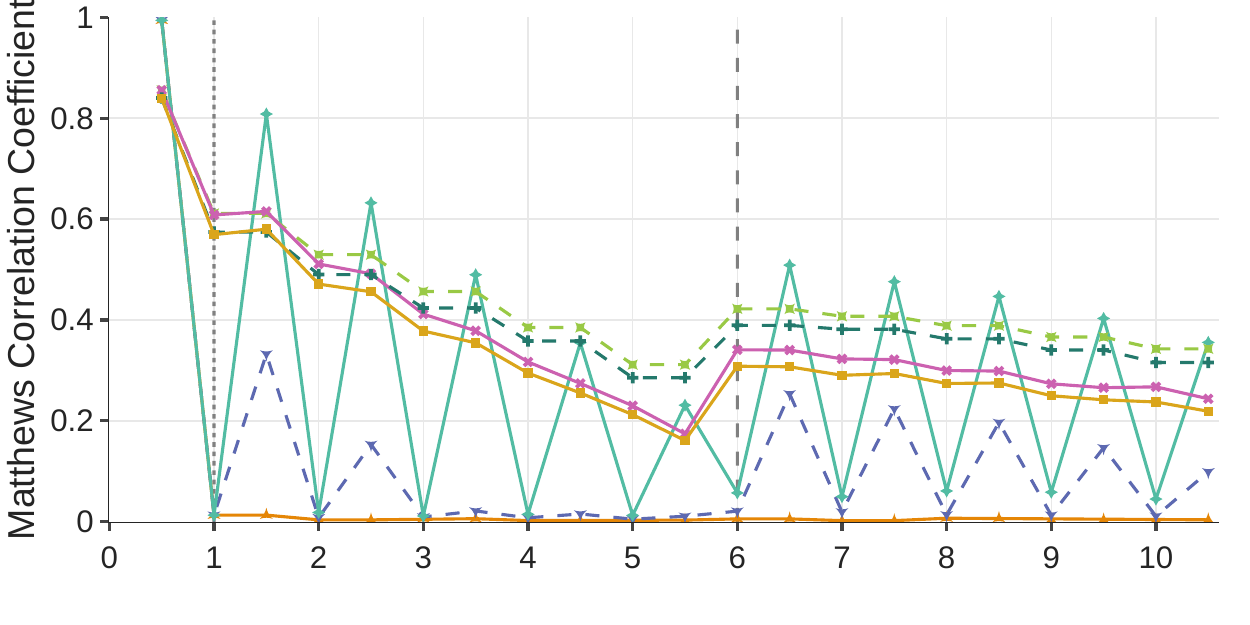}
        \includegraphics[width=\linewidth]{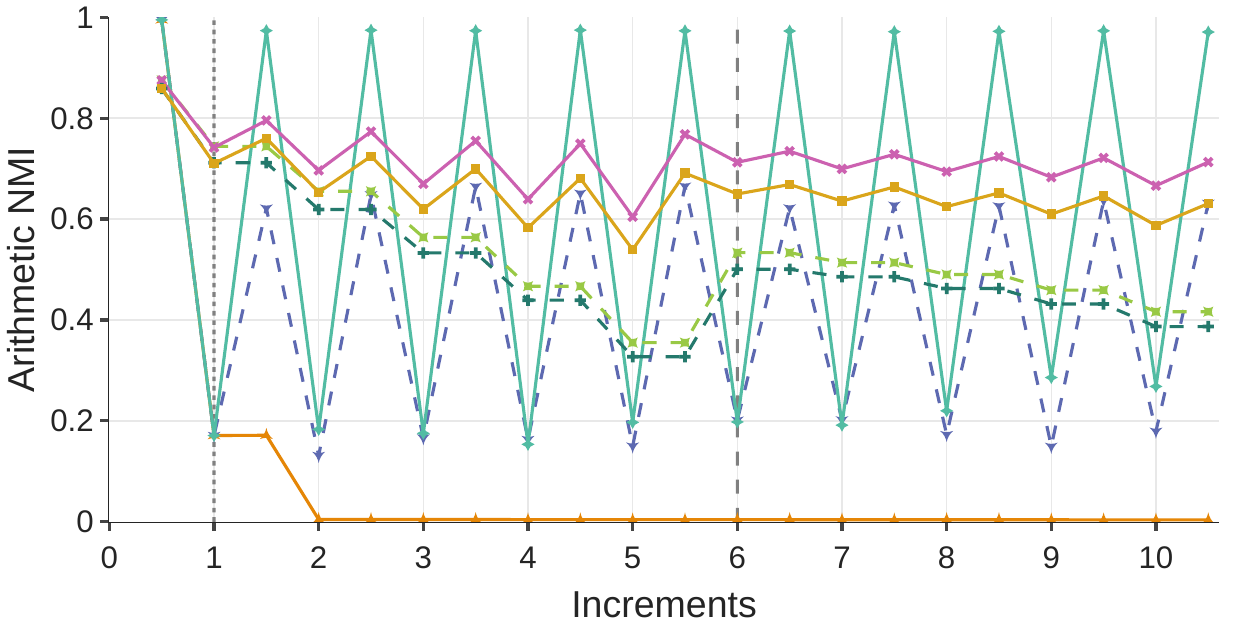}
    }
    \subcaptionbox{
        Testing Set
        \label{fig:clsf:test}
    }[.495\linewidth]{
        \includegraphics[width=\linewidth]{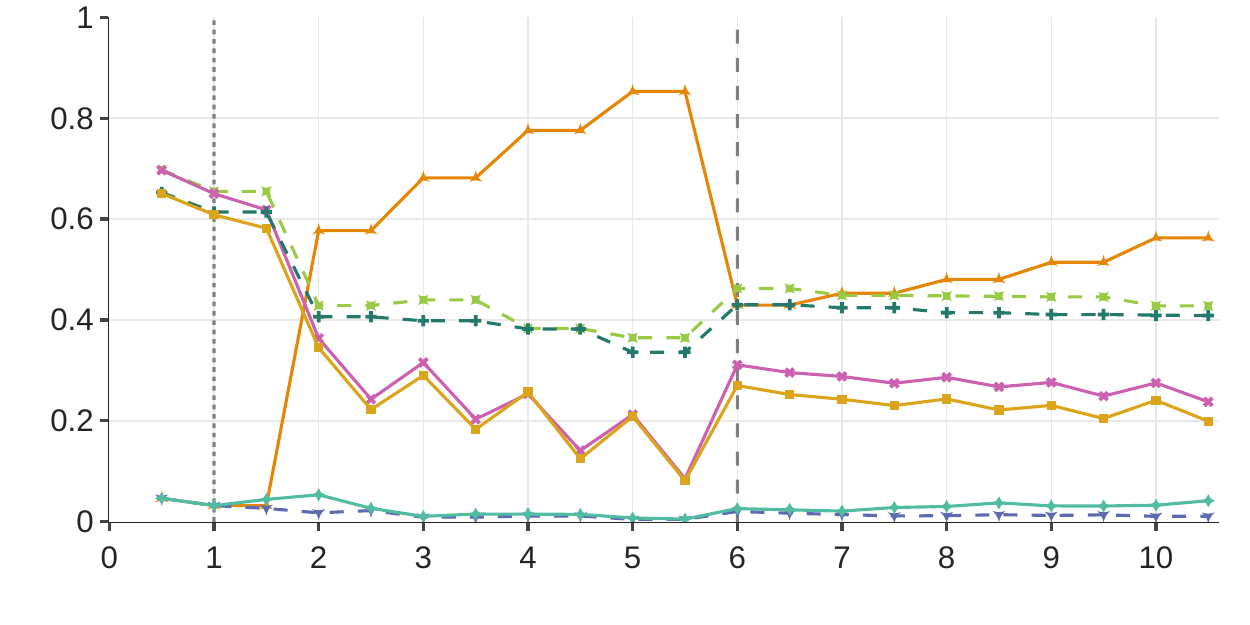}
        \includegraphics[width=\linewidth]{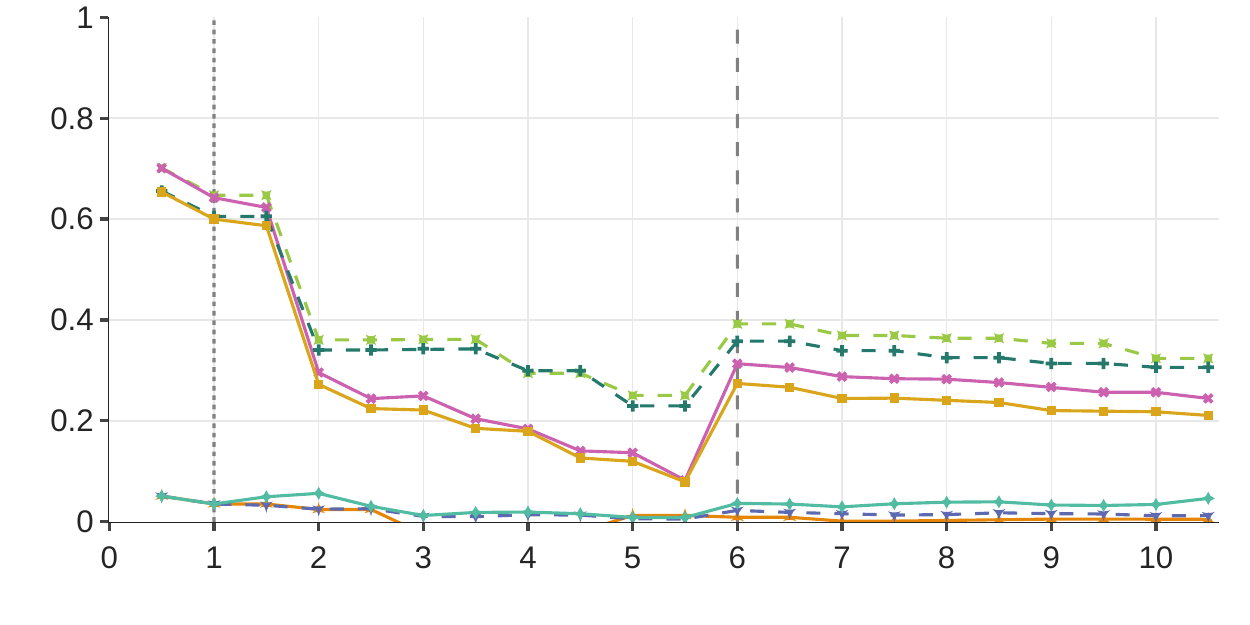}
        \includegraphics[width=\linewidth]{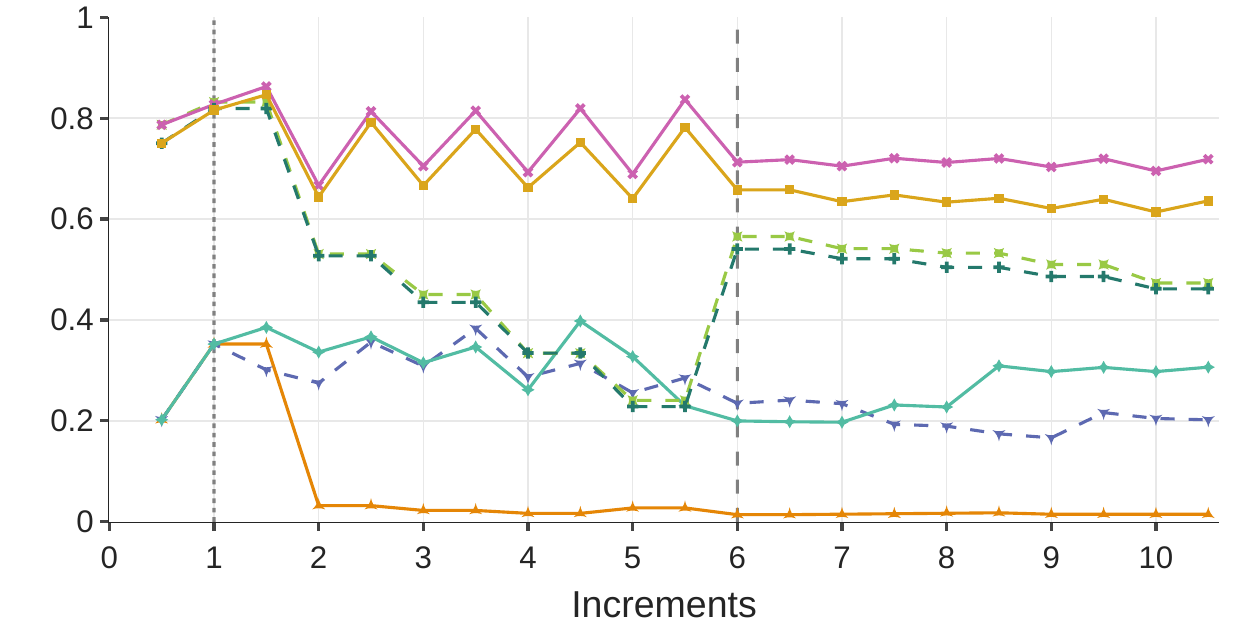}
    }
    \caption{
        \small
        The 
        \textbf{non-cumulative} incremental HAR \textbf{classification task} performance of the baseline predictors
            where the incremental evaluation for
            \textbf{pre-feedback is on whole numbers} 
            and 
            \textbf{post-feedback is on half steps}.
        All pre-feedback samples are out-of-sample, regardless of whether they are in the future training data. 
        The \textbf{predictors were fit on all prior and current experienced data} at each increment,
        \textit{except} ANNs did not use prior seen validation / test sets as unlabeled data as the GMM FINCH recognizer did.
        TimeSformer is the best feature representation.
        The ANNs with 0\% feedback 
        perform slightly worse than their incremental learning 100\% feedback counterpart.
        \vspace{-1em}
    }
    \label{fig:clsf}
\end{figure*}

\begin{figure*}[t]
    \centering
    \small
    
    \textbf{HAR Novel-to-Predictor Novelty Detection Non-Cumulative Performance on KOWL-718}
    \includegraphics[width=\linewidth]{graphics/exp2/legend_usage.pdf}
    
    \subcaptionbox{
        Training Set
        \label{fig:detect_train}
    }[.495\linewidth]{
        \includegraphics[width=\linewidth]{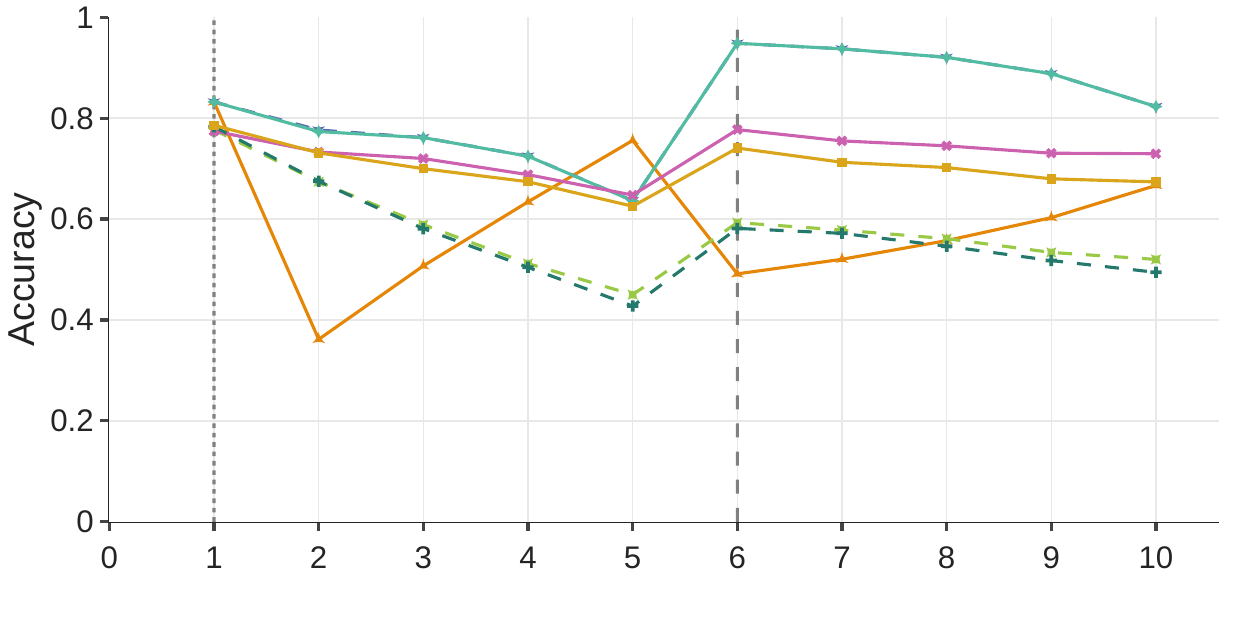}
        \includegraphics[width=\linewidth]{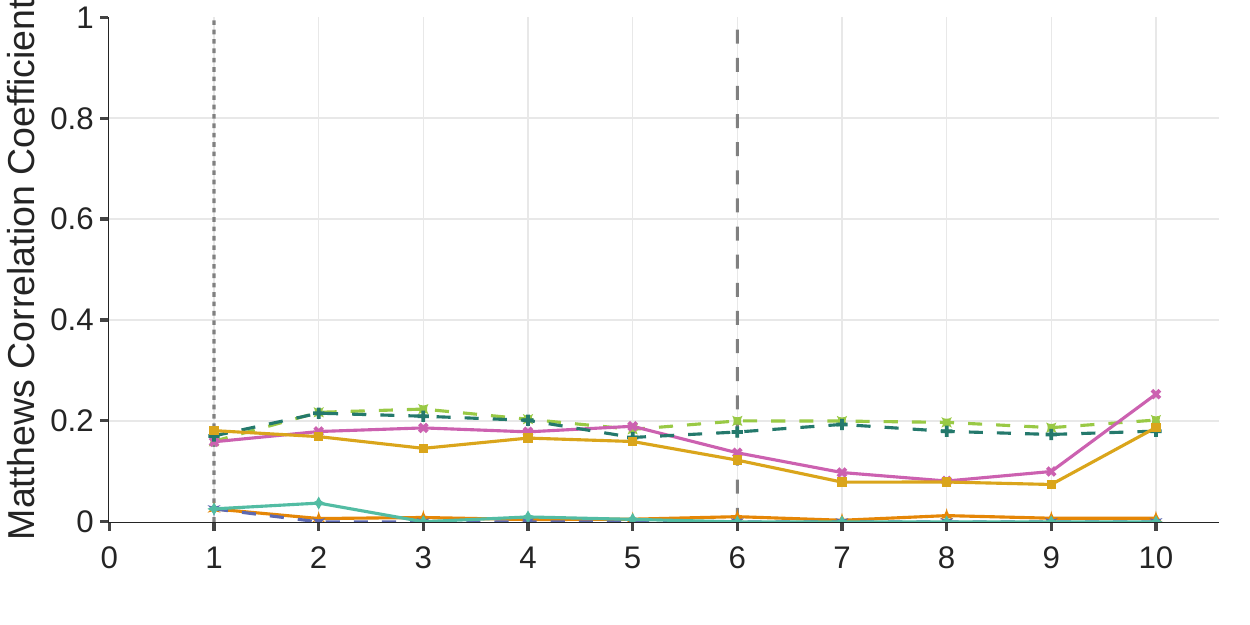}
        \includegraphics[width=\linewidth]{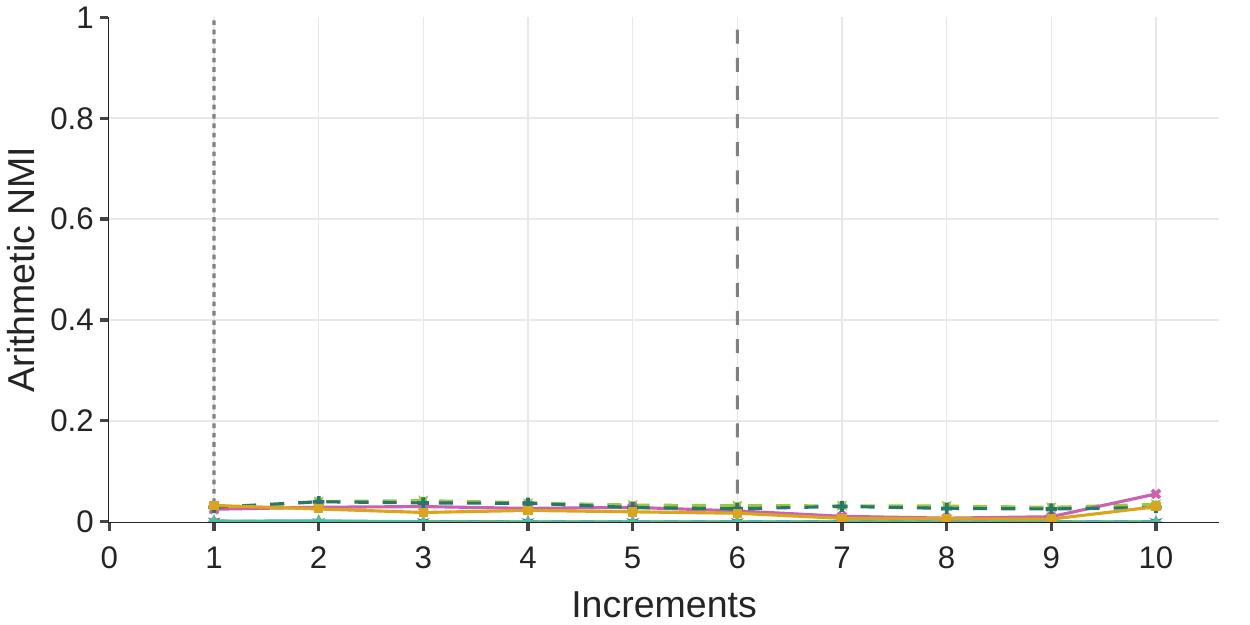}
    }
    \subcaptionbox{
        Testing Set
        \label{fig:detect_test}
    }[.495\linewidth]{
        \includegraphics[width=\linewidth]{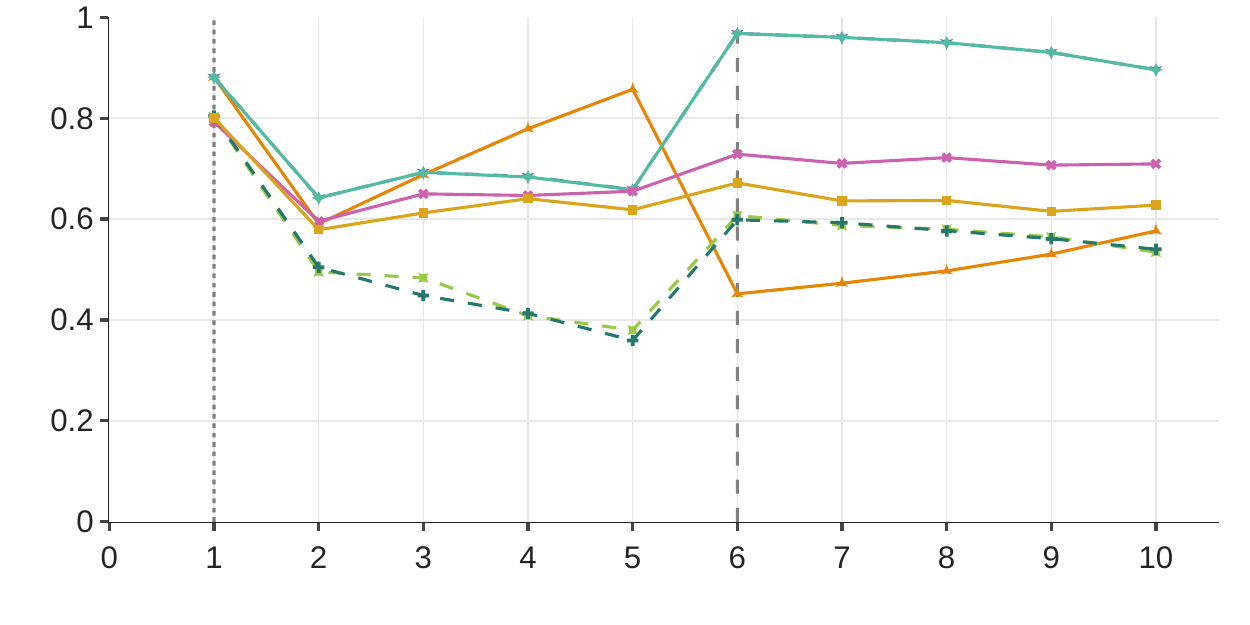}
        \includegraphics[width=\linewidth]{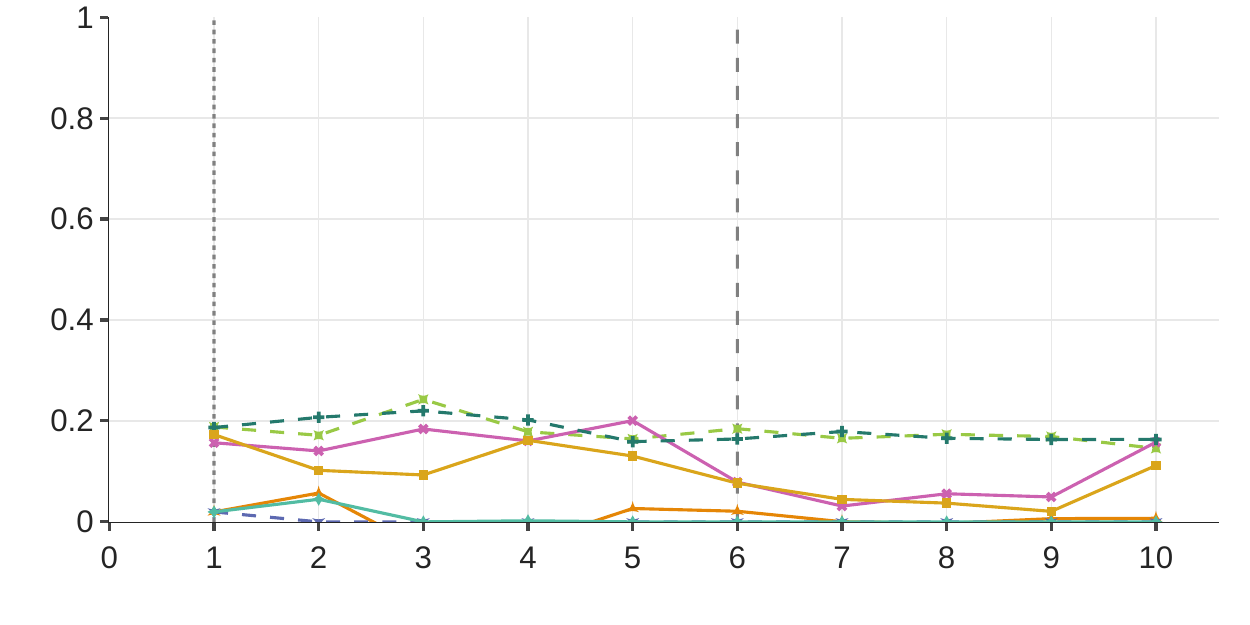}
        \includegraphics[width=\linewidth]{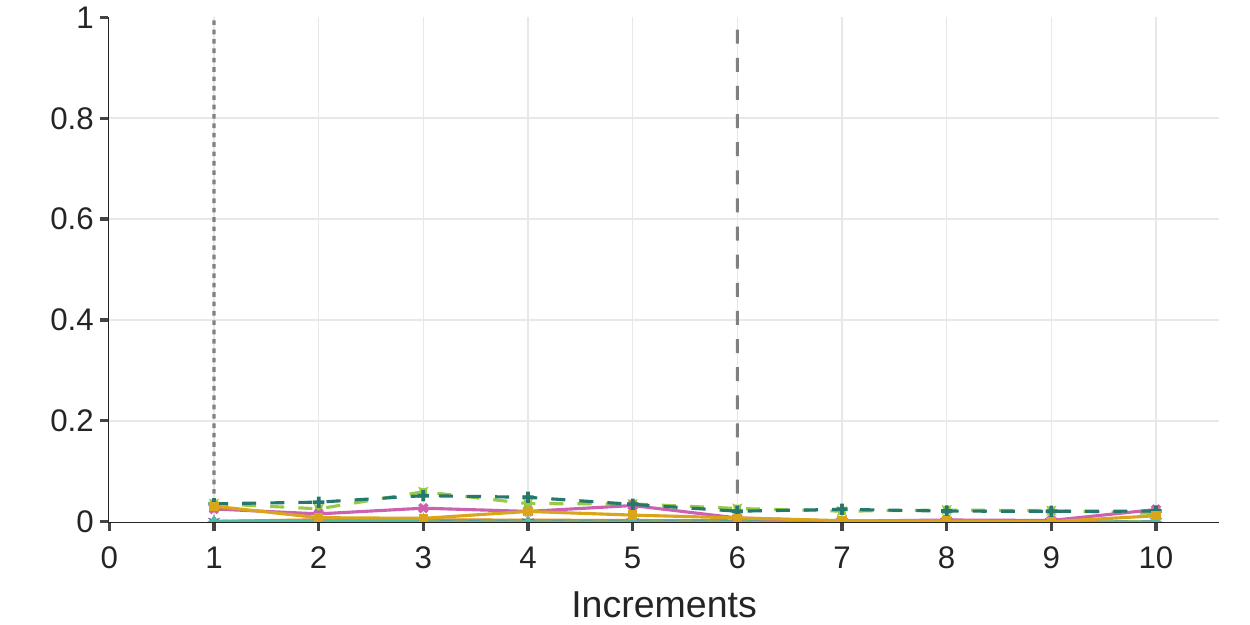}
    }
    \caption{
        \small
        The predictors' novelty detection performance at each pre-feedback phase during incremental learning on KOWL-718.
        All pre-feedback samples are out-of-sample, regardless of whether they are in the future training data. 
        \vspace{-1em}
    }
    \label{fig:detect}
\end{figure*}

The time-series line plots, such as the figures for the raw performance in Figure~\ref{fig:raw} and the classification task in Figure~\ref{fig:clsf}, are all non-cumulative point-estimate measurements.
This means that every measurement stands for the performance \textit{only} on the new data for that increment where at the whole step the new data serves to evaluate the predictor generalization in the pre-feedback phase and then the half-step evaluates the predictor after being given the feedback requested, if any.
Analyzing the performance in a non-cumulative manner indicates the effect that each increment's samples have at that moment in time to better compare the performance on pre- and post- feedback.
As such, jumps in performance from one increment to another may occur due to the number of samples and changes in class balance by introducing the increment's novel classes.
The figures for assessing how the predictors handle novelty, such as novelty detection Figure~\ref{fig:detect}, novelty recognition Figure~\ref{fig:recog}, novelty reaction time Figure~\ref{fig:react}, all show only the performance at the pre-feedback phase, as no novelty occurs in the post-feedback phase given the video samples have already been seen.

\subsubsection{HAR Classification Task Performance}
\label{sec:clsf}
The HAR classification task performance of the incremental learning predictors was assessed over time as novel classes were introduced in each increment, as seen in Figure~\ref{fig:raw}~and~\ref{fig:clsf}.
Note that this includes the reduction of actual unknown and recognized unknown classes into the ``unknown'' catch-all class.
The performance of all predictors degrades over the increments as expected, given that the feature representation is frozen at Kinetics-400 and thus does not contain information about the new classes added within Kinetics-600 and -700.
This decrease in performance happens even when 100\% feedback is given, indicating that the learned feature representations from X3D and TimeSformer do not generalize to unseen activities given the Kinetics-400 training data to the rest of Kinetics-600 and -700 samples.
Otherwise, the classifiers would be able to perform better on the accuracy and MCC measurements in Figure~\ref{fig:clsf}.
Overall, predictors using the TimeSformer as their feature representation out performed those using X3D.
This increase in performance comes at the cost of longer training and inference times as noted in Appendix~\ref{sm:sec:frepr}.
Given that TimeSformer consistently out-performs X3D, the plots include only the GMM FINCH predictors using the TimeSformer feature representations for clearer visual readability.

The post-feedback NMI for nearly all of the models using 100\% feedback are higher than their relative pre-feedback evaluation measurements with the exception of the NMI of GMM FINCH increment 5.
The finetuned ANNs with 0\% feedback have the same post-feedback performance as their pre-feedback performance because no updates are made to their internal state after the initial increment.
The GMM FINCH with 50\% feedback also experiences an increase in performance in the post-feedback phases, but in the training set either not as significant as the 100\% feedback and in the testing set not always improving relative to the increment's pre-feedback measurements.
Nonetheless, this demonstrates the value of informative feedback in learning a task.

Notably, from both Figure~\ref{fig:clsf} and the raw performance in Figure~\ref{fig:raw}, the GMM FINCH model seems to overfit the training data, as indicated by the low measurements during the pre-feedback phases (whole numbers) and high measurements during post-feedback phases (half-steps) on the training set and the low measurements on the both phases in the testing set.
The 0\% feedback GMM FINCH is an example of when to observe MCC and NMI  because on both the training and testing sets in Figure~\ref{fig:clsf} the accuracy indicates high performance  due to predicting unknowns frequently. But doing so does not correlate linearly or non-linearly with the actual labels, as captured by MCC and the NMI.

The predictors exhibit an interesting phenomena where on post-feedback their accuracy and MCC either remain the same or drop, while the opposite occurs for most of their NMI scores.
This is due to the fact that at post-feedback the novel classes are no longer reduced to unknown if feedback labels for that class were given to the predictor.
The raw measures for these predictors mostly remain similar in performance for post-feedback compared to their pre-feedback, especially the finetuned ANNs, as seen in Figure~\ref{fig:raw}.

Given the reduction of unknowns, the predictors with different feedback amounts have fewer known activities to learn and more samples that fall under ``unknown.''
Due to this, such predictors may have higher performance measurements as seen in the 0\% feedback finetuned ANNs on accuracy and MCC.
However the ANN's NMI indicates they perform worse than their 100\% feedback counterparts.
Furthermore, referencing the raw confusion matrix's measures in Figure~\ref{fig:raw}, these increases in accuracy and MCC vanish, indicating it is due to the reduction of unknowns.
Also, the GMM FINCH 50\% feedback at times exceeds its 100\% feedback counterpart on the testing set.

\subsubsection{HAR Novelty Detection Performance}
\label{sec:detect}

Figure~\ref{fig:detect} depicts the predictor's novelty detection performance during incremental learning.
This is the binary classification task when the known classes to the predictor at each increment are reduced to ``known'' and the actual unknowns and predictor's recognized unknown classes are reduced together to ``unknown.''
We note that the 100\% feedback GMM FINCH's accuracy is nearly identical to the known class frequency shown in Table~\ref{tab:kowl718}.
This means that the GMM FINCH 100\% feedback is predicting ``known'' nearly all the time, which is an ill performing novelty detector and is captured by the low measurements of MCC and NMI.
Based on the MCC and NMI measures, the baseline predictors demonstrate that novelty detection over the incremental learning of KOWL-718 is not an easy task, and also demonstrate how accuracy may be misleading.
The best novelty detectors in Figure~\ref{fig:detect} would be the finetuned ANNs using the threshold determined from the initial increment's validation set.
Although demonstrative, the baseline performance leaves much to be desired and indicates room for improvement from future predictors.

\subsubsection{HAR Novelty Recognition Performance}
\label{sec:recog}

\begin{figure*}[t]
    \centering
    \small
    
    \textbf{HAR Novel-to-Predictor Novelty Recognition Non-Cumulative Performance on KOWL-718}
    \includegraphics[width=\linewidth]{graphics/exp2/legend_usage.pdf}
    
    \subcaptionbox{
        Training Set
        \label{fig:recog_train}
    }[.495\linewidth]{
        \includegraphics[width=\linewidth]{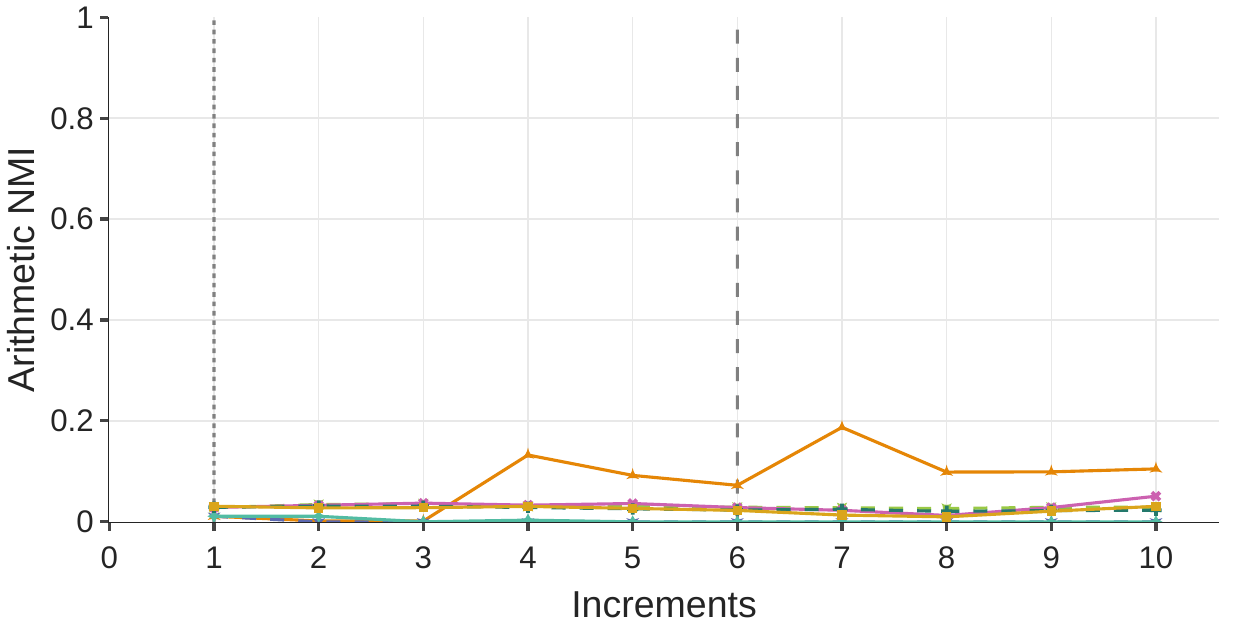}
    }
    \subcaptionbox{
        Testing Set
        \label{fig:recog_test}
    }[.495\linewidth]{
        \includegraphics[width=\linewidth]{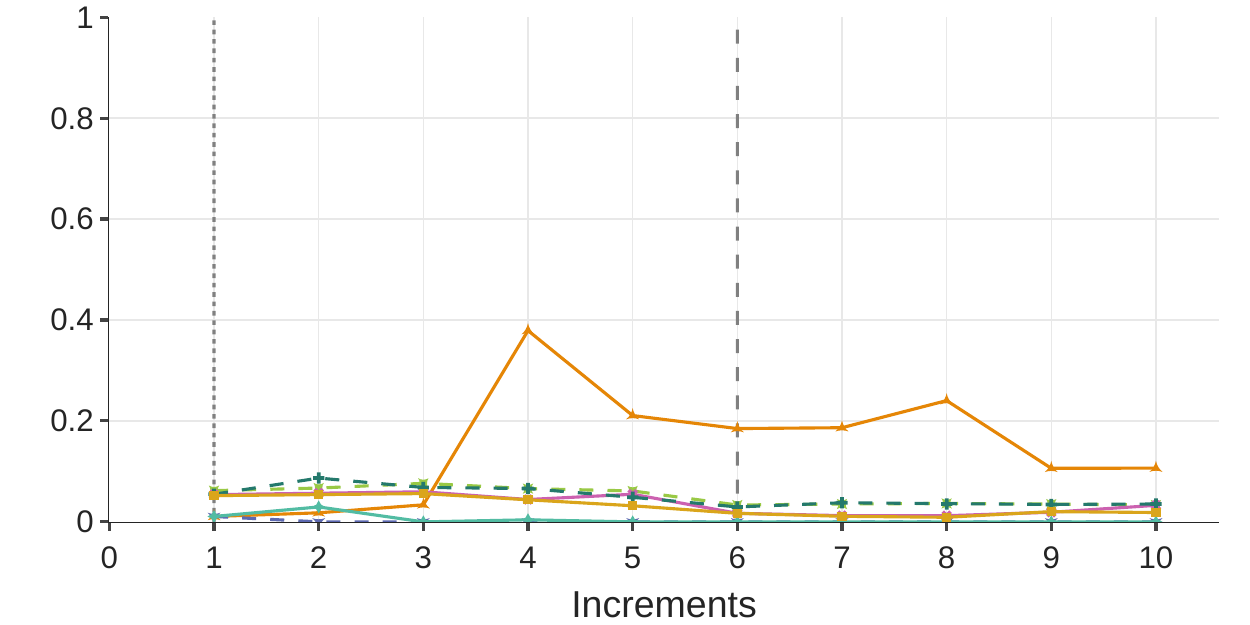}
    }
    
    \caption{
        \small
        The novelty recognition non-cumulative performance of the baseline predictors.
        The evaluation of an increment's novel samples is in the pre-feedback phase only.
        All pre-feedback samples are out-of-sample, regardless of whether they are in future training data. 
        \vspace{-1em}
    }
    \label{fig:recog}
\end{figure*}

Figure~\ref{fig:recog} depicts the non-cumulative predictor performance as they incrementally learn KOWL-718.
Accuracy and MCC are not useful to visualize given that they require the symbolic mappings to be exact mappings, where novelty recognition is inherently a comparison of clustering along with a single catch-all ``known'' class.
At best, accuracy would simply be overly optimistic when predicting samples as known the majority of the time given the balance of knowns to unknowns in KOWL-718, which is representative of real world scenarios.
Given this and the amount of samples per increment, the mutual information normalized by the arithmetic mean of entropies is measured per increment to assess the shared task-relevant information between the actual labels and the predictor's recognized unknown classes.
Perhaps as hinted at by the prior novelty detection performance, the baselines do not perform well on novelty recognition with the majority of the NMI measurements for the predictors falling below $0.1$.
The TimeSformer GMM FINCH with 0\% feedback notably outperforms all the others starting from increment 4 onwards, but does not exceed a measurement of $0.4$ NMI.
This demonstrates the novelty recognition's difficulty as a semi-supervised learning task and we are hopeful that future predictors will surpass these baselines as more researchers take up this line of work.

\subsubsection{HAR Novelty Reaction Time Performance}
\label{sec:react}

\begin{figure*}[t]
    \centering
    \small
    
    \textbf{HAR  Novel-to-Evaluator Novelty Reaction Time Non-Cumulative on KOWL-718}
    \includegraphics[width=\linewidth]{graphics/exp2/legend_usage.pdf}
    
    \subcaptionbox{
        Training Set
        \label{fig:react_train}
    }[.495\linewidth]{
        \includegraphics[width=\linewidth]{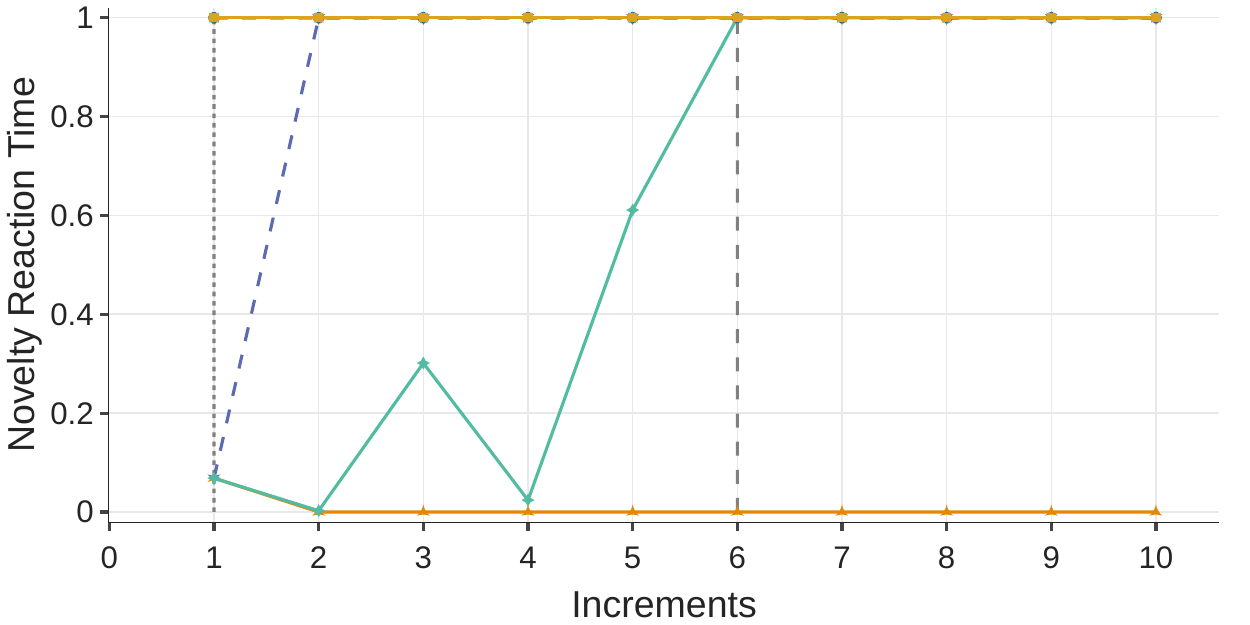}
    }
    \subcaptionbox{
        Testing Set
        \label{fig:react_test}
    }[.495\linewidth]{
        \includegraphics[width=\linewidth]{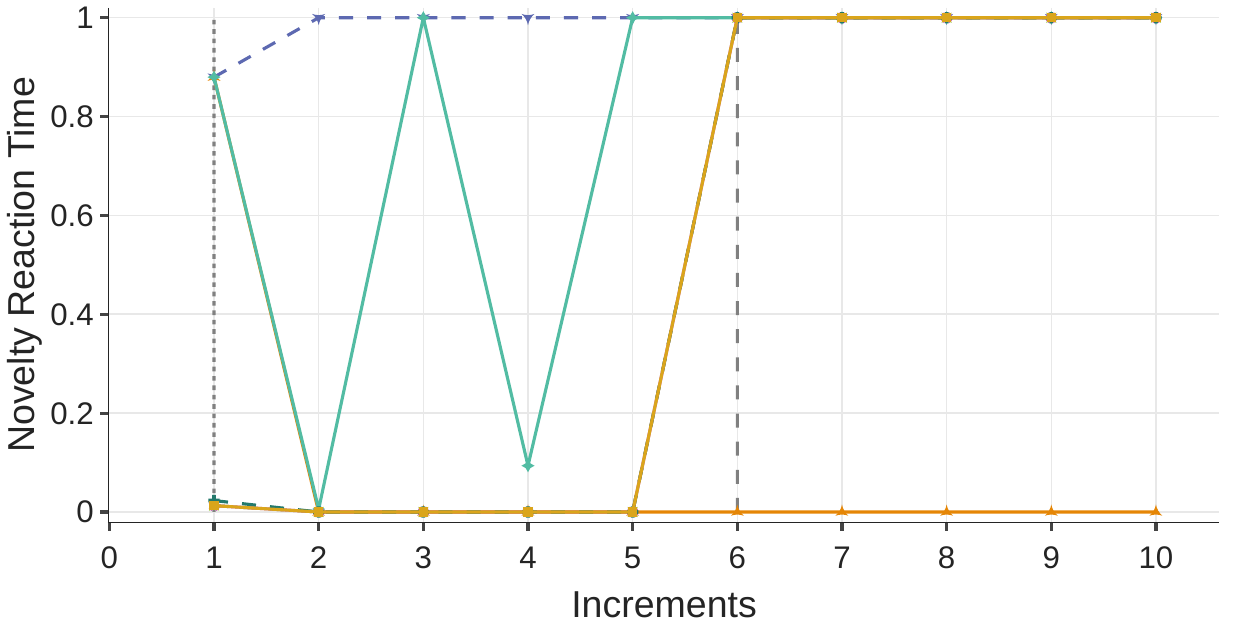}
    }
    
    \caption{
        \small
        The per-increment novelty reaction time measure from Equation~\ref{eq:react} of the baseline predictors.
        \textbf{Lower values mean a quicker detection of novel actions.}
        The first observation of an increment's novel samples is in the pre-feedback phase only.
        All pre-feedback samples are out-of-sample, regardless of whether they are in future training data. 
        TimeSformer+GMM-FINCH-0\% is the best performing on reaction time given its low values, while all of the ANNs perform the worst on the training set.
        TimeSformer+GMM-FINCH on 50\% and 100\% feedback has lower reaction times early on in both sets, but joins the ANNs in maximum reaction time value by the 6th increment when the Kinetics-700 dataset begins.
    }
    \label{fig:react}
\end{figure*}

Figure~\ref{fig:react} depicts the predictors' novelty reaction time on KOWL-718 and indicates that the ANN models with thresholding for unknown samples all react at the same time regardless of their feature representation model.
The ANN models react the slowest on the training set as it is released at the pre-feedback stage but join the GMM-FINCH 0\% model for the first five increments as the fastest reacting predictors to novelty.
The GMM-FINCH with 50\% feedback reacts the slowest of all the models over all sets, with only the first increment having a notably less than 1.0 reaction time score.
The GMM-FINCH with 100\% feedback  is the most sporadic in its reaction times as evident in its scores prior to the sixth increment.
The GMM-FINCH with 0\% feedback consistently has the fastest reaction time of all models.
This may be due to tending towards guessing ``unknown'' in favor of ``known,'' as the prior MCC and NMI measures in Figure~\ref{fig:detect} were very low. But its detection accuracy was not too low overall
and it did perform the best out of all predictors on novelty recognition, which indicates that out of all of the baseline predictors in this demonstration, it may be a contender for the best model in handling the novelty-related subtasks of HAR.

\begin{figure*}[t]
    \centering
    \small
    \textbf{AVA-Kinetics Ablation Study on KOWL-718 Testing Set for}
    
    \textbf{HAR Novel-to-Predictor Classification Task Cumulative Post-Feedback Performance}
    
    \includegraphics[width=\linewidth]{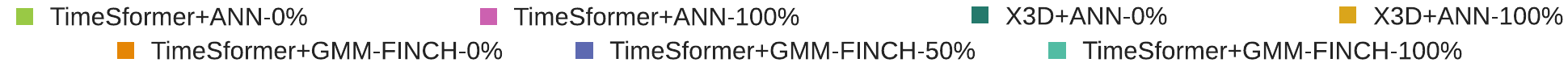}

    \subcaptionbox{
        \newline
        Kinetics Activity Type
        \label{fig:ava_type}
    }[.246\linewidth]{
        \includegraphics[width=\linewidth]{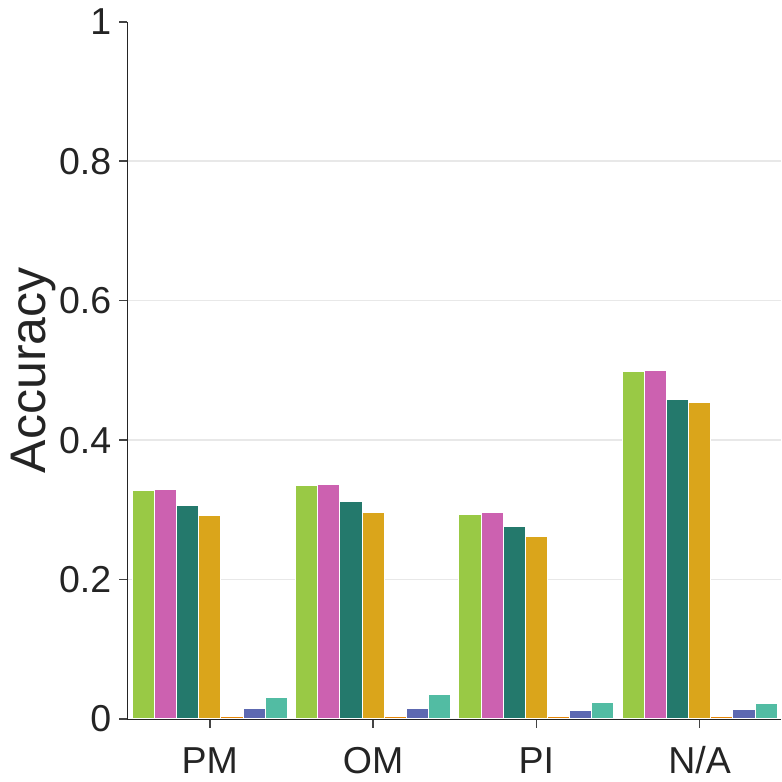}
        \includegraphics[width=\linewidth]{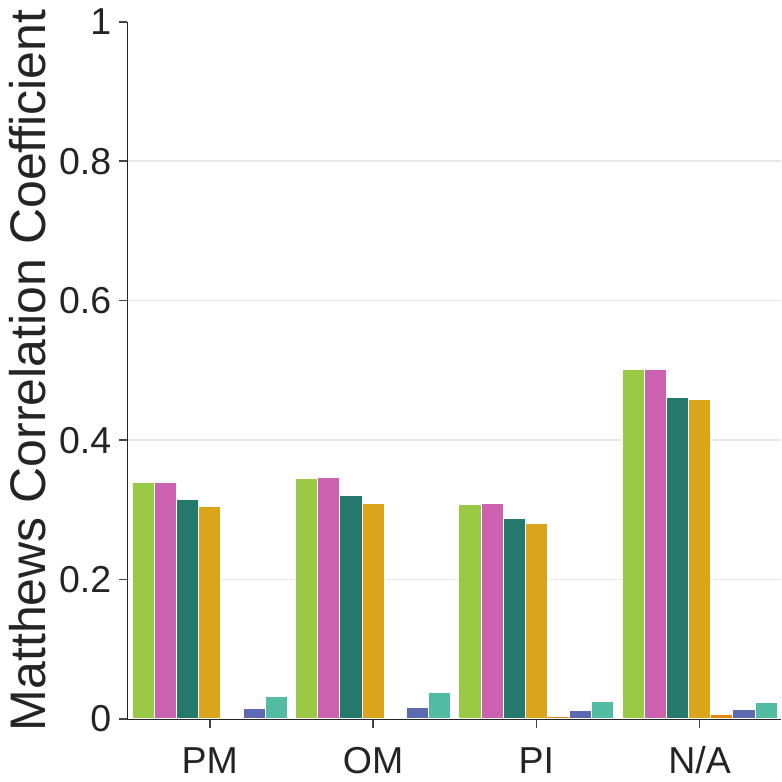}
        \includegraphics[width=\linewidth]{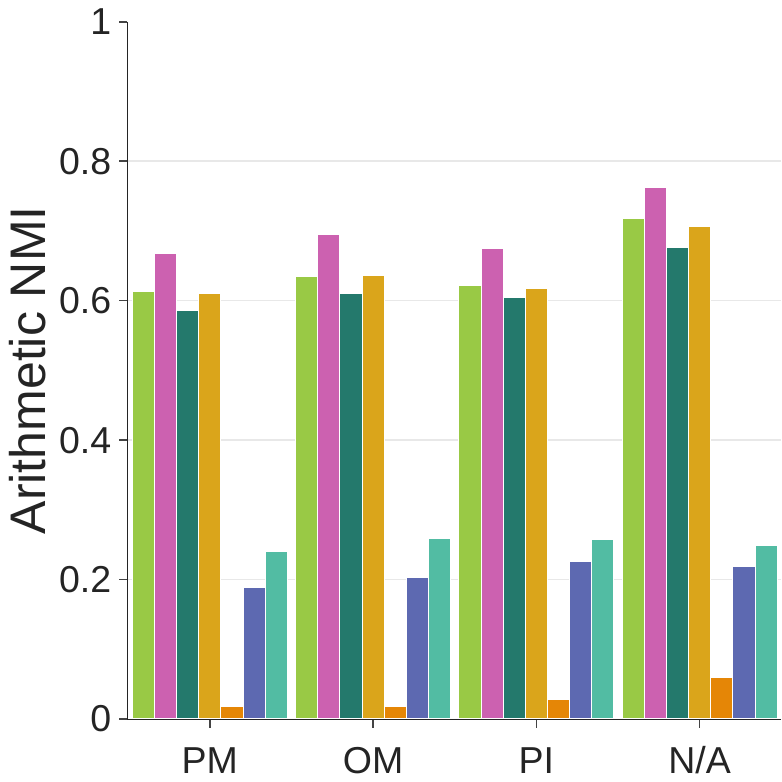}
    }
    \hfill
    \subcaptionbox{
        \newline
        Number of AVA Actions
        \label{fig:ava_acts}
    }[.246\linewidth]{
        \includegraphics[width=\linewidth]{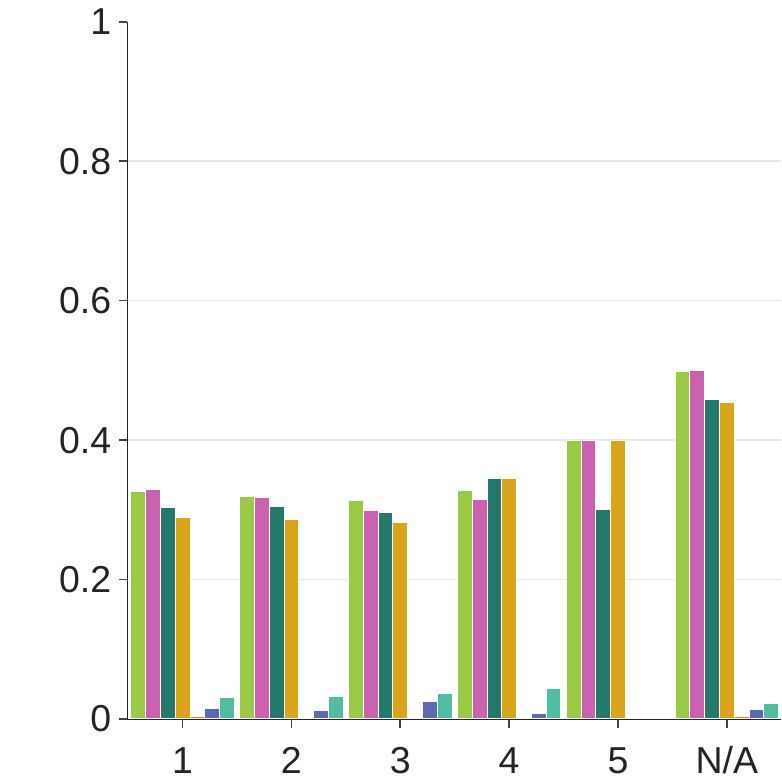}
        \includegraphics[width=\linewidth]{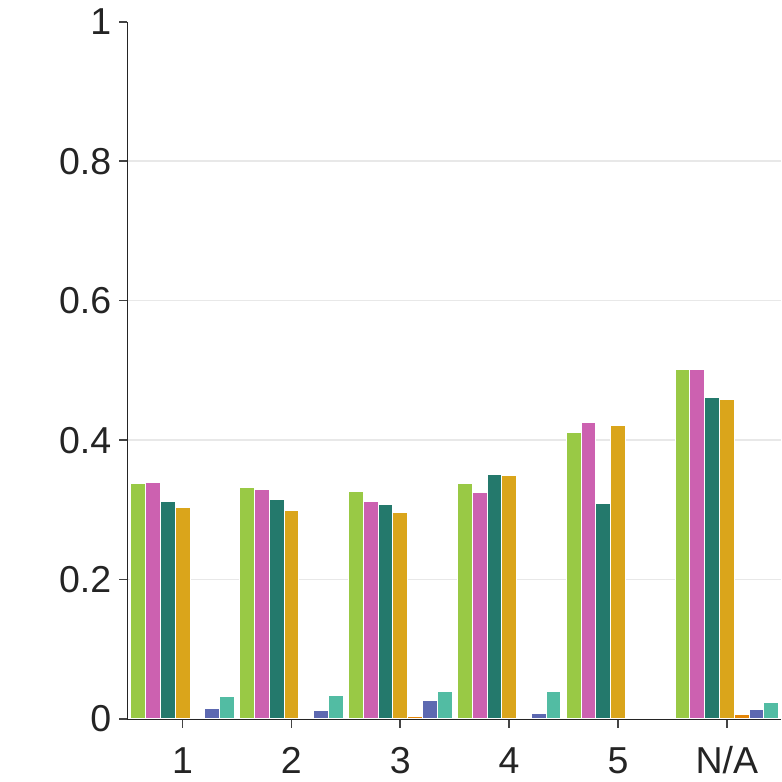}
        \includegraphics[width=\linewidth]{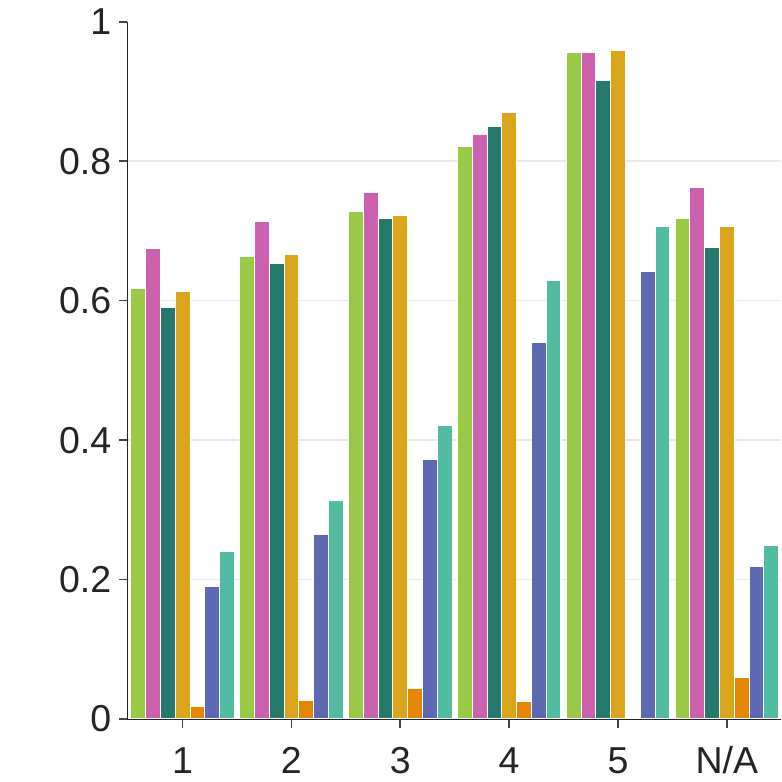}
    }
    \hfill
    \subcaptionbox{
        \newline
        Number of Humans
        \label{fig:ava_humans}
    }[.246\linewidth]{
        \includegraphics[width=\linewidth]{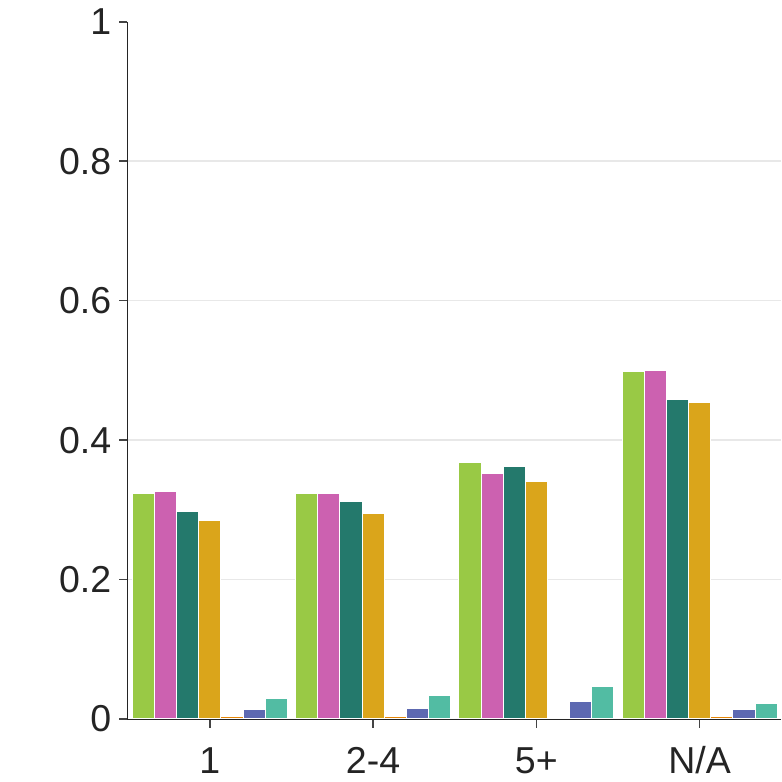}
        \includegraphics[width=\linewidth]{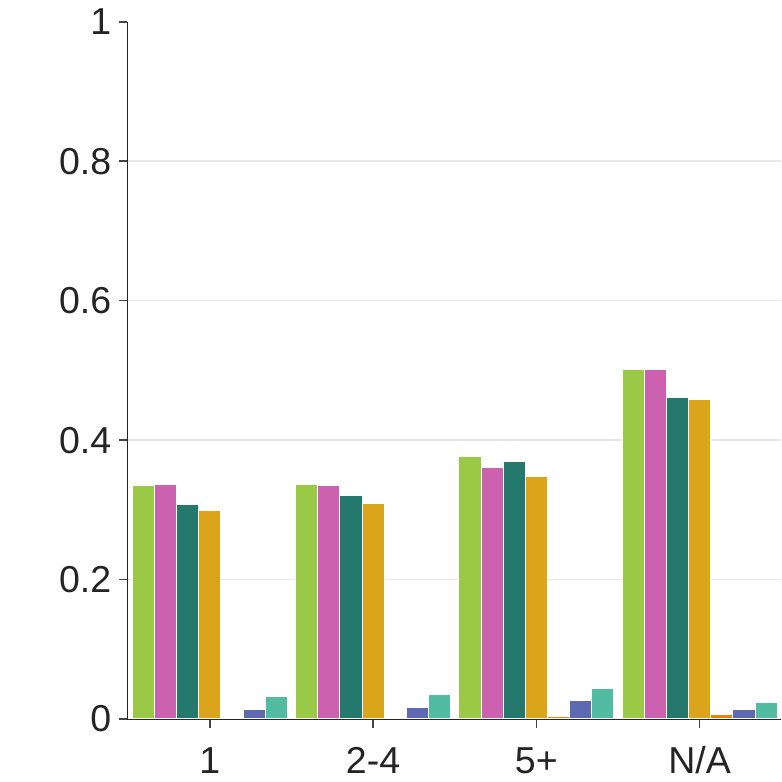}
        \includegraphics[width=\linewidth]{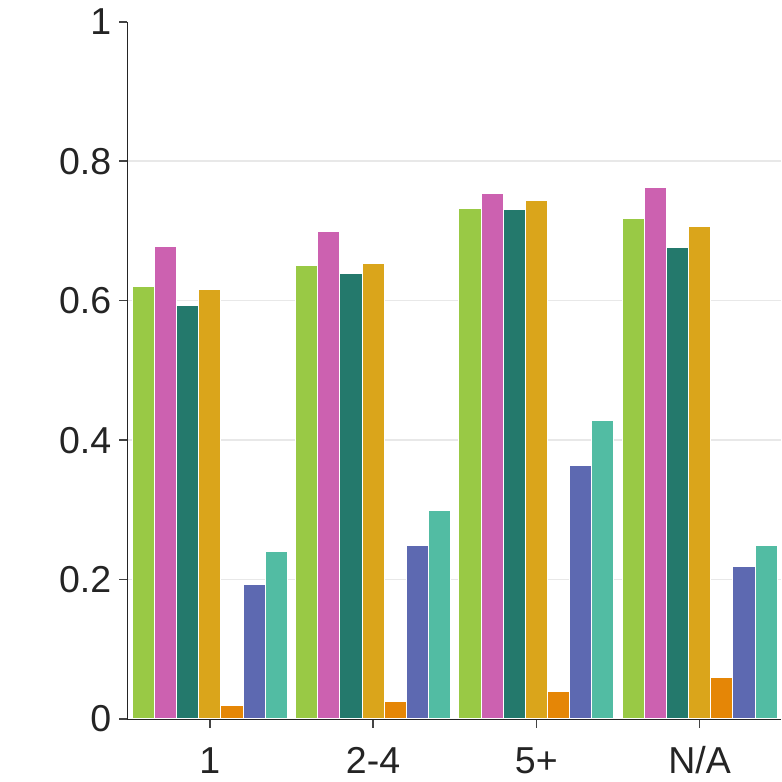}
    }
    \hfill
    \subcaptionbox{
        \newline
        Total Area of Humans
        \label{fig:ava_human_area}
    }[.246\linewidth]{
        \includegraphics[width=\linewidth]{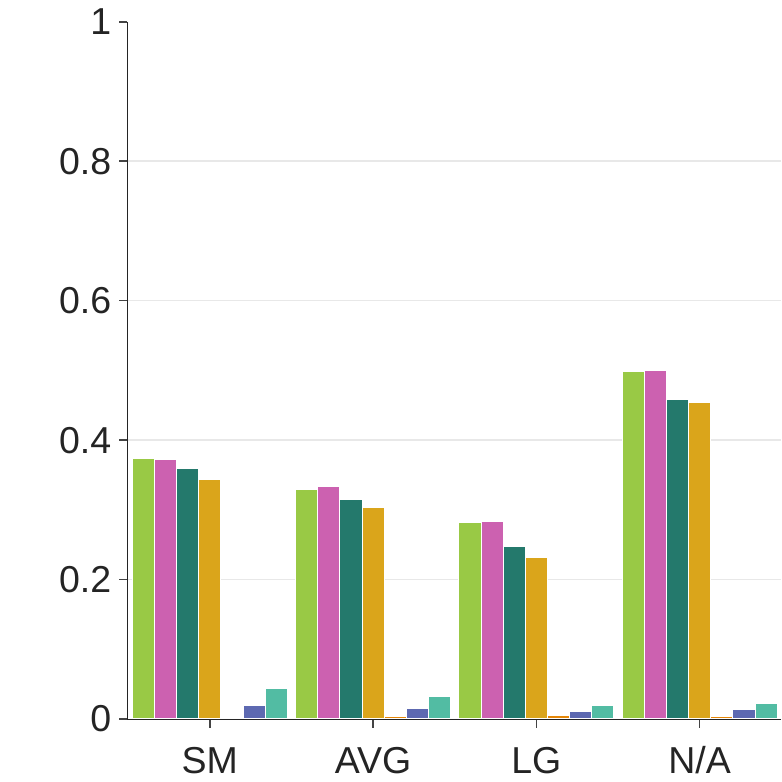}
        \includegraphics[width=\linewidth]{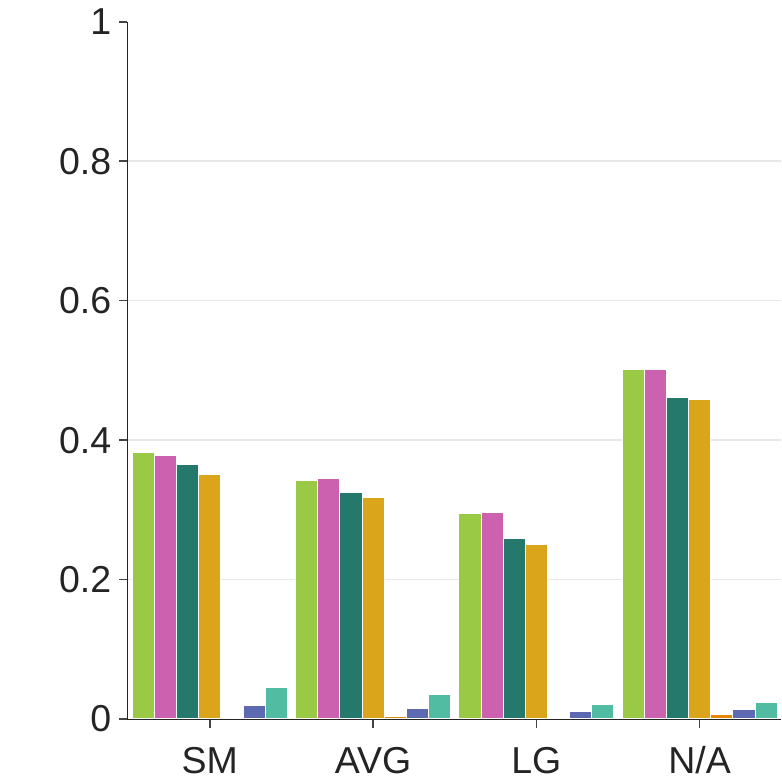}
        \includegraphics[width=\linewidth]{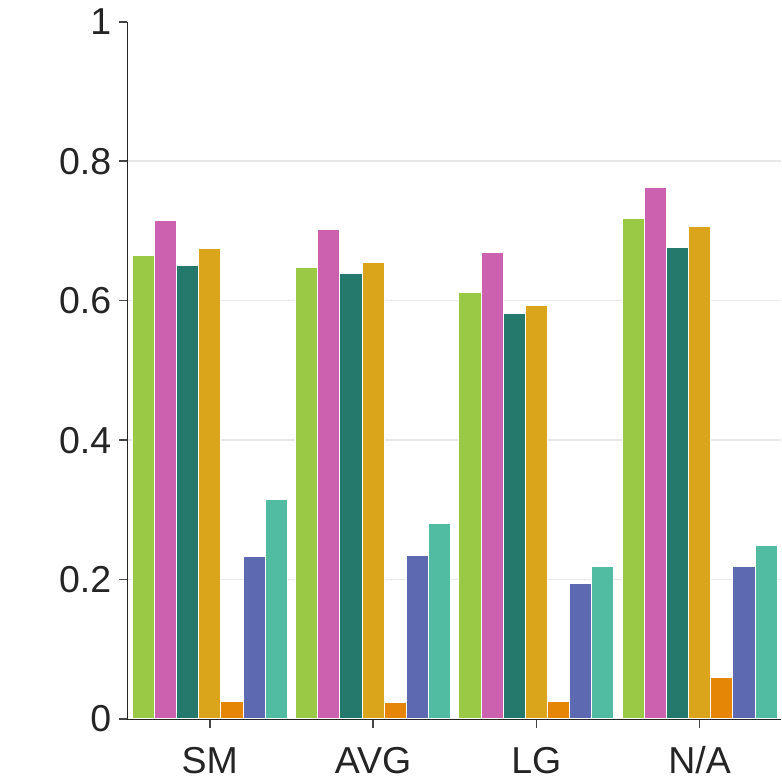}
    }
    \caption{
        \small
        Ablation study results on the classification task \textbf{cumulative post-feedback} performance on KOWL-718 under different annotation types using each video's annotated key frame in AVA-Kinetics~\cite{li_ava-kinetics_2020}, which only covers a subset of videos. 
        Any videos not annotated fall under ``N/A''.
        Cumulative means that all of the confusion matrices associated with the time-steps for the post-feedback phase were summed together, adding together the occurrences.
        AVA activity types are: Person Movement (PM), Object Manipulation (OM), and Person Interaction (PI).
        Total area of humans is the mean bounding box area normalized by the frame's total area, which was then binned into thirds as small (SM), average (AVG), and large (LG).
        See Section~\ref{sec:ava_kinetics} for details and discussion.
        \vspace{-2em}
    }
    \label{fig:ava_types}
\end{figure*}

\subsection{Ablation Study of Performance given AVA-Kinetics}
\label{sec:ava_kinetics}

The measurements so far only examine the performance of the different subtasks given novel human activities occurring in an open world.
To better analyze how performance is affected by characteristics of the video samples, 
an ablation study of the predictors' classification task performance over KOWL-718 was conducted using annotations from AVA-Kinetics~\cite{li_ava-kinetics_2020} to examine any difference in performance given certain characteristics of the videos.
AVA-Kinetics annotated different types of information for a single key frame on a subset of videos in the Kinetics datasets.
Using these annotations, we were able to observe the performance of the predictors under different circumstances, such as
    the different types of Kinetics actions,
    the different number of AVA actions occurring within a key frame,
    the number of humans that were in a key frame of a video,
    and the amount of pixel area occupied by humans using the arithmetic mean area of their bounding boxes in the key frames.
AVA-Kinetics provided a grouping of the Kinetics actions into three types.
The AVA actions differ from the Kinetics datasets' classes, which results in the possibility of multiple classes occurring in a single frame.
AVA-Kinetics records the AVA actions as bounding boxes over the humans performing the action.
This enables counting the number of AVA actions and humans occurring in a single key frame.
The total area of humans within a video was calculated by taking the arithmetic mean of the areas associated with the bounding boxes for the key frame.
Then those mean areas were binned into ``small,'' ``medium,'' and ``large,'' to enable an easier comparison of the amounts of average area the humans occupied within the frame.
To get the bins, the mean areas were put into ascending order split into thirds uniformly over the samples.

Figure~\ref{fig:ava_types} depicts the ablation study on cumulative post-feedback performance of the predictors on the HAR novel-to-predictor classification task over KOWL-718.
The measures over the three Kinetics Activity Types indicates minimal difference in performance but with Person Interaction activities having slightly worse symbolic performance across all predictors as depicted by accuracy and MCC.
The clearest depiction of a difference in performance across the ablation study is the Number of AVA Actions within a single key frame in~\ref{fig:ava_acts}.
Especially for Arithmetic NMI, the measured mutual information across all predictors increases as the number of AVA Actions within a key frame increases.
There is a subtle increase in the accuracy and MCC of the ANN predictors for five AVA actions versus less than four for AVA actions.
This could be due to more AVA actions within a key frame aiding a predictor in classifying the Kinetics activity class, but further detailed analysis and possibly more annotated data would be necessary to support this finding.
One possibility is that there are so few samples for the videos with more AVA actions, that they become easier to predict as they are learned. For example, group activities may be more easily determined if the predictors can determine 
Number of Humans follows a similar, albeit weaker, trend to the Number of AVA Actions, as seen by the measures in~\ref{fig:ava_humans}.
Total Area of Humans has a subtle decrease in the performance measures across all predictors as the mean area of the humans in the key frame increases.
This could indicate that more visual environmental context  may better inform the predictor of the Kinetics activity class.
However, it is important to be wary of this as it could also indicate overfitting due to over-dependence on the environment rather than the humans performing the action.
\begin{figure}[t]
    \centering
    \small{\textbf{
        State-of-the-art Model MCC on Kinetics-400
        Transformed with Nuisance Novelty
    }}
    \includegraphics[width=1.0\linewidth]{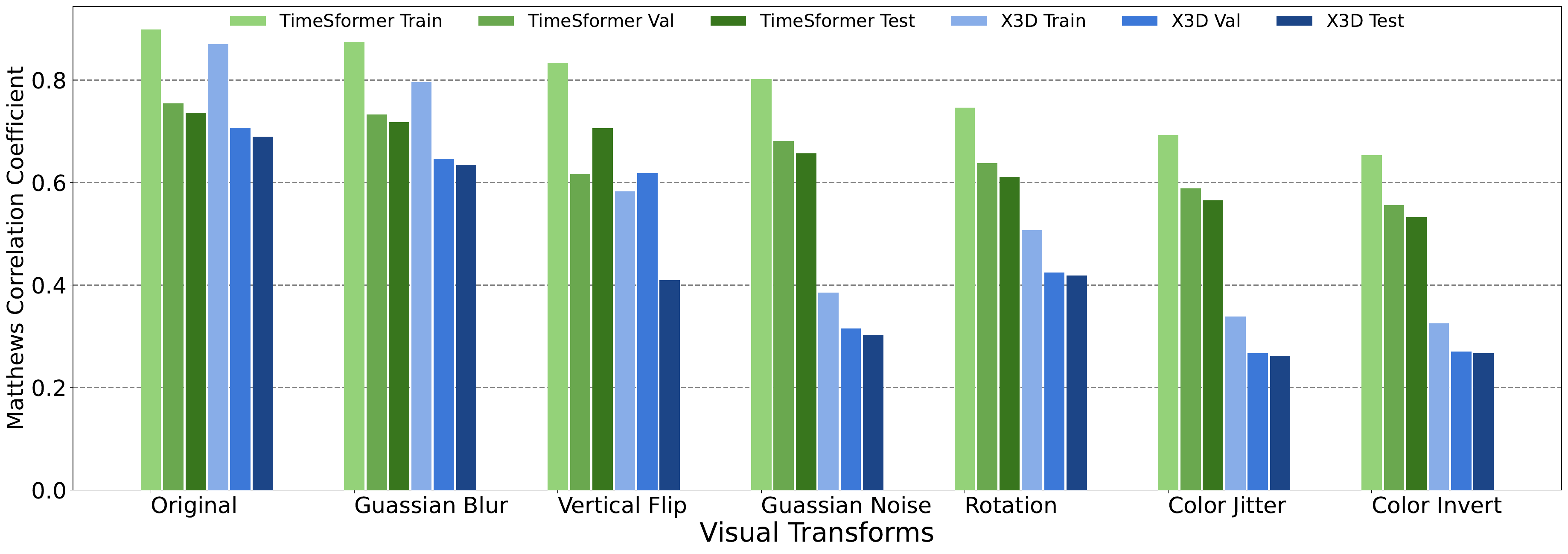}
    \vspace{-2.2em}
    \caption{
        \small
        To evaluate tolerance against novel video representations, the MCC of state-of-the-art models on visually transformed videos was recorded.
        X3D and TimeSformer 
        pretrained on Kinetics-400 given transformed Kinetics-400 videos
        both experience a performance decrease when encountering unseen visual transformations.
        The TimeSformer performed the best in all cases.
        This shows these models are not invariant to such visual transformations.
    }
    \label{fig:k400_transforms}
\end{figure}

\begin{figure*}
    \centering
    \includegraphics[width=\textwidth]{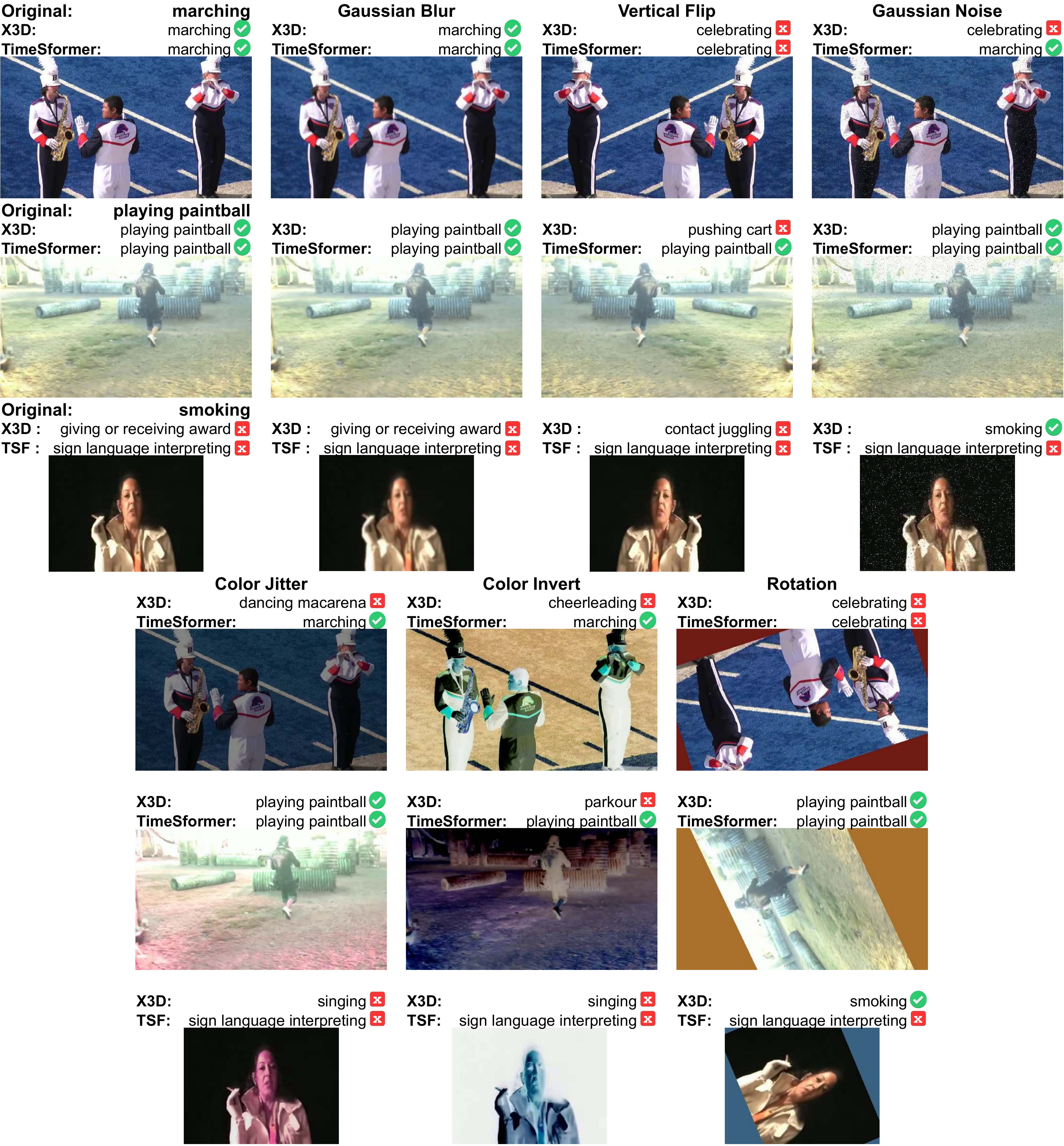}
    \caption{
        \small
        Visual transform examples on Kinetics-400 validation videos unseen by the predictors,
        along with their X3D and TimeSformer original pretrained classifier predictions marked as correct or incorrect.
        While not exact matches, the incorrect predictions have an understandable visual misinterpretation in these examples.
        The actual label was included in the top-5 predictions for all except for
            the marching example's
                X3D on color jitter and
                TimeSformer on flip and rotation;
            the playing paintball example's
                X3D on flip and color invert; and
            the smoking example's
                X3D on the original, Gaussian blur, and color jitter.
        TimeSformer always predicted ``smoking'' as its second most likely guess for the smoking example.
    }
\label{fig:visual_transforms}
\end{figure*}

\subsection{HAR Predictor Tolerance to Nuisance Novelty}
\vspace{-0.5em}
\label{sec:tolerance}
The above evaluation focused on action class specific novelty,
however
machine learning algorithms are desired to be tolerant to noise, and nuisance novelty is inherently never before seen noise patterns or other phenomena that are task-irrelevant.
We assessed the tolerance of pre-existing state-of-the-art models, X3D and TimeSformer, to novel visual representations
    by examining the performance of these models on Kinetics-400 versus Kinetics-400 with visual transformations not seen during training.
The videos are transformed using six different visual transformations unseen during training, shown in Figure~\ref{fig:visual_transforms}.
These transformations were applied to the videos with randomized parameters for each transform.
The visual transformations were performed such that identical transformations were performed on each frame from the same video, with the exception of noise, which was random for each frame.
The MCC of each predictor for each visual transform is depicted in Figure~\ref{fig:k400_transforms}.
We find that the Kinetics-400 pretrained X3D and TimeSformer are not invariant to such transformations, especially color inversion and the rotation of videos.
However, TimeSformer performs the best overall and is minimally affected by random Gaussian blur of videos.
\section{Conclusion}
With the increased interest in learning tasks with novelty~\cite{boult_towards_2021,langley_open-world_2020},
we formalized HAR in an open world and provided a reproducible and extendable OWL protocol to evaluate predictors in learning novel classes incrementally over time.
Applying this protocol to the popular Kinetics datasets, we constructed a practical benchmark, KOWL-718, that is the difficult real world problem of learning human activities from videos as new data becomes available, where unseen actions may be encountered.
This benchmark exemplifies the nuance in performing HAR in an open world as multiple subtasks are involved and intertwined together.
Our OWL protocol allows for analysis of different aspects of learning HAR in an open world, including varying levels of feedback.
Further analysis of predictor performance was recorded when the videos are visually transformed or when the number of people between videos differs.
The baseline predictors demonstrate that while these subtasks can be learned together, it is nontrivial to excel at all of the subtasks at the same time.
This primes future work on this benchmark which is unlike many others and is ripe for further research.

\section*{Acknowledgments}
This research was sponsored in part by the National Science Foundation (NSF) grant CAREER-1942151 and by the Defense Advanced Research Projects Agency (DARPA) and the Army Research Office (ARO) under multiple contracts/agreements including HR001120C0055, W911NF-20-2-0005, W911NF-20-2-0004, HQ0034-19-D-0001, W911NF2020009.
The views contained in this document are those of the authors and should not be interpreted as representing the official policies, either expressed or implied, of the DARPA or ARO, or the U.S. Government.

{\appendix
    \section{Three Spaces of Novelty for HAR}
\label{sec:spaces_novelty}

The section aligns OWL HAR with the framework introduced in~\citeA{boult_towards_2021} and the spaces in Figure~\ref{fig:novelty_wop_venn}.
The world's state space is the space of all human activities and an oracle knows the maximum likely activity class for each activity.
A dataset's ground-truth human activity labels along with the paired video serve as samples from the oracle.
The samples from the world are used to learn the task and assess performance based on feedback either directly or indirectly from an oracle, where the evaluator program serves as the oracle in a controlled benchmark setting, such as ours.
The true task in the world is approximated by sampling and it is the job of the oracle and the experimental designers to determine if the predictor is learning that actual task.
This means that the world dissimilarity and regret measures are never truly known, but are approximated via supervised learning when training the predictor.
HAR is a discrete classification task where predictors output their predicted nominal class label.
As such the confusion matrix and its derived measures
    may be used to assess predictor performance over a period of time, and thus serves as the world regret measure, given appropriate sign change for minimization of regret.
If the order or magnitude of weighting between mutually exclusive activity classes is desired, then the desired output of the predictor is a probability vector assigning probability to each of the known classes and unknown recognized class-clusters.
To analyze these, ordered top-$k$ confusion matrices may be used in order to obtain measure such as the top-$k$ accuracy.
The world experience is the past sequence of states indexed by the time-step.
The world state transition function is the experimental process used to determine the next sample video to give the predictor, based on the current time-step and world state.

The observation space is the encoding of world states after being processed by sensors, resulting in the sensor states.
In the case of visual HAR, the sensors are cameras that capture video footage of human activities, so the state space consists of the video frame with its color space.
The observation dissimilarity measure used in the literature tends to be Euclidean distance or the difference of video pixel values between frames.
Sometimes, when the observation space is considered a feature representation space by a smart camera, the inverse cosine similarity is used.
In the end, what matters is the information captured by the sensors of the world's state.
Information theoretic measures are the theoretically proper choice as noted in \cite{kinney_equitability_2014}, as long as they are able to be measured in practice.
All other measures on the same space will be equivalent or contain less information.
Learning what is task-relevant or -irrelevant from these sensors is the predictor's job.

\begin{figure}[t]
    \centering
    \includegraphics[width=0.70\linewidth]{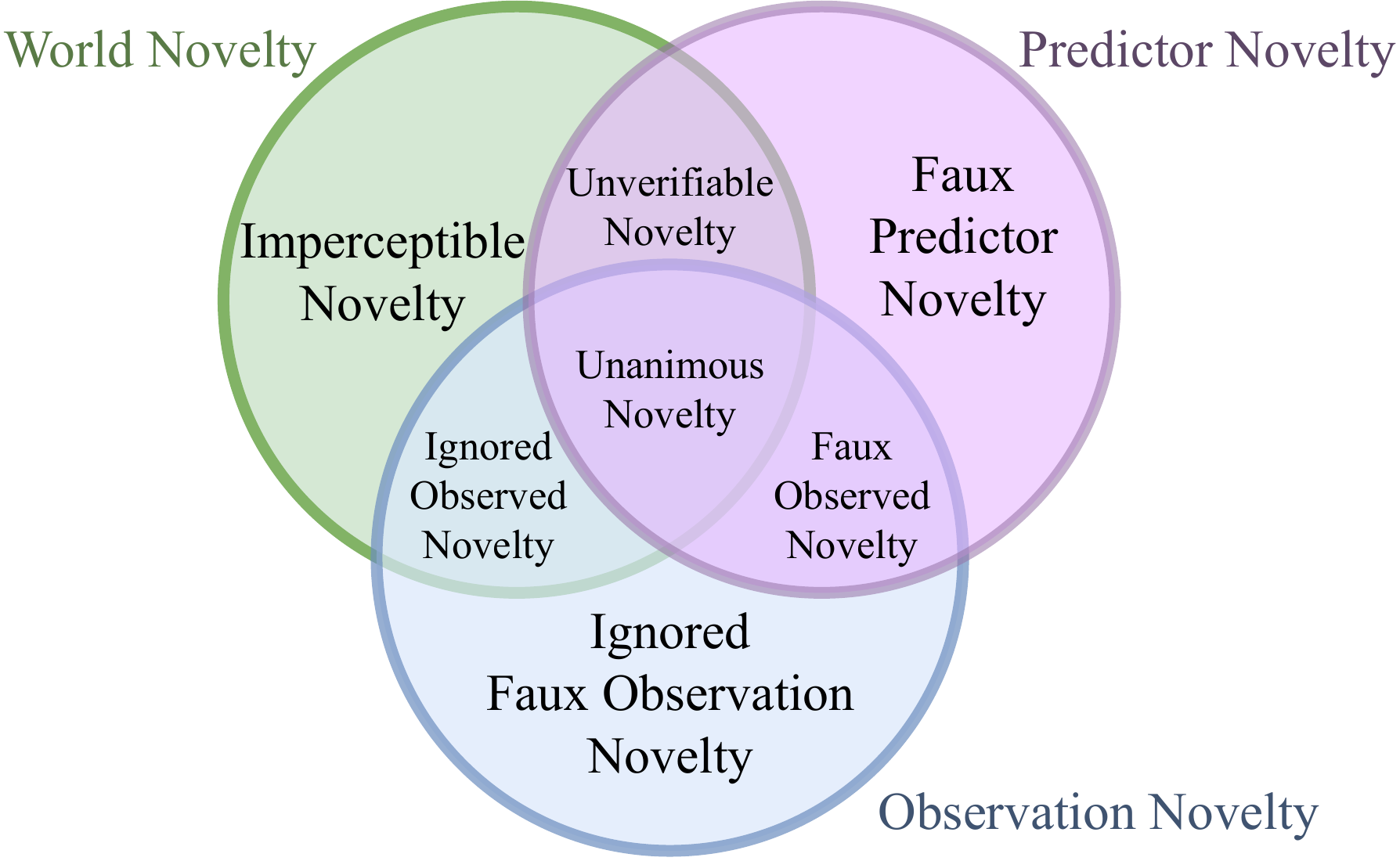}
    \vspace{-0.5em}
    \caption{
        \small
        Set representation of novelty spaces adapted from~\citeA{boult_towards_2021}.
        We use the term ``predictor'' in-place of ``agent''.
        See Section~\ref{sec:formal} and Appendix~\ref{sec:spaces_novelty}.
        \label{fig:novelty_wop_venn}
    }
\end{figure}

The observation regret measure is the regret of information in the observed state with respect to the task.
This may only known by the oracle or evaluator and differs from the world regret measure because it does not necessarily include all of the information that the world space contains.
This regret measure is only applied in practice when the experimental designers consider and address the difference in information between the world space and the observation space.
The observation regret is not directly applied in this paper's experiments as it is determined by the sensors used to capture the videos and would be expressed by meta-data or further annotations of the videos.
The observation experience is the time-step indexed history of prior sensor states, which is the videos in HAR.
In the case of real world applications, the observation experience may have a limited capacity to store the prior states, and so this may be defined as a buffer and thus forgotten states will be deemed novel if encountered again.
The oracle or experimental designer may know the truth in such a case, and assess the predictor with this knowledge.
The observation state transition function is the transformation of the given world state through the sensors that form the observation space.

The predictor space is the space of the predictor's model and states.
The HAR predictor decides on the activity labels, and depending on setup of experiments and available data, may request feedback.
The predictor's dissimilarity measure  may be cosine similarity, Euclidean distance, or any measure that applies to the predictor's encoded space of the activity phenomena. 
The predictor's regret measure  tends to be its objective function, \eg, loss function, regret function, or a heuristic.
Recall that the theoretically correct measures are those that measure the information such that task relevant and irrelevant information may be determined~\cite{kullbackInformationSufficiency1951,tishby_information_1999,kinney_equitability_2014}.
The predictor's experience is often limited by physical and practical constraints given it is a computer program, thus it may be in practice a finite sized memory buffer.
In this paper's experiments using Kinetics, we did not encounter physical limits of the memory buffer.
We neither applied any constraints to explore learning with a limited memory resource.
The predictor's state transition function takes the given observations, optionally updates its internal state, and predicts the activity class while also detecting novelty.
    \section{Formalism: HAR Novelty Set Relations}
\label{sec:har_novelty_sets}
With defined novelty spaces for HAR, we may further specify the types of novelty encountered in this domain following~\citeA{boult_towards_2021}.
Given the three novelty spaces, the presence of novelty across them forms set relationships as visualized in Figure~\ref*{fig:novelty_wop_venn}.
The entire list of defined sets of novelty given the three novelty spaces for HAR is as follows:
    \textbf{Unanimous Novelty}:
        This is novelty that occurs in the world, is observable via the sensors, and is detected by the predictor.
        This is the True Positive case of novelty detection.
        For example, correctly detecting the first ``blowdrying hair'' sample or ``swimming with dolphins'' sample in Figure~\ref{fig:teaser}.
    
    \textbf{Imperceptible Novelty}:
        Novelty that occurs in the world but is unable to be sensed by the sensors.
        In HAR, this can be wavelengths of light that are unable to be sensed, such as infrared or ultra violet light that cannot be recorded with an RGB sensor.
        This could also include novelty that is not within the camera's field of view or is occluded, which could affect the activity being recorded.
        For example, in a multi-person team sport, if something happens off camera that causes all teammates to stop and stare, then the cause of them staring is a novelty imperceptible to the camera. 
        
    \textbf{Ignored Observed Novelty}:
        Observed novelty that is ignored by the predictor because it is deemed as task-irrelevant or because the predictor is simply unable to  interpret the novelty in its input.
        The ironing hair example from Figure~\ref{fig:teaser} is a case where the predictor sees hair being brushed, but does not recognize the hair iron given it is not task-relevant based on the prior known classes and data.
        
    \textbf{Ignored Faux Observation Novelty}:
        This is novelty that is introduced in the observation novelty space that is ignored by the predictor and thus not deemed as a novelty to it.
        An example is when the predictor is tolerant to hue shifts in the visual input that occur due to sensor error and performs as expected on the HAR task.
        
    \textbf{Faux Observed Novelty}:
        In contract to Ignored Faux Observation Novelty, Faux Observed Novelty is novelty that originates in the observation novelty space but is also detected as novelty by the predictor and thus is included in the predictor novelty space.
        Following the hue shift from the prior example, this is when the predictor is not tolerant to hue shifts and detects them as novelty.
        
    \textbf{Faux Agent Novelty}:
        This is when the predictor falsely detects novelty in its input space (feature space), otherwise known as a False Positive novelty detection.
        In Figure~\ref{fig:teaser}, if the third video of ``walking the dog'' was classified as a novel human activity class, this would be the agent misunderstanding the visual input of three dogs on a leash as a new class.
        Perhaps seeing three dogs together is novel, but in this case the human activity is still known and the presence of more dogs results in a nuisance novelty.
        
    \textbf{Unverifiable Novelty}:
        This is when a predictor correctly detects novelty in the state, but this cannot be confirmed by the predictor because the sensors cannot sense this novelty.
        This occurs when the predictor makes a lucky guess, leverages latent knowledge about that task, or uses knowledge external to what was found in the sensors, such as prior knowledge included by human designers.
        If a camera's point of view misses a novelty that has occurred off camera, then at that moment of time it cannot verify that novelty occurred or exists, but may be able to do so later in the future by simply shifting its point of view.

    \section{Additional Experiment Details}
\label{sm:sec:exps}

\subsection{Data Specifics}
\label{sm:sec:data}
Collecting the Kinetics datasets is unfortunately not straight forward for such a standard benchmark.
The data is officially distributed by Google as a list of YouTube IDs.
A large portion of the videos has fallen victim to link rot where the links are broken and the videos are no longer available, especially in the older Kinetics 400 and 600 sets.  
Fortunately, there are at least two archives of this data that offer a more complete version of the dataset, a torrent for Kinetics-700\footnote{
    \url{https://academictorrents.com/details/49f203189fb69ae96fb40a6d0e129949e1dfec98}
}
and Kinetics-400\footnote{
    \url{https://academictorrents.com/details/184d11318372f70018cf9a72ef867e2fb9ce1d26}
},
and a repository hosted by the CVDF\footnote{\url{https://github.com/cvdfoundation/kinetics-dataset}}.
This paper used the CVDF repository.
Information about missing samples in our experiments is available in the project's git repository.
See Table~\ref{sm:tab:raw_k} for the breakdown of the total number of samples used and activity classes per increment, as well as their original source Kinetics dataset that was used to construct KOWL-718.

\begin{table}[t]
    \scalebox{.9}{
{\small
\begin{tabular}{l|l|lllll|lllllll}
\toprule
Source Dataset & K-400 &  
    \multicolumn{5}{c|}{Kinetics-600} &
    \multicolumn{5}{c}{Kinetics-700-2020} \\
\midrule
\multicolumn{1}{r|}{Increments} & 0          &      1 &      2 &      3 &      4 &      5 &      6 &      7 &      8 &      9 &     10 \\

\midrule

\textbf{Train}  &&&&&&&&&&& \\
Known  &&&&&&&&&&& \\
\multicolumn{1}{r|}{classes} & 409 &   409 &   455 &   501 &   547 &   593 &   636 &   653 &   670 &   687 &   704 \\
\multicolumn{1}{r|}{samples} & 218371 &  16082 &  19148 &  24661 &  32391 &  44687 &  34287 &  36112 &  38527 &  41855 &  47128 \\
Novel  &&&&&&&&&&& \\
\multicolumn{1}{r|}{classes} & 0 &     46 &     46 &     46 &     46 &     43 &     17 &     17 &     17 &     17 &     14 \\
\multicolumn{1}{r|}{samples} & 0 &   3163 &   5513 &   7730 &  12298 &  25538 &   1870 &   2415 &   3331 &   5273 &  10144 \\

\midrule

\textbf{Validate}    &&&&&&&&&&& \\
Known &&&&&&&&&&& \\
\multicolumn{1}{r|}{classes} & 409 &   409 &   455 &   501 &   547 &   593 &   636 &   653 &   670 &   687 &   704 \\
\multicolumn{1}{r|}{samples} &     18016 & 333 &        20 &        45 &        88 &       156  &   4892 &  4990 &  5197 &  5473 &  5883 \\
Novel &&&&&&&&&&& \\
\multicolumn{1}{r|}{classes} &         1 &         46 &        46 &        46 &        46 &        43 &   17 &    17 &    17 &    17 &    14   \\
\multicolumn{1}{r|}{samples} &         0 &   75 &        72 &        68 &        75 &       124 & 163 &   208 &   277 &   411 &   684 \\

\midrule
\textbf{Test}    &&&&&&&&&&& \\
Known &&&&&&&&&&& \\
\multicolumn{1}{r|}{classes} & 409 &   409 &   455 &   501 &   547 &   593 & 0 &   0 &   0 &   0 &   0    \\
\multicolumn{1}{r|}{samples} & 35256 &   795 &   287 &   446 &   644 &   940 &   0 &   0 &   0 &   0 & 0\\
Novel &&&&&&&&&&& \\
\multicolumn{1}{r|}{classes} & 1 &    46 &    46 &    46 &    46 &    43 & 0 &   0 &   0 &   0 & 0\\
\multicolumn{1}{r|}{samples} & 1 &   104 &   160 &   198 &   297 &   488 &0 &   0 &   0 &   0 & 0 \\

\bottomrule
\end{tabular}
}}
    \caption{
        \small
        The resulting number of known and novel classes and their samples \textbf{per increment} for the Kinetics-400, -600, and -700-2020 datasets using the configuration in Section~\ref{sec:exps}:
        most recent label first, first come first assigned data split, and most frequent novel class first per future dataset.
        The Kinetics datasets unfortunately experienced link rot and some original samples have been lost.
        The single novel class and its single sample in Kinetics-400 test is still included in KOWL-718, but Table~\ref{tab:kowl718} excluded it for brevity.
    }
    \label{sm:tab:raw_k}
\end{table}

\subsection{Predictor Specifics}
\label{sm:sec:agents}

\subsubsection{Feature Representation Model Details}
\label{sm:sec:frepr}

\paragraph{X3D Details}
The X3D~\shortcite{feichtenhofer_x3d_2020}~ model used in this work uses a pure pytorch implementaiton of X3D Multigrid~\cite{wu_multigrid_2020}, found at:~\url{https://github.com/kkahatapitiya/X3D-Multigrid}. We used the pretrained weights for the ``M'' model trained on Kinetics 400 in the official distribution from FAIR found in~\cite{fan2020pyslowfast}. The reported performance of this model is as follows: Top-1 10 view: 75.1, Top-1 30 view: 76.1. The aforementioned repo's model zoo from which we pulled the weights can be found at~\url{https://github.com/facebookresearch/SlowFast/blob/main/MODEL_ZOO.md} ported into Kahatapitiya's pure PyTorch implementation.
After porting the weights, we were able to approximately verify the model's performance on our version of Kinetics 400, with some variance due to issues such as dataset collection difficulties and differences in video preprocessing, our performance is shown in MCC as ``Original" in Table~\ref{sm:tbl:task}

\paragraph{TimeSformer Details}
The TimeSformer feature extractor used in this work was adapted from the Kinetics-400 reference model distributed by \cite{bertasius_is_2021} from 
\url{https://github.com/facebookresearch/TimeSformer}.
Specifically, we used the one labeled as ``TimeSformer'' for which they gave the following statistics: number of frames: 8, spatial crop: 224, top 1 accuracy: 77.9, top 5 accuracy: 93.2.
The code was only modified to return the feature embedding from the encoder as an additional output.
The raw TimeSformer result in MCC are shown in the ``Original" row of Table~\ref{sm:tbl:task}

\subsubsection{Finetuned Classifier Model Details}
\label{sm:sec:finetunes}

The primary finetuned classifier models include an ANN that takes as its input the feature representation of the samples and outputs a softmax classifier output, which is the unknown class with the currently known classes together as a probability vector.
The ANNs used for both of the feature representation models was a fully connected dense network of one hidden layer with a LeakyReLu activation that then went to the softmax classsifier layer which was sized as known classes + 1 for the unknown catch-all class.
In experimentation to find a suitable ANN architecture, multiple hidden layers and hidden layer widths were examined, including multiple layers with skip connections through concatenation.
In the end, the single hidden layer model with the same size as the feature representation vector was found to be the most performant, using a dropout during training of 0.5 probability.
We also explored different numbers of total training epochs ranging from [1, 100].
We found that 100 epochs would result in overfitting, as seen in Fig.~\ref{sm:fig:frepr}, which compares the performance of the Finetuned ANN for 100 epochs versus the original logit predictions of the feature representation's original classifier layer output. 
The runtimes of the 100 epochs ANN was approximately 8 hours of wall time on the Kinetics-400 dataset only using Kinetics-400 labels.
The runtime of the single epoch ANN was approximately 1 half-hour of wall time for the incremental learning experiment at the cost of an estimated 0.05 MCC performance loss based on the training run on only the Kinetics-400 data.

\begin{figure}[t]
    \centering
    \scalebox{.9}{
\begin{tabular}{lllll}
\toprule
F. Repr. & Classifier &  Train &    Val. &   Test \\
\midrule
                                X3D & Original & 0.8701 & 0.7067 & 0.6893 \\
                                 X3D & ANN 100 & \textbf{0.9891} & 0.7054 & 0.6890 \\
                                   X3D & ANN 1 & 0.8414 & 0.6929 & 0.6740 \\
                             X3D & ANN 1 + EVM & 0.5777 & 0.4836 & 0.4698 \\
                                TSF & Original & 0.8988 & \textbf{0.7548} & \textbf{0.7361} \\
                                 TSF & ANN 100 & \textit{0.9817} & \textit{0.7546} & \textit{0.7344} \\
                                   TSF & ANN 1 & 0.8566 & 0.7410 & 0.7237 \\
                                     TSF & EVM & 0.5552 & 0.4995 & 0.4938 \\
\bottomrule
\end{tabular}
    }
    \captionof{table}{
        \small
        The HAR \textbf{classification task} performance as measured by MCC
        on \textbf{Kinetics-400 with its original labels}
        comparing different feature representation and classifier combinations of the predictor.
        The Original uses the X3D's~\cite{feichtenhofer_x3d_2020} and TimeSformer's~\cite{bertasius_is_2021} (TSF) original classifier layers with pretrained weights.
        These MCC values correspond to the bar chart in Fig.~\ref{sm:fig:frepr}.
    }
    \label{sm:tbl:task}
\end{figure}

An Extreme Value Machine (EVM) was also evaluated.
However, as seen in Fig.~\ref{sm:fig:frepr}, the EVM performed worse than the ANN on the HAR task.
Given this in addition to its longer computation time, we excluded it from the experiment in the main paper.
Given that Kinetics-400 does is not necessarily representative of Kinetics-600 and -700, note that the EVM may have resulted in better novelty detection performance over the incremental learning than the base finetuned ANN without any further novelty handling than a single unknown class included at each time-step.
The EVM's runtime on the Kinetics-400 only training totaled approximately 16 hours with a loss of performance relative to both the feature representations' original classifiers and the fine-tuned ANN.

\begin{figure}[t]
    \centering
    \includegraphics[width=.75\linewidth]{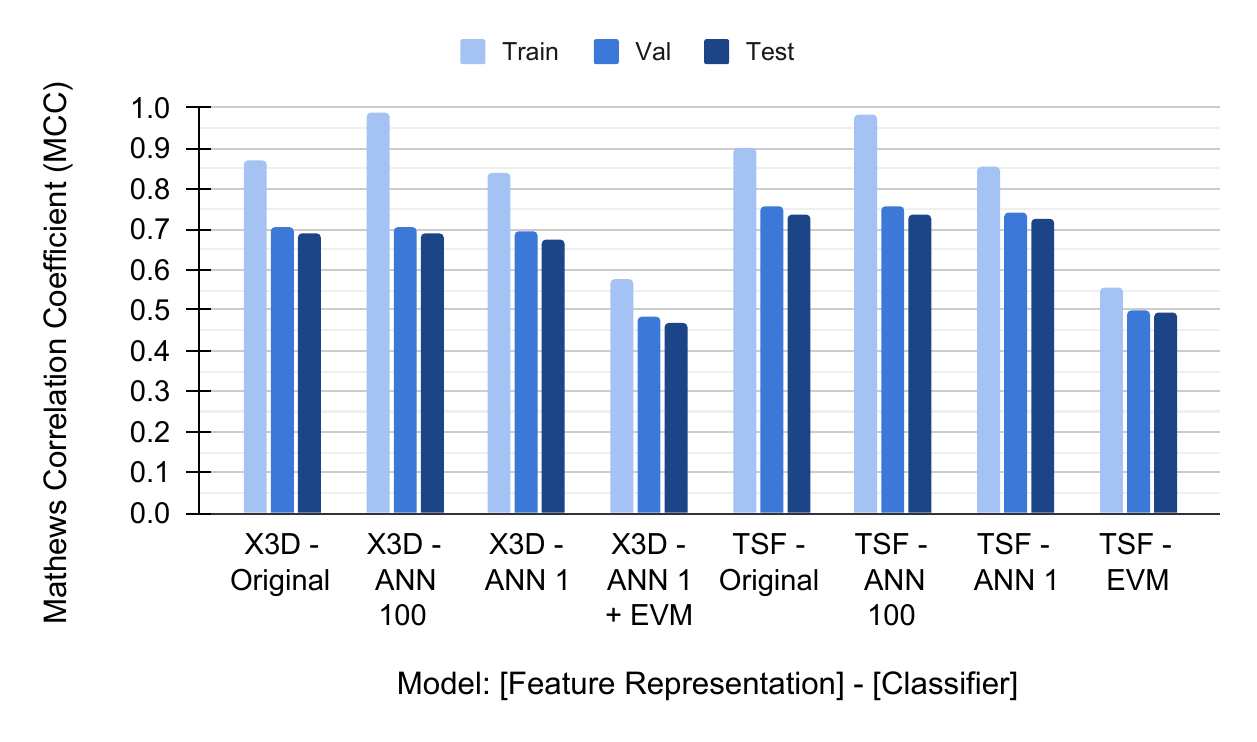}
    \caption{
        \small
        The Extreme Value Machine (EVM) \cite{rudd_extreme_2018} was  less performant with a longer run-time to fit of about 8 hr.
        The ANN 100 was a fully connected ANN with a single hidden layer trained for 100 epochs, which cost about the same time as the EVM to fit. 
        The ANN fit with only 1 epoch took dramatically less run-time to fit and predict versus the rest at the cost of less than a $\sim$2\% decrease in validation and test performance to the original classifier layer.
    }
    \label{sm:fig:frepr}
\end{figure}

\subsubsection{Hardware, Resources, and Computational Runtimes}
\label{sm:sec:resources}
These models were run on a variety of compute resources.
In order to facilitate experimentation, we first ran feature extraction on the entire unified datasets (KOWL-718) with both the X3D and TimeSformer models.
A full run of the feature extraction takes approximately 30 hours for the X3D feature extractor and approximately 23 days for the TimeSformer feature extractor when using an Nvidia Quadro RTX 6000.
In order to speed up the wall clock time of this process, we parallel processed the data in 60 equal sized batches, where each batch took about 30 minutes for X3D and 9 hours for TimeSformer on the RTX 6000.

\subsection{Predictor Tolerance to Nuisance Novelty}
\label{sm:sec:exp1}
In order to evaluate how the models held up to spurious representation novelty, we evaluated model performance after the video was modified using a number of common visual transformation techniques.
This meant that we needed to run a full pass on the unified dataset, KOWL-718, for each of the 6 types of visual transformations.
This resulted in approximately 194 GPU days for feature extraction.

Once feature extraction was completed, the computational cost for the Kinetics 400 visual transform assessment was quite low in Section~\ref{sec:eval_nuisance}.
During the feature extraction process, the extracted features were saved to reduce the overhead of running experiments multiple times.
Without saving the state, the compute overhead would have been substantially higher.
}

\bibliography{har_novelty}

\begin{thebibliography}{}

\bibitem[\protect\BCAY{Beddiar, Nini, Sabokrou,\ \BBA\ Hadid}{Beddiar et~al.}{2020}]{beddiar_vision-based_2020}
Beddiar, D.~R., Nini, B., Sabokrou, M., \BBA\ Hadid, A. \BBOP2020\BBCP.
\newblock \BBOQ Vision-based human activity recognition: a survey\BBCQ\
\newblock {\Bem Multimed Tools Appl}, {\Bem 79\/}(41), 30509--30555.

\bibitem[\protect\BCAY{Belouadah, Popescu,\ \BBA\ Kanellos}{Belouadah et~al.}{2021}]{belouadah_comprehensive_2021}
Belouadah, E., Popescu, A., \BBA\ Kanellos, I. \BBOP2021\BBCP.
\newblock \BBOQ A comprehensive study of class incremental learning algorithms for visual tasks\BBCQ\
\newblock {\Bem Neural Networks}, {\Bem 135}, 38--54.

\bibitem[\protect\BCAY{Bendale\ \BBA\ Boult}{Bendale\ \BBA\ Boult}{2015}]{bendale_towards_2015}
Bendale, A.\BBACOMMA\  \BBA\ Boult, T. \BBOP2015\BBCP.
\newblock \BBOQ Towards {Open} {World} {Recognition}\BBCQ\
\newblock In {\Bem Conference on {Computer} {Vision} and {Pattern} {Recognition}}, \BPGS\ 1893--1902.

\bibitem[\protect\BCAY{Bertasius, Wang,\ \BBA\ Torresani}{Bertasius et~al.}{2021}]{bertasius_is_2021}
Bertasius, G., Wang, H., \BBA\ Torresani, L. \BBOP2021\BBCP.
\newblock \BBOQ Is {Space}-{Time} {Attention} {All} {You} {Need} for {Video} {Understanding}?\BBCQ\
\newblock In {\Bem Proceedings of the 38th {International} {Conference} on {Machine} {Learning}}, \BPGS\ 813--824. PMLR.
\newblock ISSN: 2640-3498.

\bibitem[\protect\BCAY{Boult, Cruz, Dhamija, Gunther, Henrydoss,\ \BBA\ Scheirer}{Boult et~al.}{2019}]{boult_learning_2019}
Boult, T.~E., Cruz, S., Dhamija, A.~R., Gunther, M., Henrydoss, J., \BBA\ Scheirer, W.~J. \BBOP2019\BBCP.
\newblock \BBOQ Learning and the {Unknown}: {Surveying} {Steps} toward {Open} {World} {Recognition}\BBCQ\
\newblock {\Bem Proceedings of the AAAI Conference on Artificial Intelligence}, {\Bem 33\/}(01), 9801--9807.
\newblock Number: 01.

\bibitem[\protect\BCAY{Boult, Grabowicz, Prijatelj, Stern, Holder, Alspector, Jafarzadeh, Ahmad, Dhamija, Li, Cruz, Shrivastava, Vondrick,\ \BBA\ Walter}{Boult et~al.}{2021}]{boult_towards_2021}
Boult, T., Grabowicz, P., Prijatelj, D., Stern, R., Holder, L., Alspector, J., Jafarzadeh, M.~M., Ahmad, T., Dhamija, A., Li, C., Cruz, S., Shrivastava, A., Vondrick, C., \BBA\ Walter, S. \BBOP2021\BBCP.
\newblock \BBOQ Towards a {Unifying} {Framework} for {Formal} {Theories} of {Novelty}\BBCQ\
\newblock {\Bem Proceedings of the AAAI Conference on Artificial Intelligence}, {\Bem 35\/}(17), 15047--15052.
\newblock Number: 17.

\bibitem[\protect\BCAY{Brighi, Franco,\ \BBA\ Maio}{Brighi et~al.}{2021}]{brighi_activityexplorer_2021}
Brighi, M., Franco, A., \BBA\ Maio, D. \BBOP2021\BBCP.
\newblock \BBOQ {ActivityExplorer}: {A} semi-supervised approach to discover unknown activity classes in {HAR} systems\BBCQ\
\newblock {\Bem Pattern Recognition Letters}, {\Bem 151}, 340--347.

\bibitem[\protect\BCAY{Carreira, Noland, Banki-Horvath, Hillier,\ \BBA\ Zisserman}{Carreira et~al.}{2018}]{carreira_short_2018}
Carreira, J., Noland, E., Banki-Horvath, A., Hillier, C., \BBA\ Zisserman, A. \BBOP2018\BBCP.
\newblock \BBOQ A {Short} {Note} about {Kinetics}-600\BBCQ\
\newblock {\Bem arXiv:1808.01340 [cs]}, {\Bem ""}.
\newblock arXiv: 1808.01340.

\bibitem[\protect\BCAY{Carreira, Noland, Hillier,\ \BBA\ Zisserman}{Carreira et~al.}{2019}]{carreira_short_2019}
Carreira, J., Noland, E., Hillier, C., \BBA\ Zisserman, A. \BBOP2019\BBCP.
\newblock \BBOQ A {Short} {Note} on the {Kinetics}-700 {Human} {Action} {Dataset}\BBCQ\
\newblock {\Bem arXiv:1907.06987 [cs]}, {\Bem ""}.
\newblock arXiv: 1907.06987.

\bibitem[\protect\BCAY{Chalmers}{Chalmers}{2006}]{chalmers_strong_2006}
Chalmers, D.~J. \BBOP2006\BBCP.
\newblock \BBOQ Strong and weak emergence\BBCQ\
\newblock In {\Bem The {Re}-{Emergence} of {Emergence}: {The} {Emergentist} {Hypothesis} from {Science} to {Religion}}, \BPGS\ 244--415. OUP Oxford.
\newblock Google-Books-ID: gQZREAAAQBAJ.

\bibitem[\protect\BCAY{Cover\ \BBA\ Thomas}{Cover\ \BBA\ Thomas}{1991}]{cover_elements_1991}
Cover, T.~M.\BBACOMMA\  \BBA\ Thomas, J.~A. \BBOP1991\BBCP.
\newblock {\Bem Elements of {Information} {Theory}}.
\newblock Wiley {Series} in {Telecommunications}. John Wiley \& Sons, Inc., New York, USA.

\bibitem[\protect\BCAY{Dhamija, Ahmad, Schwan, Jafarzadeh, Li,\ \BBA\ Boult}{Dhamija et~al.}{2021}]{dhamija_self-supervised_2021}
Dhamija, A.~R., Ahmad, T., Schwan, J., Jafarzadeh, M., Li, C., \BBA\ Boult, T.~E. \BBOP2021\BBCP.
\newblock \BBOQ Self-{Supervised} {Features} {Improve} {Open}-{World} {Learning}\BBCQ\
\newblock {\Bem arXiv:2102.07848 [cs]}, {\Bem ""}.
\newblock arXiv: 2102.07848.

\bibitem[\protect\BCAY{Dosovitskiy, Beyer, Kolesnikov, Weissenborn, Zhai, Unterthiner, Dehghani, Minderer, Heigold, Gelly, Uszkoreit,\ \BBA\ Houlsby}{Dosovitskiy et~al.}{2022}]{dosovitskiy_image_2022}
Dosovitskiy, A., Beyer, L., Kolesnikov, A., Weissenborn, D., Zhai, X., Unterthiner, T., Dehghani, M., Minderer, M., Heigold, G., Gelly, S., Uszkoreit, J., \BBA\ Houlsby, N. \BBOP2022\BBCP.
\newblock \BBOQ An {Image} is {Worth} 16x16 {Words}: {Transformers} for {Image} {Recognition} at {Scale}\BBCQ\
\newblock In {\Bem International {Conference} on {Learning} {Representations}}.

\bibitem[\protect\BCAY{Fan, Li, Xiong, Lo,\ \BBA\ Feichtenhofer}{Fan et~al.}{2020}]{fan2020pyslowfast}
Fan, H., Li, Y., Xiong, B., Lo, W.-Y., \BBA\ Feichtenhofer, C. \BBOP2020\BBCP.
\newblock \BBOQ Pyslowfast\BBCQ\
\newblock \url{https://github.com/facebookresearch/slowfast}.

\bibitem[\protect\BCAY{Feichtenhofer}{Feichtenhofer}{2020}]{feichtenhofer_x3d_2020}
Feichtenhofer, C. \BBOP2020\BBCP.
\newblock \BBOQ {X3D}: {Expanding} {Architectures} for {Efficient} {Video} {Recognition}\BBCQ\
\newblock In {\Bem Proceedings of the {IEEE}/{CVF} {Conference} on {Computer} {Vision} and {Pattern} {Recognition}}, \BPGS\ 203--213.

\bibitem[\protect\BCAY{Geng, Huang,\ \BBA\ Chen}{Geng et~al.}{2021}]{geng_recent_2021}
Geng, C., Huang, S.-J., \BBA\ Chen, S. \BBOP2021\BBCP.
\newblock \BBOQ Recent {Advances} in {Open} {Set} {Recognition}: {A} {Survey}\BBCQ\
\newblock {\Bem IEEE Transactions on Pattern Analysis and Machine Intelligence}, {\Bem 43\/}(10), 3614--3631.
\newblock Conference Name: IEEE Transactions on Pattern Analysis and Machine Intelligence.

\bibitem[\protect\BCAY{Gershman\ \BBA\ Niv}{Gershman\ \BBA\ Niv}{2015}]{gershman_novelty_2015}
Gershman, S.~J.\BBACOMMA\  \BBA\ Niv, Y. \BBOP2015\BBCP.
\newblock \BBOQ Novelty and {Inductive} {Generalization} in {Human} {Reinforcement} {Learning}\BBCQ\
\newblock {\Bem Topics in Cognitive Science}, {\Bem 7\/}(3), 391--415.
\newblock \_eprint: https://onlinelibrary.wiley.com/doi/pdf/10.1111/tops.12138.

\bibitem[\protect\BCAY{Gutoski, Lazzaretti,\ \BBA\ Lopes}{Gutoski et~al.}{2021}]{gutoski_deep_2021}
Gutoski, M., Lazzaretti, A.~E., \BBA\ Lopes, H.~S. \BBOP2021\BBCP.
\newblock \BBOQ Deep metric learning for open-set human action recognition in videos\BBCQ\
\newblock {\Bem Neural Comput \& Applic}, {\Bem 33\/}(4), 1207--1220.

\bibitem[\protect\BCAY{He, Zhang, Ren,\ \BBA\ Sun}{He et~al.}{2016}]{he_deep_2016}
He, K., Zhang, X., Ren, S., \BBA\ Sun, J. \BBOP2016\BBCP.
\newblock \BBOQ Deep {Residual} {Learning} for {Image} {Recognition}\BBCQ\
\newblock In {\Bem Proceedings of the {IEEE} {Conference} on {Computer} {Vision} and {Pattern} {Recognition}}, \BPGS\ 770--778.

\bibitem[\protect\BCAY{Ho\ \BBA\ Vasconcelos}{Ho\ \BBA\ Vasconcelos}{2020}]{ho_contrastive_2020}
Ho, C.-H.\BBACOMMA\  \BBA\ Vasconcelos, N. \BBOP2020\BBCP.
\newblock \BBOQ Contrastive {Learning} with {Adversarial} {Examples}\BBCQ\
\newblock {\Bem arXiv:2010.12050 [cs]}, {\Bem ""}.
\newblock arXiv: 2010.12050.

\bibitem[\protect\BCAY{Inácio, Gutoski, Lazzaretti,\ \BBA\ Lopes}{Inácio et~al.}{2021}]{inacio_osvidcap_2021}
Inácio, A. D.~S., Gutoski, M., Lazzaretti, A.~E., \BBA\ Lopes, H.~S. \BBOP2021\BBCP.
\newblock \BBOQ {OSVidCap}: a {Framework} for the {Simultaneous} {Recognition} and {Description} of {Concurrent} {Actions} in {Videos} in an {Open}-{Set} {Scenario}\BBCQ\
\newblock {\Bem IEEE Access}, {\Bem ""}, 1--1.
\newblock Conference Name: IEEE Access.

\bibitem[\protect\BCAY{Kay, Carreira, Simonyan, Zhang, Hillier, Vijayanarasimhan, Viola, Green, Back, Natsev, Suleyman,\ \BBA\ Zisserman}{Kay et~al.}{2017}]{kay_kinetics_2017}
Kay, W., Carreira, J., Simonyan, K., Zhang, B., Hillier, C., Vijayanarasimhan, S., Viola, F., Green, T., Back, T., Natsev, P., Suleyman, M., \BBA\ Zisserman, A. \BBOP2017\BBCP.
\newblock \BBOQ The {Kinetics} {Human} {Action} {Video} {Dataset}\BBCQ\
\newblock {\Bem arXiv:1705.06950 [cs]}, {\Bem ""}.
\newblock arXiv: 1705.06950.

\bibitem[\protect\BCAY{Kinney\ \BBA\ Atwal}{Kinney\ \BBA\ Atwal}{2014}]{kinney_equitability_2014}
Kinney, J.~B.\BBACOMMA\  \BBA\ Atwal, G.~S. \BBOP2014\BBCP.
\newblock \BBOQ Equitability, mutual information, and the maximal information coefficient\BBCQ\
\newblock {\Bem PNAS}, {\Bem 111\/}(9), 3354--3359.

\bibitem[\protect\BCAY{Kullback\ \BBA\ Leibler}{Kullback\ \BBA\ Leibler}{1951}]{kullbackInformationSufficiency1951}
Kullback, S.\BBACOMMA\  \BBA\ Leibler, R.~A. \BBOP1951\BBCP.
\newblock \BBOQ On {Information} and {Sufficiency}\BBCQ\
\newblock {\Bem The Annals of Mathematical Statistics}, {\Bem 22\/}(1), 79--86.
\newblock Publisher: Institute of Mathematical Statistics.

\bibitem[\protect\BCAY{Langley}{Langley}{2020}]{langley_open-world_2020}
Langley, P. \BBOP2020\BBCP.
\newblock \BBOQ Open-{World} {Learning} for {Radically} {Autonomous} {Agents}\BBCQ\
\newblock {\Bem AAAI}, {\Bem 34\/}(09), 13539--13543.

\bibitem[\protect\BCAY{Li, Thotakuri, Ross, Carreira, Vostrikov,\ \BBA\ Zisserman}{Li et~al.}{2020}]{li_ava-kinetics_2020}
Li, A., Thotakuri, M., Ross, D.~A., Carreira, J., Vostrikov, A., \BBA\ Zisserman, A. \BBOP2020\BBCP.
\newblock \BBOQ The {AVA}-{Kinetics} {Localized} {Human} {Actions} {Video} {Dataset}\BBCQ\
\newblock {\Bem arXiv:2005.00214 [cs, eess]}, {\Bem ""}.
\newblock arXiv: 2005.00214.

\bibitem[\protect\BCAY{Losing, Hammer,\ \BBA\ Wersing}{Losing et~al.}{2018}]{losing_incremental_2018}
Losing, V., Hammer, B., \BBA\ Wersing, H. \BBOP2018\BBCP.
\newblock \BBOQ Incremental on-line learning: {A} review and comparison of state of the art algorithms\BBCQ\
\newblock {\Bem Neurocomputing}, {\Bem 275}, 1261--1274.

\bibitem[\protect\BCAY{Masana, Liu, Twardowski, Menta, Bagdanov,\ \BBA\ van~de Weijer}{Masana et~al.}{2021}]{masana_class-incremental_2021}
Masana, M., Liu, X., Twardowski, B., Menta, M., Bagdanov, A.~D., \BBA\ van~de Weijer, J. \BBOP2021\BBCP.
\newblock \BBOQ Class-incremental learning: survey and performance evaluation on image classification\BBCQ.
\newblock arXiv:2010.15277 [cs].

\bibitem[\protect\BCAY{Minh~Dang, Min, Wang, Jalil~Piran, Hee~Lee,\ \BBA\ Moon}{Minh~Dang et~al.}{2020}]{minh_dang_sensor-based_2020}
Minh~Dang, L., Min, K., Wang, H., Jalil~Piran, M., Hee~Lee, C., \BBA\ Moon, H. \BBOP2020\BBCP.
\newblock \BBOQ Sensor-based and vision-based human activity recognition: {A} comprehensive survey\BBCQ\
\newblock {\Bem Pattern Recognition}, {\Bem 108}, 107561.

\bibitem[\protect\BCAY{Pang, Shen, Cao,\ \BBA\ Hengel}{Pang et~al.}{2021}]{pang_deep_2021}
Pang, G., Shen, C., Cao, L., \BBA\ Hengel, A. V.~D. \BBOP2021\BBCP.
\newblock \BBOQ Deep {Learning} for {Anomaly} {Detection}: {A} {Review}\BBCQ\
\newblock {\Bem ACM Comput. Surv.}, {\Bem 54\/}(2), 38:1--38:38.

\bibitem[\protect\BCAY{Qiu, Sun, Xu, Shao, Dai,\ \BBA\ Huang}{Qiu et~al.}{2020}]{qiu_pre-trained_2020}
Qiu, X., Sun, T., Xu, Y., Shao, Y., Dai, N., \BBA\ Huang, X. \BBOP2020\BBCP.
\newblock \BBOQ Pre-trained models for natural language processing: {A} survey\BBCQ\
\newblock {\Bem Sci. China Technol. Sci.}, {\Bem 63\/}(10), 1872--1897.

\bibitem[\protect\BCAY{Radford, Kim, Hallacy, Ramesh, Goh, Agarwal, Sastry, Askell, Mishkin, Clark, Krueger,\ \BBA\ Sutskever}{Radford et~al.}{2021}]{radford_learning_2021}
Radford, A., Kim, J.~W., Hallacy, C., Ramesh, A., Goh, G., Agarwal, S., Sastry, G., Askell, A., Mishkin, P., Clark, J., Krueger, G., \BBA\ Sutskever, I. \BBOP2021\BBCP.
\newblock \BBOQ Learning {Transferable} {Visual} {Models} {From} {Natural} {Language} {Supervision}\BBCQ\
\newblock {\Bem arXiv:2103.00020 [cs]}, {\Bem ""}.
\newblock arXiv: 2103.00020.

\bibitem[\protect\BCAY{Rebuffi, Kolesnikov, Sperl,\ \BBA\ Lampert}{Rebuffi et~al.}{2017}]{rebuffi_icarl_2017}
Rebuffi, S.-A., Kolesnikov, A., Sperl, G., \BBA\ Lampert, C.~H. \BBOP2017\BBCP.
\newblock \BBOQ {iCaRL}: {Incremental} {Classifier} and {Representation} {Learning}\BBCQ\
\newblock In {\Bem Proceedings of the IEEE Conference on Computer Vision and Pattern Recognition (CVPR)}, \BPGS\ 2001--2010.

\bibitem[\protect\BCAY{Roitberg, Al-Halah,\ \BBA\ Stiefelhagen}{Roitberg et~al.}{2018}]{roitberg_informed_2018}
Roitberg, A., Al-Halah, Z., \BBA\ Stiefelhagen, R. \BBOP2018\BBCP.
\newblock \BBOQ Informed {Democracy}: {Voting}-based {Novelty} {Detection} for {Action} {Recognition}\BBCQ\
\newblock {\Bem arXiv:1810.12819 [cs]}, {\Bem ""}.
\newblock arXiv: 1810.12819.

\bibitem[\protect\BCAY{Roitberg, Ma, Haurilet,\ \BBA\ Stiefelhagen}{Roitberg et~al.}{2020}]{roitberg_open_2020}
Roitberg, A., Ma, C., Haurilet, M., \BBA\ Stiefelhagen, R. \BBOP2020\BBCP.
\newblock \BBOQ Open {Set} {Driver} {Activity} {Recognition}\BBCQ\
\newblock In {\Bem 2020 {IEEE} {Intelligent} {Vehicles} {Symposium} ({IV})}, \BPGS\ 1048--1053.
\newblock ISSN: 2642-7214.

\bibitem[\protect\BCAY{Rudd, Jain, Scheirer,\ \BBA\ Boult}{Rudd et~al.}{2018}]{rudd_extreme_2018}
Rudd, E.~M., Jain, L.~P., Scheirer, W.~J., \BBA\ Boult, T.~E. \BBOP2018\BBCP.
\newblock \BBOQ The {Extreme} {Value} {Machine}\BBCQ\
\newblock {\Bem IEEE Transactions on Pattern Analysis and Machine Intelligence}, {\Bem 40\/}(3), 762--768.

\bibitem[\protect\BCAY{Ruff, Kauffmann, Vandermeulen, Montavon, Samek, Kloft, Dietterich,\ \BBA\ Müller}{Ruff et~al.}{2021}]{ruff_unifying_2021}
Ruff, L., Kauffmann, J.~R., Vandermeulen, R.~A., Montavon, G., Samek, W., Kloft, M., Dietterich, T.~G., \BBA\ Müller, K.-R. \BBOP2021\BBCP.
\newblock \BBOQ A {Unifying} {Review} of {Deep} and {Shallow} {Anomaly} {Detection}\BBCQ\
\newblock {\Bem Proceedings of the IEEE}, {\Bem 109\/}(5), 756--795.
\newblock Conference Name: Proceedings of the IEEE.

\bibitem[\protect\BCAY{Sarfraz, Sharma,\ \BBA\ Stiefelhagen}{Sarfraz et~al.}{2019}]{sarfraz_efficient_2019}
Sarfraz, S., Sharma, V., \BBA\ Stiefelhagen, R. \BBOP2019\BBCP.
\newblock \BBOQ Efficient {Parameter}-{Free} {Clustering} {Using} {First} {Neighbor} {Relations}\BBCQ\
\newblock In {\Bem Proceedings of the IEEE Conference on Computer Vision and Pattern Recognition (CVPR)}, \BPGS\ 8934--8943.

\bibitem[\protect\BCAY{Scheirer, Rocha, Sapkota,\ \BBA\ Boult}{Scheirer et~al.}{2013}]{scheirer_toward_2013}
Scheirer, W.~J., Rocha, A. d.~R., Sapkota, A., \BBA\ Boult, T.~E. \BBOP2013\BBCP.
\newblock \BBOQ Toward {Open} {Set} {Recognition}\BBCQ\
\newblock {\Bem IEEE Transactions on Pattern Analysis and Machine Intelligence}, {\Bem 35\/}(7), 1757--1772.

\bibitem[\protect\BCAY{Schiappa, Rawat,\ \BBA\ Shah}{Schiappa et~al.}{2022}]{schiappa_self-supervised_2022}
Schiappa, M.~C., Rawat, Y.~S., \BBA\ Shah, M. \BBOP2022\BBCP.
\newblock \BBOQ Self-{Supervised} {Learning} for {Videos}: {A} {Survey}\BBCQ\
\newblock {\Bem arXiv:2207.00419 [cs]}, {\Bem ""}.

\bibitem[\protect\BCAY{Smaira, Carreira, Noland, Clancy, Wu,\ \BBA\ Zisserman}{Smaira et~al.}{2020}]{smaira_short_2020}
Smaira, L., Carreira, J., Noland, E., Clancy, E., Wu, A., \BBA\ Zisserman, A. \BBOP2020\BBCP.
\newblock \BBOQ A {Short} {Note} on the {Kinetics}-700-2020 {Human} {Action} {Dataset}\BBCQ\
\newblock {\Bem arXiv:2010.10864 [cs]}, {\Bem ""}.
\newblock arXiv: 2010.10864.

\bibitem[\protect\BCAY{Tishby, Pereira,\ \BBA\ Bialek}{Tishby et~al.}{1999}]{tishby_information_1999}
Tishby, N., Pereira, F.~C., \BBA\ Bialek, W. \BBOP1999\BBCP.
\newblock \BBOQ The {Information} {Bottleneck} {Method}\BBCQ\
\newblock {\Bem The 37th annual Allerton Conference on Communication, Control, and Computing}, {\Bem ""}, 368--377.
\newblock arXiv: physics/0004057.

\bibitem[\protect\BCAY{Wu, Girshick, He, Feichtenhofer,\ \BBA\ Krahenbuhl}{Wu et~al.}{2020}]{wu_multigrid_2020}
Wu, C.-Y., Girshick, R., He, K., Feichtenhofer, C., \BBA\ Krahenbuhl, P. \BBOP2020\BBCP.
\newblock \BBOQ A {Multigrid} {Method} for {Efficiently} {Training} {Video} {Models}\BBCQ\
\newblock In {\Bem Proceedings of the IEEE Conference on Computer Vision and Pattern Recognition (CVPR)}, \BPGS\ 153--162.

\bibitem[\protect\BCAY{Zaeemzadeh, Bisagno, Sambugaro, Conci, Rahnavard,\ \BBA\ Shah}{Zaeemzadeh et~al.}{2021}]{zaeemzadeh_out--distribution_2021}
Zaeemzadeh, A., Bisagno, N., Sambugaro, Z., Conci, N., Rahnavard, N., \BBA\ Shah, M. \BBOP2021\BBCP.
\newblock \BBOQ Out-of-{Distribution} {Detection} {Using} {Union} of 1-{Dimensional} {Subspaces}\BBCQ\
\newblock In {\Bem Proceedings of the {IEEE}/{CVF} {Conference} on {Computer} {Vision} and {Pattern} {Recognition}}, \BPGS\ 9452--9461.

\end{thebibliography}
\bibliographystyle{theapa}

\end{document}